\documentclass[pdflatex,sn-mathphys]{sn-jnl}
\usepackage{amsmath} 
\usepackage{amssymb} 
\usepackage{placeins} 
\jyear{2021}%

\usepackage{array,booktabs} 
\begin{document}

\title[Conditional Feature Importance for Mixed Data]{Conditional Feature Importance for Mixed Data}


\author*[1,2]{\fnm{Kristin} \sur{Blesch}}\email{blesch@leibniz-bips.de}

\author[3]{\fnm{David S.} \sur{Watson}}\email{david.watson@kcl.ac.uk}

\author[1,2,4]{\fnm{Marvin N.} \sur{Wright}}\email{wright@leibniz-bips.de}

\affil*[1]{\orgname{Leibniz Institute for Prevention Research \& Epidemiology – BIPS}, \orgaddress{\city{Bremen}, \country{Germany}}}

\affil[2]{\orgdiv{Faculty of Mathematics and Computer Science}, \orgname{University of Bremen}, \orgaddress{ \city{Bremen}, \country{Germany}}}

\affil[3]{\orgdiv{Department of Informatics}, \orgname{King's College London}, \orgaddress{\city{London}, \country{United Kingdom}}}

\affil[4]{\orgdiv{Department of Public Health}, \orgname{University of Copenhagen}, \orgaddress{\city{Copenhagen}, \country{Denmark}}}


\abstract{
Despite the popularity of feature importance (FI) measures in interpretable machine learning, the statistical adequacy of these methods is rarely discussed. From a statistical perspective, a major distinction is between analyzing a variable's importance before and after adjusting for covariates -- i.e., between \textit{marginal} and \textit{conditional} measures. Our work draws attention to this rarely acknowledged, yet crucial distinction and showcases its implications.

We find that few methods are available for testing conditional FI, and practitioners have hitherto been severely restricted in method application due to mismatched data requirements. Most real-world data exhibits complex feature dependencies and incorporates both continuous and categorical features (i.e., mixed data). Both properties are oftentimes neglected by conditional FI measures.

To fill this gap, we propose to combine the conditional predictive impact (CPI) framework with sequential knockoff sampling. The CPI enables conditional FI measurement that controls for any feature dependencies by sampling valid knockoffs -- hence, generating synthetic data with similar statistical properties -- for the data to be analyzed. Sequential knockoffs were deliberately designed to handle mixed data and thus allow us to extend the CPI approach to such datasets.

We demonstrate through numerous simulations and a real-world example that our proposed workflow controls type I error, achieves high power, and is in line with results given by other conditional FI measures, whereas marginal FI metrics can result in misleading interpretations. Our findings highlight the necessity of developing statistically adequate, specialized methods for mixed data.}


\maketitle
\section{Introduction}\label{sec::Intro}
Interpretable machine learning is on the rise as practitioners become interested in not only achieving high prediction accuracy in supervised learning tasks, but also understanding why certain predictions were made. Evaluating the importance of input variables (features) to the target prediction plays a crucial role in facilitating such endeavours. Several feature importance (FI) measures have been proposed by the machine learning community, but differing conceptualizations are spread across the literature.

{ 
We identify at least five dichotomies that orient FI methods: (1) global vs. local; (2) model-agnostic vs. model-specific; (3) testing vs. scoring; (4) methods that do and do not accommodate mixed tabular data; and (5) conditional vs. marginal measures. This defines a grid with $2^5 = 32$ cells that helps categorize FI measures. For example, the popular SHAP algorithm \citep{lundberg2017} produces local, model-agnostic FI scores that can accommodate mixed data and measures marginal FI. We emphasize that there is no ``ideal'' configuration of these five options---each is the right answer to a different question that is irreducibly context-dependent. However, this grid helps identify a notable lacuna: There are few global, model-agnostic FI methods that accommodate mixed data with error control for conditional FI measurement. 

Explaining the dichotomies in more detail, local FI measures \citep{lundberg2017, ribeiro2016} are optimized for a particular point or region of the feature space, e.g. a single observation, while global FI scores \citep{fisher2019, friedman2001} measure a variable's overall importance. Model-specific measures \citep{breiman2001, kursa2010, shrikumar2017} exploit the properties of a particular function class for more efficient or precise FI calculation, while model-agnostic measures \citep{apley2020, ribeiro2018} treat the underlying model as a black box. Testing methods include some inference procedure for error control \citep{lei2018}, while scoring methods \citep{covert2020} do not. Some methods are proposed with limited applicability to certain data types, e.g., only continuous inputs \citep{watson2021}, while others are more flexible \citep{molnar2023}. We discuss a selection of FI methods briefly in Section~\ref{sec::methods}, but refer readers to review papers on FI interpretability methods, e.g. \cite{linardatos2020}, for a wider discussion on the topic.

Through the lens of statistics, the division (5), conditional vs. marginal measures, is particularly important yet insufficiently acknowledged in both literature and practice \citep{apley2020, hooker2021, molnar2023, watson2021}. The complementary concepts become evident when relating the statistical conception of independence testing to the machine learning view on FI measurement. We can think of the marginal null hypothesis as testing whether the input feature $X_j$ is independent of other covariates ${X_{-j}}$ or the target variable $Y$: 
\begin{equation}
    H_{0}^M: X_j \perp\!\!\!\perp \{Y, X_{-j}\}\label{eq::marginal}
\end{equation}
On the other hand, testing against \eqref{eq::conditional} accounts for the covariates $X_{-j}$ and hence corresponds to conditional FI: 
\begin{equation}
    H_{0}^C: X_j \perp\!\!\!\perp Y \mid X_{-j} \label{eq::conditional}
\end{equation} 
These tests clearly target different objectives. In this setup, we have $H_0^M$ entailing $H_0^C$, but not the other way around. However, this strength comes with a certain loss of specificity, because rejecting $H_{0}^M$ leaves it unclear whether $X_j$ is correlated with $Y$, $X_{-j}$, or both.

The relationship between FI and independence testing sheds light on another aspect, which may even be considered another dichotomy: does the FI measure aim to investigate model behaviour or the underlying data structure \citep{chen2020}? For example, conditional independence tests that are part of some conditional FI measures \citep{watson2021} may be used for causal structure learning, which often is based on repeated conditional independence testing \citep{glymour2019}. Therefore, conditional FI measures can help explain the underlying data structure, whereas marginal FI measures differentiate between variables the predictive model relies on, which can be used to evaluate the fairness of a model. This does not preclude practitioners from using marginal and conditional FI measures in conjunction, and since marginal measures are often faster to compute, they might be preferable for quick assessments in large pipelines with many iterations. However, practitioners must be careful to interpret these measures properly and not infer a conditional signal from a marginal test. 

In Fig.~\ref{fig::marginal_vs_conditional}, we illustrate the difference between marginal \citep[permutation feature importance (PFI),][]{fisher2019, breiman2001} and conditional \citep[conditional predictive impact with Gaussian knockoffs (CPIgauss),][]{watson2021} FI measures. In this example, the confounding variable $C$ is a common cause of both $X$ and $Y$. This causal structure induces spurious correlation between $X$ and $Y$, leading the marginal FI measure to attribute nonzero importance to both $C$ and $X$ in predicting $Y$. On the contrary, the conditional FI measure attributes nonzero FI only to $C$, since $X$ has no \textit{additional} predictive value for $Y$ above $C$.

\begin{figure}[h] 
	\centering
	\includegraphics[width=.9\textwidth]{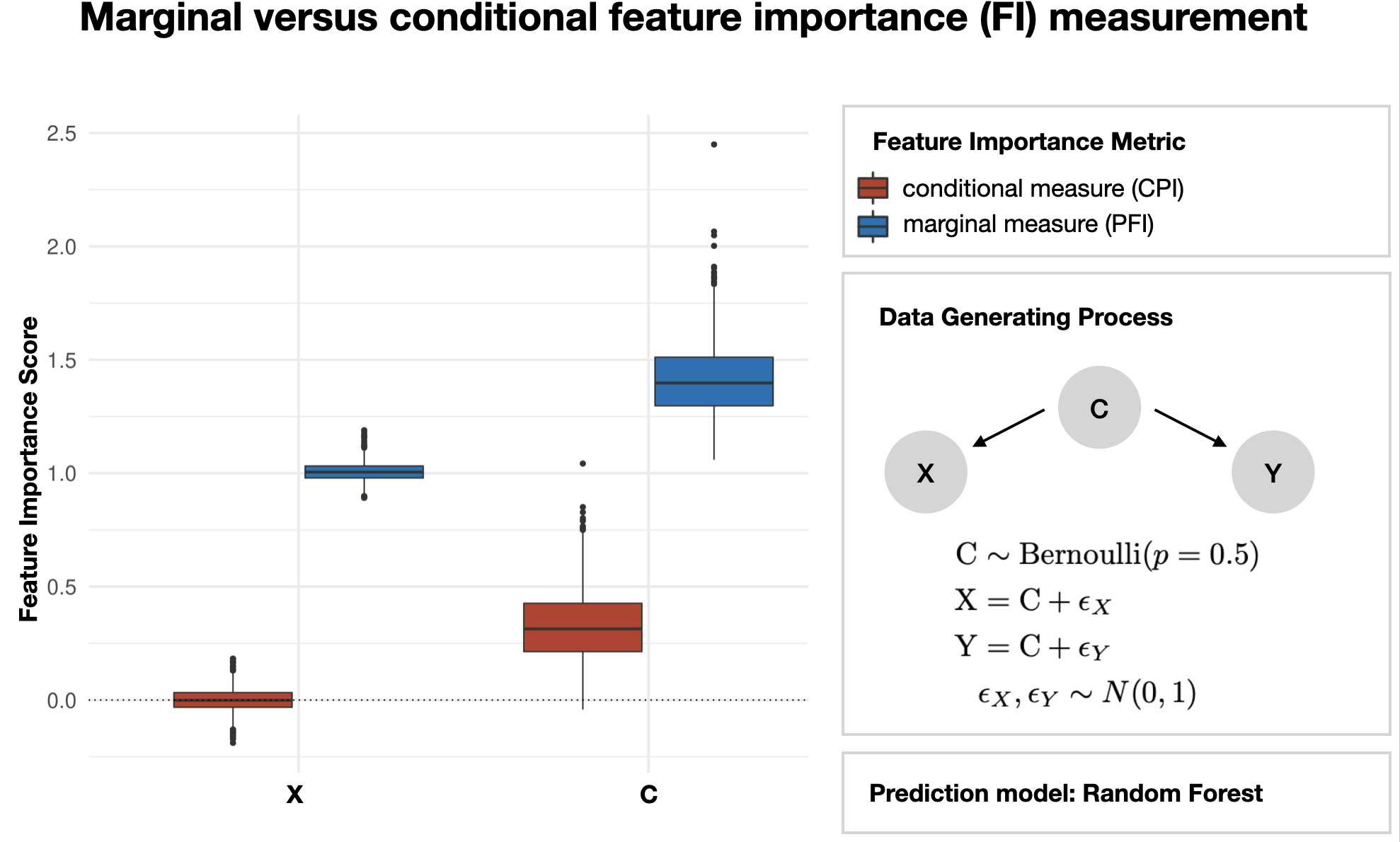}
	\caption{{ Boxplots contrasting marginal and conditional FI metrics for a prediction of $Y$ with $C, X$ ($N = 200$) through a random forest prediction model across $1\, 000$ replicates. The conditional FI measure attributes no importance to $X$, whereas the marginal measure attributes non-zero importance to $X$ because (due to induced correlation between $X$ and $Y$ by $C$) it is predictive of $Y$.}
	\label{fig::marginal_vs_conditional}}
\end{figure}

This paper explores global, model-agnostic FI methods that accommodate mixed data with error control for conditional FI measurement. This is not a niche problem: mixed tabular data is the norm in many important areas such as healthcare, economics, and industry, and inference procedures are essential for decision making in high risk domains to minimize costly errors. With the proliferation of machine learning algorithms, model-agnostic approaches can help standardize FI tasks without recalibrating to a particular function class for each new application. Conditional, global measures are valuable when practitioners seek mechanistic understanding that takes data covariance into account and go beyond individual model outputs. 

Even though the empirical relevance of this kind of FI measurement is eminent, specialized methods are lacking. Some FI methods have yet to be evaluated in mixed data settings \citep{covert2020, molnar2023, lei2018}, while others are currently inapplicable \citep{watson2021}. The consequences of neglecting the special nature of mixed data for conditional FI measurement remain unexplored, and therefore practitioners currently have no guidance on how to proceed with conditional FI measurement in such cases, which proves a severe limitation in real-world applications.}

We propose to combine the \textbf{conditional predictive impact} (CPI) testing framework proposed by \citet{watson2021} with the use of \textbf{sequential knockoffs} \citep{kormaksson2021} in order to enable conditional, global, model-agnostic FI testing for mixed data. CPI is a flexible, model-agnostic tool that relies on the usage of so-called knockoffs \citep{candes2018}. In short, knockoffs are synthetic variables that carry over the major statistical properties of the original variables, such as the correlation structure among covariates. While \cite{watson2021} claim that the CPI should in principle work with any valid set of knockoffs, it has thus far only been applied and evaluated with Gaussian knockoffs \citep{candes2018}. This currently limits practitioners to using the CPI method only with continuous variables  or to disregard the specialities of mixed data. We analyse consequences of such a disregard when using CPI with Gaussian knockoffs \citep{candes2018} (CPIgauss) and deep knockoffs \citep{romano2020} (CPIdeep) and propose a specialized solution strategy to tackle the mixed data case: using sequential knockoffs \citep{kormaksson2021} -- a knockoff sampling algorithm explicitly developed for mixed data -- within the CPI framework (CPIseq).

The paper will be structured as follows. We present relevant methodology and FI measures in Section~\ref{sec::methods}. Section~\ref{sec::knockoffs} reviews several knockoff sampling algorithms, demonstrating the need for specialized procedures with mixed data and motivating our proposed solution, CPIseq. Through simulation studies in Sections~\ref{sec::experiments1} and \ref{sec::experiments2}, we will evaluate our newly proposed workflow in more depth and further compare it to other methods. Finally, we illustrate method application to a real-world dataset in Section~\ref{sec::real_data} before concluding and discussing our findings in Section~\ref{sec::conclusion}. 

\section{Methods} \label{sec::methods}
With a focus on the measurement of model-agnostic, global, conditional FI, this section presents related measures proposed by previous literature and discusses their applicability to mixed data. We acknowledge that methods from the statistical literature on conditional independence testing \citep{shah2020, williamson2021} might also be utilized for conditional FI measurement, however, a full comparison of such methods is beyond the scope of this paper. Further, it is worth clarifying at this point that we understand FI here as a concept that is tied to the variable's effect on the predictive performance in a supervised learning task.

\subsection{Feature Importance Measures}
\paragraph{Conditional subgroup approach (CS)}
A global, model-agnostic FI measure that acknowledges the crucial distinction between conditional and marginal measures of importance is the \textbf{conditional subgroup (CS)} approach proposed by \cite{molnar2023}. CS partitions the data into interpretable subgroups, i.e. groups whose feature distributions are homogeneous within but heterogeneous between groups. The method is promising, as it explicitly specifies the conditioning between subgroups and further allows for an unconditional interpretation within subgroups. This means the method provides both a global conditional and a within-group unconditional interpretation, which sheds light on feature dependence structures. 

To determine FI, CS evaluates the change in loss when the variable of interest is permuted within subgroups, which lowers extrapolation to low-density regions of the feature space, thereby mitigating a common problem with permutation-based approaches \citep{hooker2021}. To decide on a suitable partition, the authors suggest determining subgroups via transformation trees. Using a pre-specified loss function, the average increase in loss is reported for multiple permutations versus the original ordering of variables. 

CS is not affected by mixed data other than through the choice of an appropriate prediction algorithm, which is why this method is suspected to work equally well with mixed data. However, for this approach to work, researchers must assume that the data is separable into subgroups. Further, for testing FI, the method would need to rely on computationally expensive permutation tests, as no inherent testing procedure is provided.

\paragraph{Leave-One-Covariate-Out (LOCO)}
\textbf{Leave-one-covariate-out (LOCO)}  is a fairly simple approach to measuring FI, which, as the name suggests, evaluates the change in predictive performance of a model when leaving out a covariate of interest  \citep{lei2018}. This means, FI is determined by comparing the loss of the model fitted including or excluding the covariate of interest.

While this is a very intuitive approach, it does involve several drawbacks. First, the model has to be retrained with a different set of variables, which not only incurs high computational cost, but also yields an entirely different model, raising concerns about comparability in general. Further, if correlations or other complex dependencies are present within the data, LOCO might give misleading results if only one covariate at a time is excluded, as this neglects potential interaction effects between groups of variables. In the presence of such group-wise structures, the exclusion of multiple covariates at a time is advisable \citep{au2022, rinaldo2016}.

For the speciality of mixed data, we can again see that all reliance is on the level of model choice, hence, as long as the prediction model is able to process mixed data, LOCO is not affected by different data types.

\paragraph{Shapley Additive Global Importance (SAGE)}
\textbf{Shapley Additive Global Importance (SAGE)} \citep{covert2020} is a model-agnostic FI measure that aims to take into account feature interactions on a global level. The method is based on Shapley values \citep{shapley1953}, which have received much attention in interpretable machine learning recently. While Shapley values are widespread in their use for giving local explanations, i.e. explaining the role of features in individual predictions made by the model, \cite{covert2020} propose a global extension such that the role of features can be understood on a model-wide level. SAGE values are Shapley values for the features with regard to the predictive power of the model. Therefore, SAGE values can also be calculated by directly calculating  Shapley values for the model loss, e.g. as proposed in LossSHAP \citep{lundberg2020}, and then average across all instances to achieve a global measure. However, \cite{covert2020} propose a fast approximation algorithm. 

The SAGE methodology allows for taking feature interaction effects into account, however, in practice,  implementations typically use marginal sampling as an approximation to the conditional densities when sampling to replace the respective feature in various coalitions. This results in explanations that are comparable to marginal measures of FI when applied to real-world data.

Mixed data affects SAGE at the variable sampling step to build the coalitions and through the choice of the predictive model. With the use of marginal imputation and a model that is able to process mixed data, SAGE should not be affected by mixed data types. 

\paragraph{Conditional Predictive Impact (CPI)} 
A fairly general approach to tackle conditional FI measurement is the \textbf{conditional predictive impact (CPI)} proposed by \cite{watson2021}. To capture conditional FI, a flexible conditional independence test is introduced that works with any supervised learning algorithm, valid knockoff sampler and well-defined loss function. CPI ties FI to predictive performance, arguing that the inclusion of a relevant variable in the model should improve its predictive performance. Building on this idea, first, a supervised learning algorithm is trained to predict the outcome from given input variables. Then, using a knockoff sampling algorithm, so-called knockoff copies of the input features are generated. These knockoffs retain the covariance structure of the input features\footnote{This holds true for knockoffs that are at least of second-order, i.e. exhibit the same first two moments as the original data.}  but are (conditional on the input features) independent of the response variable. They therefore serve as a set of negative controls against which to compare the original data. In detail, to compute the CPI statistic, the trained model from the first step is used to predict the target twice: first using the original test data, and again after replacing one or several features of interest in the test data by their knockoff copies. The change in loss is then averaged across samples. Finally, the authors propose to apply inference procedures, such as a paired $t$-test, to get valid $p$-values and confidence intervals for the FI scores. 

Given that the prediction algorithm works with mixed data, sampling valid knockoffs for mixed data is the sticking point. As \cite{watson2021} claim, the CPI setup is knockoff-agnostic and hence works for any knockoff sampler. However, their simulations are limited to settings of continuous data and Gaussian knockoff sampling, i.e. using CPIgauss, only. Resulting from this, practitioners facing mixed data cannot use CPIgauss directly and are forced to use workarounds that may perform poorly in practice, e.g. dummy encoding variables and treating them as continuous, of which the effects on the method are thus far unknown. The present work sheds light on the consequences of such procedures, see further Section~\ref{sec::experiments1}. To propose an efficient way of making CPI applicable to mixed data, we will now delve into the methodology of knockoffs in greater depth.

\subsection{Model-X Knockoffs}\label{sec::knockoffs}
The model-X knockoff framework \citep{candes2018} was proposed for variable selection while controlling the false discovery rate (FDR). The idea is to use knockoffs as negative controls in the model, which prevents spuriously correlated variables from being detected as important. These knockoffs are a set of variables $\Tilde{\mathbf{X}}$ that mimic the correlational structure between the original input variables $\mathbf{X}$, but crucially are known to be irrelevant to the target variable $Y$, conditional on the input data. Intuitively, if $X_j$ does not significantly outperform $\Tilde{X_j}$ by some importance measure, then $X_j$ can be removed from the model \citep{candes2018}.

More formally, to construct a valid knockoff matrix $\Tilde{\mathbf{X}}$ for the $p$-dimensional feature matrix $\mathbf{X}$, two conditions have to be met. The first is pairwise exchangeability, i.e. for any proper subset $S \subset \{1, \dots, p\}$: \begin{equation}
   (\mathbf{X}, \mathbf{\Tilde{X})}_{swap(S)} \overset{d}{=} (\mathbf{X}, \mathbf{\Tilde{X})}, \label{eq::knockoffs_a}
\end{equation} where $ \overset{d}{=} $ represents equality in distribution and \textit{swap($S$)} indicates swapping the respective variables in $S$ with their knockoff counterparts. The second condition is conditional independence, i.e.

\begin{equation}
    \mathbf{\Tilde{X}} \perp\!\!\!\perp Y\mid \mathbf{X}. \label{eq::knockoffs_b}
\end{equation} 
Knockoff methodology is an active field of research. Numerous approaches to knockoff sampling have been proposed, for example, methods based on distributional assumptions \citep{bates2021, candes2018, seisa2018}, Bayesian frameworks \citep{gu2021} or deep learning \citep{jordon2019, liu2018, romano2020, sudarshan2020}. While a comprehensive review of knockoff samplers is beyond the scope of this paper, we will present a selection of knockoff samplers that is particularly interesting for applications on mixed data. Namely, we will investigate Gaussian knockoffs \citep{candes2018} because of their widespread use, deep knockoffs \citep{romano2020} as a representative of deep learning based knockoff generation, and sequential knockoffs as a specialized approach to tackle mixed data.

\paragraph{Gaussian Knockoffs}
As the name suggests, the \textbf{Gaussian knockoff} sampler  \citep{candes2018} is based on the assumption that the input data matrix $\mathbf{X} \in \mathbb{R}^{N \times p}$ is multivariate Gaussian, i.e. $ \mathbf{X} \sim N(\mu, \mathbf{\Sigma})$. The joint distribution which satisfies Eq. \ref{eq::knockoffs_a} is then 
\[
(\mathbf{X}, \mathbf{\Tilde{X}}) \sim N(0, \textbf{G}), \text{ where } \textbf{G} =\begin{bmatrix}
    \mathbf{\Sigma}       & \mathbf{\Sigma} - \text{diag}\{s\} \\
    \mathbf{\Sigma} - \text{diag}\{s\}     & \mathbf{\Sigma} \\

\end{bmatrix}
\]
with diagonal matrix $\text{diag}\{s\}$ to ensure positive semi-definiteness of the joint covariance matrix $\textbf{G}$. Knockoffs can then be sampled from the conditional distribution $\mathbf{\Tilde{X}} \mid \mathbf{X} \overset{d}{=} N( \mu , \textbf{V})$, where $\mu, \textbf{V}$ can be calculated from regular regression formulas. For details see \cite{candes2018}.

Clearly, it is reasonable to suspect this knockoff sampler to work well with Gaussian data. However, with mixed data types, discrete values can only be handled after encoding, e.g. introducing dummy variables, which are evidently non-Gaussian. The consequences of such transformations, i.e. neglecting the special nature of mixed data, have not yet been evaluated for the Gaussian knockoff sampler. In an attempt to quantify such implications to some extent, we will include this knockoff sampler in our analysis in Section~\ref{sec::experiments1} and compare it to more well-suited alternatives.

\paragraph{Deep Knockoffs} 
\textbf{Deep knockoffs} as proposed by \cite{romano2020} rely on a random generator, consisting of a deep neural network, to sample valid knockoffs. For variables $\mathbf{X}$ sampled independently from an unknown distribution $P_\mathbf{X}$, the random generator is trained such that the joint distribution of $(\mathbf{X}, \mathbf{\Tilde{X}})$ is invariant under swapping, such that Eq. \eqref{eq::knockoffs_a} is satisfied. In detail, the neural network takes variables $\mathbf{X}$ and i.i.d. sampled noise $\mathbf{\mathcal{E}}$ as input to optimize a scoring function that quantifies the extent to which $\mathbf{\Tilde{X}}$ is a good knockoff copy for $\mathbf{X}$ by evaluating how well Eq. \eqref{eq::knockoffs_a} is approximated. 
Considering the neural network architecture, the authors suggest using a width $h$ that is ten times the dimensionality of the input feature space, i.e. $h ~= 10p$ and $6$ hidden layers which they claim should work well for a ``wide range of scenarios", but acknowledge that ``more effective designs" might be found \citep{romano2020}. Making use of recent deep learning advances, deep knockoffs should -- according to the authors -- generalize well to the mixed data case. \cite{romano2020} claim that this framework samples approximate knockoffs for arbitrary distributions. However, it is worth noting that there is little explicit methodology available to the user beyond making general claims about the generalizability of the method. Therefore, an applied user is again left with a knockoff sampler that does not return valid mixed data knockoffs.\footnote{Deep generative models for mixed data is an active and promising area of research \citep{xu2019, watson2022}, although we are unaware of any implementation for knockoff sampling.}

\paragraph{Sequential Knockoffs}
\textbf{Sequential knockoff} \citep{kormaksson2021} sampling is based on the conditional independent pairs algorithm \citep{candes2018} given in Supplementary Information A 
with a specialized strategy to model the conditional distribution ${P(X_j \mid X_{-j}, \Tilde{X}_{i:j-1}}$) and sample knockoffs for mixed data.

Sequential knockoffs are synthesized by sampling continuous knockoffs from a Gaussian distribution and categorical knockoffs from a multinomial distribution with distribution parameters that have been sequentially estimated through penalized\footnote{In our experiments, we follow the advice of \cite{kormaksson2021} to use an elastic net \citep{zou2005}. Note that the ordering of variables might be of relevance in finite samples and that the procedure requires the various levels of the categorical variable to occur sufficiently often.} linear or multinomial logistic regression models. The procedure is given in more detail in Algorithm \ref{alg::seq_ko}, where $ {X_{-j} := (X_1, \dots, X_{j-1}, X_{j+1}, \dots X_p)}$ and ${\Tilde{X}_{1:j-1} := (\Tilde{X_1}, \dots, \Tilde{X}_{j-1})}$.
\begin{algorithm}
\caption{Sequential knockoffs through conditional independent pairs}\label{alg::seq_ko}
\begin{algorithmic}
\State $j = 1$
\While{ $j \leq p$} 
\If{$X_j$ continuous}
\State sample $\Tilde{X_j}$ from $N(\hat{\mu}, \hat{\sigma})$ with $\hat{\mu}, \hat{\sigma}$ estimated from 
\State penalized linear regression ${\Tilde{X_j}} \sim X_{-j}, \Tilde{X}_{1:j-1}$
\ElsIf{$X_j$ categorical}
\State sample $\Tilde{X_j}$ from $Multinom(\hat{\pi})$ with $\hat{\pi}$ estimated from
\State penalized multinomial logistic regression $\Tilde{X}_j \sim X_{-j}, \Tilde{X}_{1:j-1}$
\EndIf
\State $j = j+1$
\EndWhile
\end{algorithmic}
\end{algorithm}

Algorithm \ref{alg::seq_ko} yields valid knockoff copies for data that may consist of both categorical and continuous covariates. Hence, the present paper puts a special focus on this method and evaluates its suitability for conditional FI measurement with mixed data. 
\newpage
\subsection{CPI with Sequential Knockoffs: CPIseq}\label{sec::our_workflow}
We propose to combine two frameworks that have, thus far, not been analyzed in conjunction, the CPI \citep{watson2021} and sequential knockoffs \citep{kormaksson2021}, as a viable solution for conditional FI measurement with mixed data. Section~\ref{sec::methods} revealed that amongst the limited number of conditional FI measurement methods available, CPI is one of the few conditional FI methods that allows for the direct application of statistical testing procedures. Further, we have seen that the major obstacle of CPI with mixed data is the knockoff generation step. When surveying the literature on knockoffs in Section~\ref{sec::knockoffs}, the sequential knockoff sampler stands out as a solution that tackles the special nature of mixed data.  Algorithm \ref{alg::cpi_seq} presents details on the procedure we propose here. Note that for calculating CPIseq for several features (or groups) $j$, steps $1$ and $2$ of the algorithm do not have to be recalculated for each $j$. 

\begin{algorithm}
\caption{CPI with sequential knockoffs (CPIseq)}\label{alg::cpi_seq}
\begin{algorithmic}[1]
\Require{$(X^{train}, Y^{train}), (X^{test}, Y^{test})$, supervised learner $f$, feature (or group) of interest $j$, sequential knockoff sampler $s$ (Alg. \ref{alg::seq_ko}), loss function $L$, inference procedure $h$ }
\State learn $\hat{f} \gets f(X^{train}, Y^{train})$
\State sample knockoffs $\Tilde{X}^{test} \gets s(X^{test})$
\State for feature (or group) $j$ calculate instance-wise loss difference 

$\Delta^{(i)} \gets L(\hat{f}, X^{test(i)}) - L(\hat{f}, \{X^{test(i)}_{-j}, \Tilde{X}^{test(i)}_{j}\})$
\State calculate conditional predictive impact

$\widehat{\text{CPI}} \gets \frac{1}{N}\sum_{i=1}^N \Delta^{(i)}$
\State apply inference procedure for $p$-value $p$ and confidence interval $ci$

$p, ci \gets h(\Delta)$
\Ensure{$\widehat{\text{CPI}}, p, ci$}
\end{algorithmic}
\end{algorithm}
{ 
The CPIseq we propose here combines the features of the CPI methodology with ease of applicability to real data, which often consists of mixed data types. Providing frequentist  inference procedures without model refitting is the major advantage over other conditional FI methods, such as CS and LOCO. To ensure high power for these testing procedures, adequate handling of mixed data is a prerequisite and CPIseq assures this through the flexible sequential knockoff subroutine. 
\section{Experiments}
In this section, we analyze the performance of various FI measures on both simulated and empirical data. Through simulation studies, we evaluate the performance of our newly proposed workflow in comparison to other approaches. First, we investigate how CPIseq compares to CPI with other knockoff samplers, namely CPIgauss and CPIdeep (Section~\ref{sec::experiments1}) in terms of power and effective control of false positives. Further, we compare feature rankings given by our proposed approach and other conditional FI related measures that do not use knockoffs (Section~\ref{sec::experiments2}). Finally, we use a real-world data example to illustrate method application (Section~\ref{sec::real_data}).}
\subsection{Comparing Knockoffs} \label{sec::experiments1}
\begin{figure}[h] 
	\centering
	\includegraphics[trim={0 0 0 0},clip, width=1\textwidth]{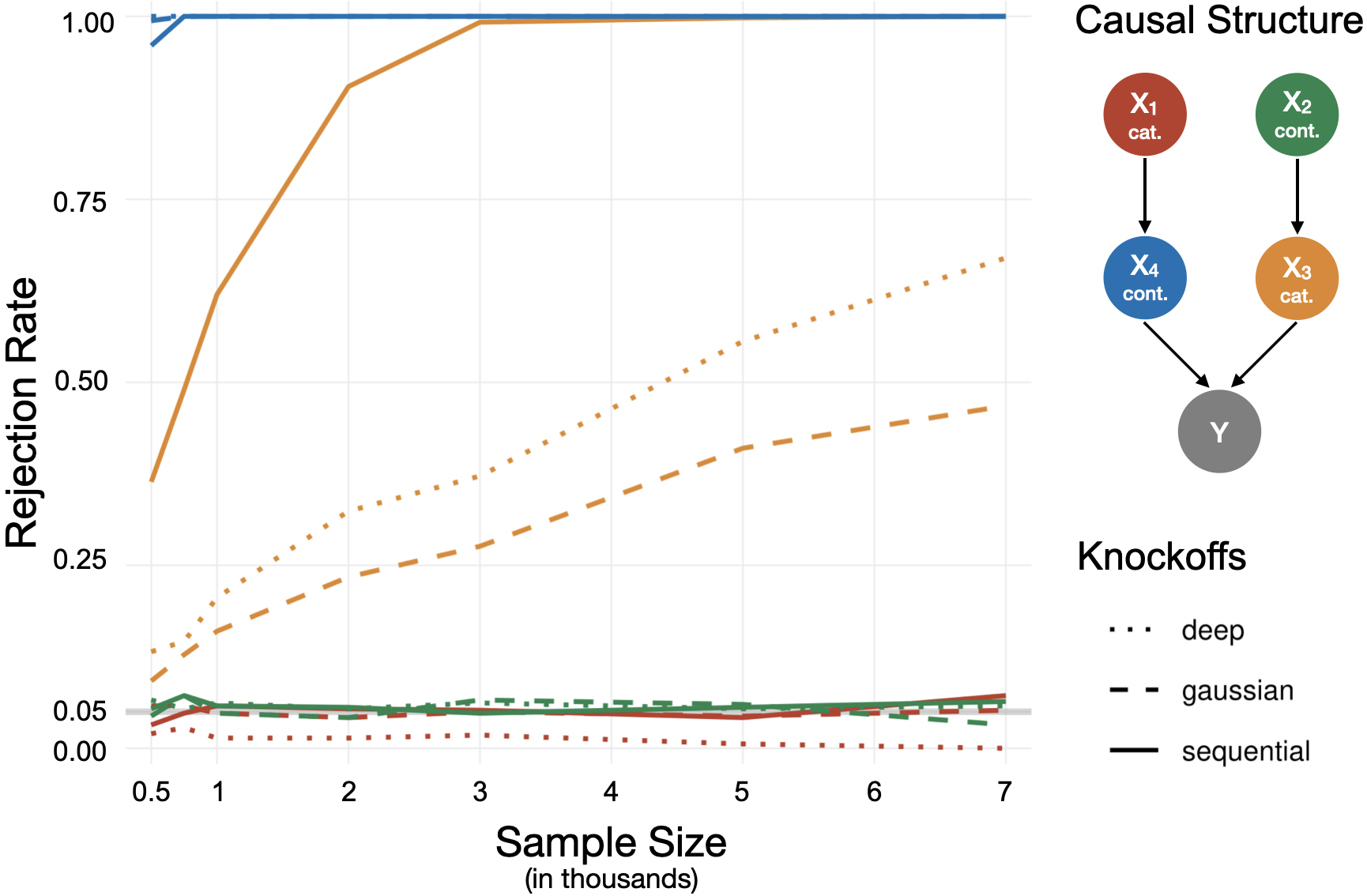}
	\caption{Rejection rates of one-sided paired $t$-tests at $\alpha ~= 0.05$ to detect relevant variables, i.e. power and type I error rates, for CPI with various knockoff samplers across $500$ simulation runs. $X_1, X_3$ are 10-level categoricals, $X_2, X_4$ are Gaussian. Effect size $\beta ~= 0.5$ {  and random forest prediction model}.}
	\label{fig::sim1}
\end{figure}

Major differences in the performance of CPIgauss, CPIdeep, and CPIseq on mixed data are illustrated using the following simulation setup. 
Consider a linear system of input variables $S =~\{X_1, X_2, X_3, X_4 \}$ and target variable $Y$, visualized by the directed acyclic graph (DAG) $\mathcal{G}$ in Fig.~\ref{fig::sim1}. Since the joint distribution is Markov with respect to $\mathcal{G}$, it follows by $d$-separation \citep{pearl2009} that $X_1 \perp\!\!\!\perp Y \mid S \setminus \{X_1\}$ and $X_2 \perp\!\!\!\perp Y \mid S \setminus \{X_2\}$, whereas $X_3 \not\!\perp\!\!\!\perp Y \mid S \setminus \{X_3\}$  and $X_4  \not\!\perp\!\!\!\perp Y \mid S \setminus \{X_4\}  $. Therefore, a conditional FI measure should only attribute nonzero importance to variables $X_3, X_4$, but not to $X_1, X_2$.
We consider three scenarios to track consequences of mixed data closely. For the baseline scenario (I), $S$ will be Gaussian; for scenario (II), $X_1$ or $X_3$ will be binary; and in scenario (III), $X_1$ and/or $X_3$ will be categorical with $c \in \{4, 10\}$ levels. {  Scenarios (II) and (III) further include an all categorical setting, i.e. $S$ will be categorical, as a point of reference. We carefully select relevant combinations of category levels (2, 4 or 10), type of the target variable (continuous or binary) and fitted model (generalized linear model or random forest). See Supplementary Information B.1 to B.4  
for further details on the experimental setup, including details on the prediction models and their validation.}
\subparagraph{Results}
For scenario (I), we find CPI achieving high power and effective type I error control with every knockoff sampling algorithm. Naturally, as the data is Gaussian, we see CPIgauss achieving high power in this setting, see Supplementary Information Fig.~\ref{fig::all_gaussian}.
When transforming $X_1$ and $X_3$ into binary variables, (scenario (II)), we still observe high power and type I error control. 

For input data consisting of mixed data types where the categorical variables are of high-cardinality (scenario (III)), we can see from Fig.~\ref{fig::sim1} that the sequential knockoff sampler provides greater sensitivity than the deep or Gaussian alternatives across all tested sample sizes. Rejection rates for CPIseq grow quickly with sample size, reaching about $90\%$ power around $N~=2\,000$. By contrast, CPIgauss only reaches about $50\%$ and the deep knockoff sampler about $70\%$ power at the maximal $N~=7\,000$. In terms of type I error control, all methods seem to be robust against the categorical nature of the irrelevant variable $X_1$, as the rejection rate in Fig.~\ref{fig::sim1} is kept close to $\alpha ~= 0.05$ for all knockoff samplers.

{ A full presentation of results is given in Supplementary Information B.5, including Figures for the all categorical cases, for which we find similar results as in mixed data settings.} 

This simulation study demonstrates that the power of CPIgauss and CPIdeep might be severely affected by high-cardinality features. We find CPIseq to provide a powerful solution to conditional FI measurement, i.e. to detect conditionally important categorical features, whereas CPIgauss and CPIdeep are less sensitive with such data. It is worth noting that CPIgauss and CPIdeep perform surprisingly well when mixed data is limited to continuous and binary data types, even though Gaussian and deep knockoffs inevitably generate data outside the support of Boolean variables. Nevertheless, CPIseq appears to be the most powerful solution for conditional FI measurement with high-cardinality categorical data. 

\subsection{Comparing Feature Importance Measures} \label{sec::experiments2}
Through a simulation study, our newly proposed workflow CPIseq will now be set in comparison with LOCO \citep{lei2018}, CS \citep{molnar2023}, SAGE \citep{covert2020}, and permutation feature importance (PFI) \citep{breiman2001, fisher2019}. {Even though CPIgauss and CPIdeep have been shown to be outperformed by CPIseq in Section 3.1, we add these two methods to the simulation in order to provide a complete picture on how they relate to other measures of FI. Further enriching the picture of FI measure comparison, we discuss a random forest model-specific FI procedure \citep{kursa2010} and its performance in comparison to the other FI measures in the Supplementary Information C.6
.}

We simulate multivariate normal data with a pre-specified correlation structure to ensure a simple setup while incorporating a larger number of variables than in our toy example in Section~\ref{sec::experiments1}. Again, we transform several variables into categoricals, such that we end up with mixed data. We distinguish between variables having zero, weak, or strong effect on the outcome $Y$, and for the continuous variables we further separate variables with a linear or nonlinear effect on $Y$. Further, we ensure that there is an equal number of relevant and irrelevant variables, such that each relevant variable is correlated with exactly one irrelevant variable of the same type, yielding a total of $p ~= 12$ variables.{  In sum, we analyze a total of 24 settings by varying the correlation strength ($\rho$ = 0.5 or 0.8), type of target variable $Y$ (continuous or binary), varying number of category levels ($c$ = 2 or 5) and fitting various machine learning prediction models (generalized linear model, random forest or neural network), see Supplementary Information C.1 and C.2  
for further details.}

Some of the methods included in the comparison do not provide statistical testing procedures. Therefore, we will compare methods by their tendency to rank relevant features higher than irrelevant alternatives. By construction, $p~=6$ variables are relevant to the outcome, whereas the other $p~=6$ variables are not. Hence, when we ask the methods to rank the variables according to their importance, ideally, the $6$ relevant variables are ranked amongst the top $6$. We will use the area under the receiver operating characteristic curve (AUC) as a measure of performance and will further report sensitivity and 1-specificity for each of the methods. See further Supplementary Information C.3. 

\begin{figure}[h] 
	\centering
	\includegraphics[width=0.9\textwidth]{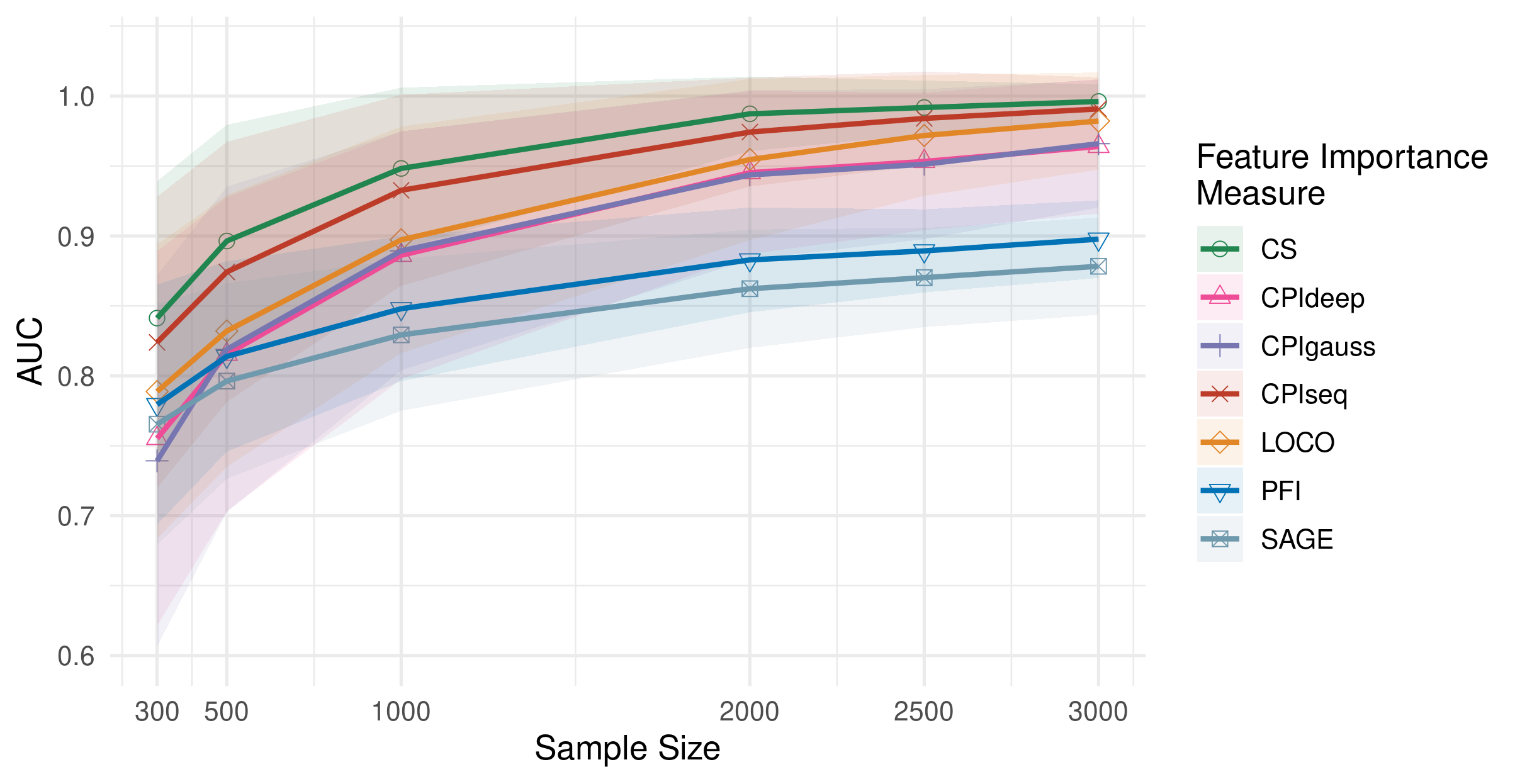}
	\caption{{  Mean AUC value with $\pm$ one standard deviation across $500$ simulation runs.  Categorical variables with $c=5$ levels, pairwise correlation $\rho ~= 0.8$ and a random forest prediction model for continuous target $Y$.}}
	\label{fig::auc}
\end{figure}

\begin{figure}[h] 
	\centering
	\includegraphics[trim={0 0 0 2cm},clip,width=1\textwidth]{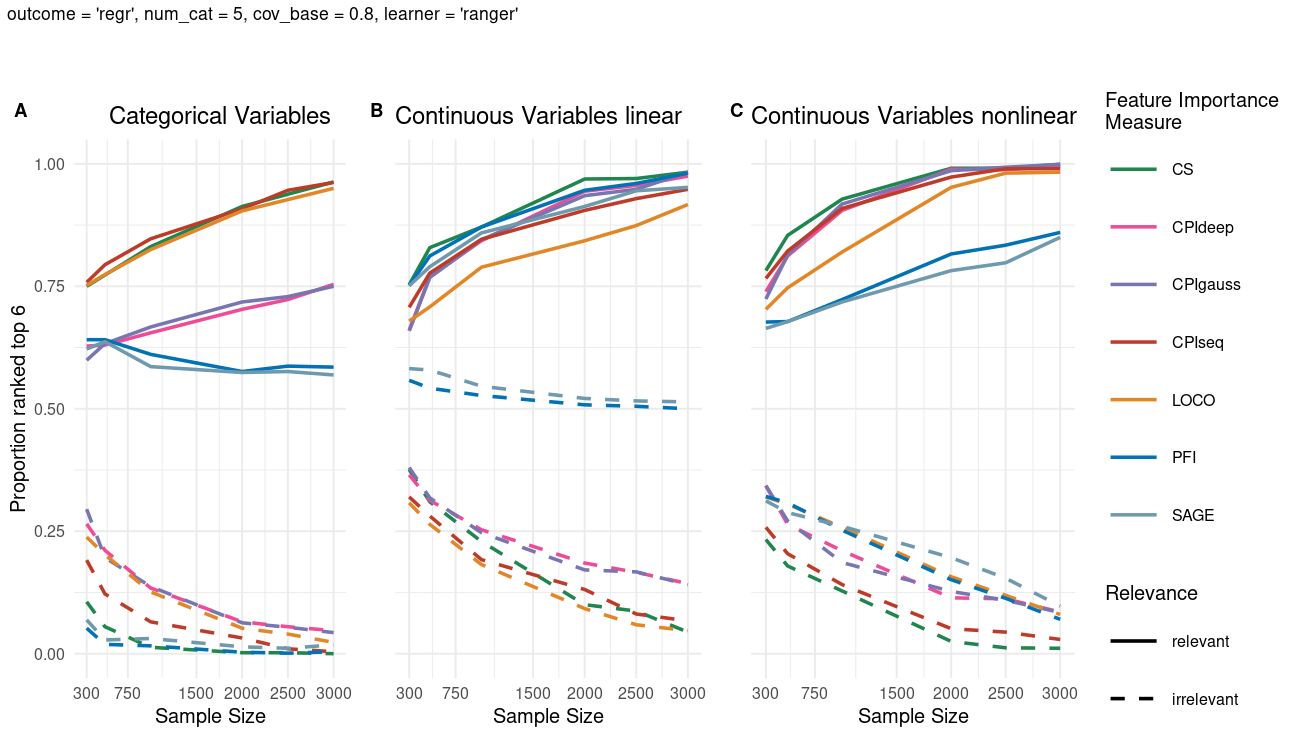}
	\caption{{  Proportion of features ranked amongst the top $6$ of $12$ by variable type across $500$ simulation runs. Solid lines (relevant variables) correspond to sensitivity, dashed lines (irrelevant variables) correspond to 1-specificity. Categorical variables with $c=5$ levels, pairwise correlation $\rho ~= 0.8$ and a random forest prediction model for continuous target $Y$.}}
	\label{fig::FI_trio}
\end{figure}
 
\subparagraph{Results} 
We find CS, CPIseq,CPIgauss, CPIdeep and LOCO outperforming PFI and SAGE in ranking the relevant variables amongst the top $6$ variables in terms of AUC scores (Fig.~\ref{fig::auc}). AUC scores rise with increasing sample size, however, while the conditional measures form a group that gets close to the optimal score of $1$, the performance of marginal measures\footnote{Note that SAGE here is closer to a marginal measure because of the marginal imputation subroutine.} flattens out. This behaviour stems from the phenomenon of marginal methods to attribute nonzero importance to correlated, but irrelevant variables, affecting the methods ability to separate the top $6$ from the bottom $6$ variables, as can be further investigated from Fig.~\ref{fig::FI_trio}.

Fig.~\ref{fig::FI_trio} depicts the proportion of the respective variable types being ranked amongst the top $6$ variables. Ideally, this proportion should be high for relevant variables (solid lines) and low for irrelevant variables (dashed lines). Panel (B) shows that both PFI and SAGE mistakenly rank the irrelevant continuous variables with a linear effect, which are correlated to the relevant continuous linear variables, amongst the top $6$ variables. This is unsurprising, because relevant continuous variables with a linear effect on the target are the easiest to detect, and hence, irrelevant variables correlated to these variables are most likely to be mixed up by marginal measures in the full ranking. Note that because each of the methods has to assign ranks $1-12$, an irrelevant variable being mistakenly ranked amongst the top $6$ variables in return leads to a relevant variable being ranked within the bottom 6 ranks. {  For example, due to the marginal measurement of FI, the PFI measure is ranking correlated yet irrelevant variables amongst the most important predictors (dashed line in Fig.~\ref{fig::FI_trio}, Panel B), which in turn forces PFI to mistakenly rank some relevant variables low (solid line in Fig.~\ref{fig::FI_trio}, Panel A).

Regarding the comparison of CPI based methods, we find CPIseq outperforming CPIgauss and CPIdeep in detecting relevant categorical variables in the mixed data setting, see Fig.~\ref{fig::FI_trio}, Panel A, which underpins the findings of simulations in Section 3.1. }

To check for robustness, we used several predictive models (generalized linear model, random forest, neural network), varied the type of the target variable (regression or classification task) and the number of categories for the categorical variables (2,5), and found similar results. {  Further, we analyzed the fit of the prediction models on test data to ensure reliable FI measurement. See Supplementary Information C.4 and C.5 for details on the robustness analyses.} 

In sum, this simulation demonstrates both that CPIseq is competitive with other conditional FI measures, and illustrates the importance of distinguishing between marginal and conditional measures. It is worth emphasizing again that the CPIseq workflow not only ranks features, but also enables powerful conditional FI testing. We will see the practical relevance of this in the following section.

\subsection{Real-World Data} \label{sec::real_data}
We conclude the section on experiments with a real-world data application to illustrate our proposed workflow on empirical mixed data. As an example, we use the \texttt{diamonds} dataset which is publicly available on \texttt{OpenML}\footnote{\url{https://www.openml.org/search?type=data&sort=runs&id=42225}} \citep{openml}. Consisting of $9$ covariates ($6$ numerical, $3$ categorical) which relate to characteristics of diamonds such as length, depth and color. We predict the selling price of the diamond in USD (\texttt{price}) using a random forest prediction model. Similar to the experiments in Section~\ref{sec::experiments2}, the importance of the covariates for the prediction model will be determined by CPIseq, CS, LOCO, PFI and SAGE. {  For further details on the dataset and the procedure, as well as a comparison to results given by another prediction model (neural network), see Supplementary Information D.} 

\begin{figure}[h] 
	\centering
	\includegraphics[width=1\textwidth]{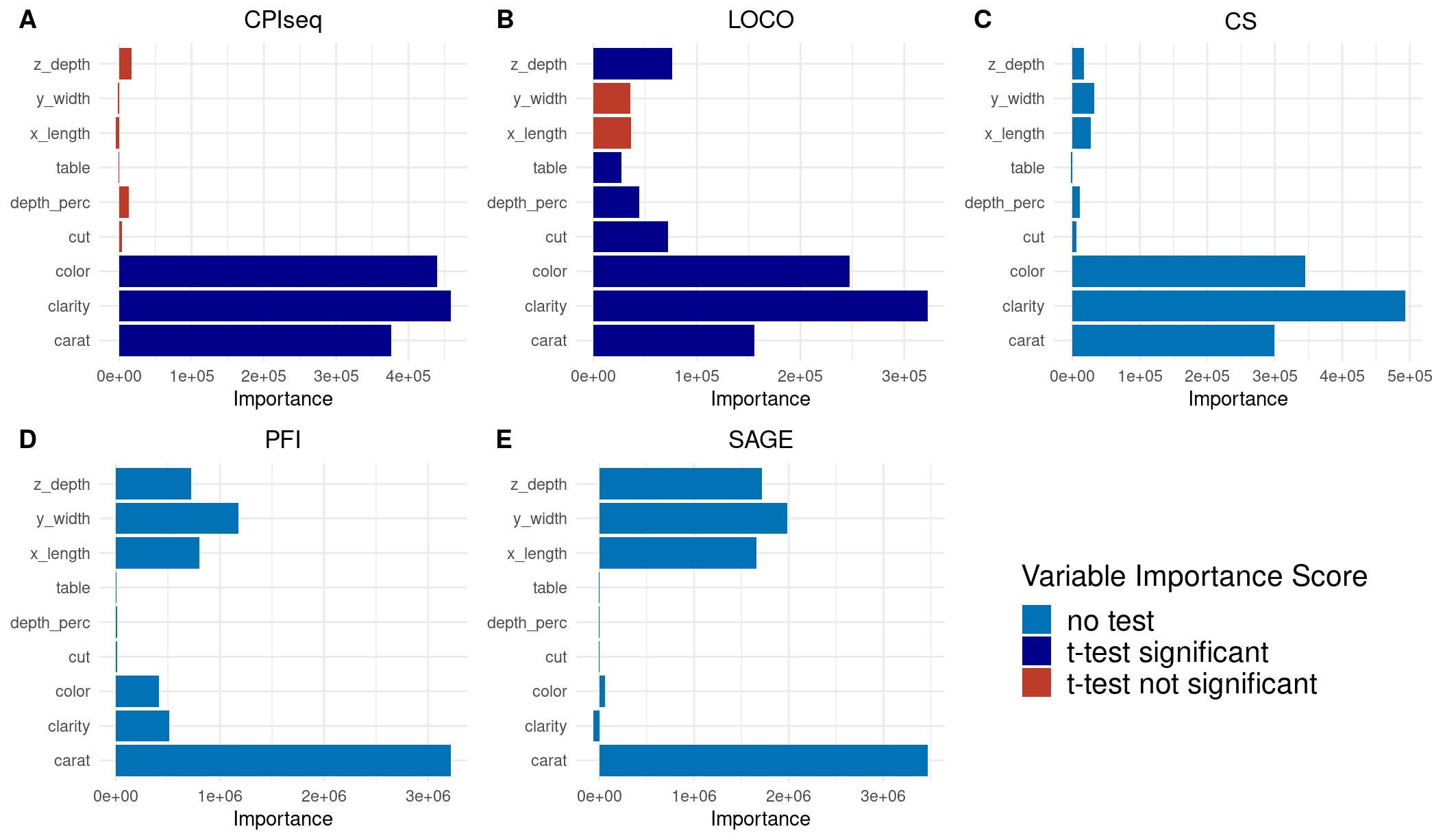}
	\caption{Feature importance scores for predicting the selling price of diamonds using a random forest model. For the CPIseq and LOCO, t-tests are at $\alpha ~= 5\%$, using the Holm procedure to adjust for multiple testing.}
	\label{fig::diamonds}
\end{figure}

Fig.~\ref{fig::diamonds} illustrates the difference between conditional and marginal measures of feature
importance. The marginal measures (Fig.~\ref{fig::diamonds}, Panels~\textbf{D, E}) attribute high importance scores to the covariates \texttt{x\_length, y\_width, z\_depth} and \texttt{carat}, whereas the conditional measures (Fig.~\ref{fig::diamonds}, Panels~\textbf{A, B, C}) attribute high importance scores to the covariates \texttt{color, clarity} and \texttt{carat}. {  Note that the scale of the FI measures in Fig. ~\ref{fig::diamonds} differs, since marginal measures also incorporate the importances of correlated variables and hence, by construction,  exhibit much larger values than conditional FI measures.   }

With some background knowledge on the physical characteristics of diamonds, we can understand the causal relationships that lead to this result. Carat is a measure of weight, and with round diamonds, this weight can be approximated by the formula $\textit{carat} ~= \textit{length} \times \textit{width} \times \textit{depth} \times 0.0061$ \citep{miller1988}. Note that to ensure this formula holds, we only considered diamonds with a deviation $< 0.02$mm from a perfect round shape, yielding a subset of $N ~= 4\,463$ observations. The covariates \texttt{x\_length, y\_width} and \texttt{z\_depth} therefore determine the weight (\texttt{carat}), which all the importance measures suggest as an important predictor variable for \texttt{price}. Conditional FI measures then suggest that \texttt{x\_length, y\_width} and \texttt{z\_depth} do not carry further information on the price, given the other covariates, including \texttt{carat}. Marginal measures, however, attribute importance irrespective of other covariates and hence do not condition on the information given by \texttt{carat}, which leads to high importance values for \texttt{x\_length, y\_width, z\_depth} as well as \texttt{carat}, even though it is reasonable to assume that \texttt{carat} absorbs all relevant information given by \texttt{x\_length, y\_width} and \texttt{z\_depth} on the \texttt{price} of diamonds.

The conditional FI measures further detect the variables \texttt{color} and \texttt{clarity} to be relevant for the prediction of \texttt{price}. Note that we here again have to see this in a conditional sense. Given the other covariates, the variables \texttt{color} and \texttt{clarity} do provide additional information on the \texttt{price}, whereas marginal measures estimate a rather low importance of these variables. 

This real-world example emphasizes the difference between conditional and marginal FI measures and its implications. Again, it is worth repeating that out of the conditional measures, CPIseq facilitates the interpretation through inference procedures providing a clear indication of the relevant variables, whereas this indication is less clear with the LOCO testing procedure and CS not providing the user with testing procedures at all. 

\section{Conclusion and Discussion}
\label{sec::conclusion}
In this work, we highlight the importance of taking statistical considerations into account when measuring FI in interpretable machine learning. Specifically, we focus on conditional versus marginal perspectives on FI measurement, and analyze conditional FI methods with special regard for mixed data. We introduce the combination of CPI and sequential knockoffs (CPIseq) as a strategy { that enables testing of conditional, model-agnostic, global FI with mixed data.} Through simulation studies, we show that CPIseq achieves high power, whereas CPIgauss and CPIdeep are less sensitive for categorical features. Further, we benchmark this method against other conditional FI measures, finding competitive performance, and use a real-world data example to illustrate empirical implications. In sum, we demonstrate that the CPIseq provides researchers with a powerful test for conditional FI while working on a global, model-agnostic level.

Our analyses are limited by the availability of specialized knockoff sampling algorithms for the generation of mixed data knockoffs. Astonishingly, the case of mixed data has not received much attention in the knockoff literature so far and even if some methods were claimed to generalize to the mixed data case \citep{romano2020}, there is a lack of concrete methodology and software implementation. Also, the scarce availability of conditional FI measures that allow for effective statistical testing impedes efficient comparison between FI metrics, forcing the evaluation to rely on rankings. While rankings are oftentimes used in the literature on FI for illustrative purposes, a systematic gold standard for comparing rankings between methods has not emerged. We hypothesize that this might be due to the fact that in the machine learning community, simulation studies -- a standard procedure in the statistics community -- are relatively rare, and hence evaluations involving, e.g., ground truth variable rankings are not in the focus. In particular, with mixed data, a ground truth ranking of simulated variables is not straightforward since it is unclear how the categorical nature should be respected and challenging disagreements across methods are likely to occur \citep{Satyapriya2022}. Methodological development that bridges evaluation strategies commonly applied in statistics with the setting faced in interpretable machine learning, e.g. FI rankings, is highly desirable. 

This work highlights the necessity for procedures that respect data-specific requirements, such as respecting the categorical nature of variables in mixed datasets. Our simulations show that a neglect of such requirements and the application of workarounds might lead to undesirable consequences. We encourage researchers to develop methods that are specifically designed for realistic (mixed) data, instead of leaving practitioners with broad claims of the generalizability of their method. {  While some generalizations are indeed effortless, e.g. for conditional independence testing with all categorical data exact $p$-values can be computed through permutations \citep{tsamardinos2010}, whereas conditional independence testing in general, including mixed data cases, is severely more challenging \citep{shah2020}. Moreover, other data type specific adjustments such as the presence of ordinal data might be of interest for future research, for example, random forest regression models yield the same results with ordinal as with numeric data \citep{hastie2009} and hence FI methods that  exploit model-specific advantages for ordinal data might be proposed.
}

Further, the present work raises awareness of the fact that even though the concept of FI might sound intuitive at first, statistical perspectives on the problem reveal that, for example, the question of marginal in contrast to conditional measurement is of fundamental relevance. We hope this paper elucidates the potential of advancing interpretable machine learning methodology through statistical considerations, which might in turn be mutually beneficial for the future development of the field of explainable artificial intelligence and statistics. 

\subsubsection*{Data and Code Availability}
Data and code to reproduce all the results presented in this paper is available at \url{https://github.com/bips-hb/CFI_mixedData}.

\subsubsection*{Acknowledgements}
MNW and KB received funding for this project from the German Research Foundation (DFG), Emmy Noether Grant 437611051. 

\clearpage

\bibliography{main}

\begin{thebibliography}{41}
\providecommand{\natexlab}[1]{#1}
\providecommand{\url}[1]{{#1}}
\providecommand{\urlprefix}{URL }
\providecommand{\doi}[1]{\url{#1}}
\providecommand{\eprint}[2][]{\url{#2}}
 \bibcommenthead

\bibitem[{Apley and Zhu(2020)}]{apley2020}
Apley D.~W., Zhu J.: {Visualizing the effects of predictor variables in black
  box supervised learning models}. Journal of the Royal Statistical Society
  Series B 82(4), 1059--1086 (2020) \doi{https://doi.org/10.1111/rssb.12377}

\bibitem[{Au et~al.(2022)Au, Herbinger, Stachl, Bischl, and
  Casalicchio}]{au2022}
Au Q., Herbinger J., Stachl C., Bischl B., Casalicchio G.: Grouped feature
  importance and combined features effect plot. Data Mining and Knowledge
  Discovery 36(4), 1401--1450 (2022)
  \doi{https://doi.org/10.1007/s10618-022-00840-5}

\bibitem[{Bates et~al.(2021)Bates, Candès, Janson, and Wang}]{bates2021}
Bates S., Candès E., Janson L., Wang W.: Metropolized knockoff sampling.
  Journal of the American Statistical Association 116(535), 1413--1427 (2021)
  \doi{https://doi.org/10.1080/01621459.2020.1729163}

\bibitem[{Breiman(2001)}]{breiman2001}
Breiman L.: Random forests. Machine learning 45(1), 5--32 (2001)
  \doi{https://doi.org/10.1023/A:1010933404324}

\bibitem[{Cand\`{e}s et~al.(2018)Cand\`{e}s, Fan, Janson, and Lv}]{candes2018}
Cand\`{e}s E., Fan Y., Janson L., Lv J.: Panning for gold: Model-free knockoffs
  for high-dimensional controlled variable selection. Journal of the Royal
  Statistical Society: Series B (Statistical Methodology) 80, 551 -- 577 (2018)
  \doi{https://doi.org/10.1111/rssb.12265}

\bibitem[{Chen et~al.(2020)Chen, Janizek, Lundberg, and Lee}]{chen2020}
Chen H., Janizek J.~D., Lundberg S., Lee S.-I.: True to the model or true to
  the data? ArXiv preprint  (2020)
  \doi{https://doi.org/10.48550/arXiv.2006.16234}

\bibitem[{Covert et~al.(2020)Covert, Lundberg, and Lee}]{covert2020}
Covert I., Lundberg S.~M., Lee S.-I.: Understanding global feature
  contributions with additive importance measures. Advances in Neural
  Information Processing Systems 33, 17\,212--17\,223 (2020)

\bibitem[{Fisher et~al.(2019)Fisher, Rudin, and Dominici}]{fisher2019}
Fisher A., Rudin C., Dominici F.: All models are wrong, but many are useful:
  Learning a variable's importance by studying an entire class of prediction
  models simultaneously. Journal of Machine Learning Research 20(177), 1--81
  (2019)

\bibitem[{Friedman(2001)}]{friedman2001}
Friedman J.~H.: Greedy function approximation: a gradient boosting machine.
  Annals of statistics pp 1189--1232 (2001)

\bibitem[{Glymour et~al.(2019)Glymour, Zhang, and Spirtes}]{glymour2019}
Glymour C., Zhang K., Spirtes P.: Review of causal discovery methods based on
  graphical models. Frontiers in Genetics 10 (2019)
  \doi{https://doi.org/10.3389/fgene.2019.00524}

\bibitem[{Gu and Yin(2021)}]{gu2021}
Gu J., Yin G.: Bayesian knockoff filter using gibbs sampler. ArXiv preprint
  (2021) \doi{https://doi.org/10.48550/arXiv.2102.05223}

\bibitem[{Hastie et~al.(2009)Hastie, Tibshirani, Friedman, and
  Friedman}]{hastie2009}
Hastie T., Tibshirani R., Friedman J.~H., Friedman J.~H.: The elements of
  statistical learning: data mining, inference, and prediction. Vol. 2.
  Springer, New York, NY, USA (2009)

\bibitem[{Hooker et~al.(2021)Hooker, Mentch, and Zhou}]{hooker2021}
Hooker G., Mentch L., Zhou S.: Unrestricted permutation forces extrapolation:{
  V}ariable importance requires at least one more model, or there is no free
  variable importance. Statistics and Computing 31(6), 1--16 (2021)
  \doi{https://doi.org/10.1007/s11222-021-10057-z}

\bibitem[{Jordon et~al.(2019)Jordon, Yoon, and van~der Schaar}]{jordon2019}
Jordon J., Yoon J., van~der Schaar M.: Knockoffgan: Generating knockoffs for
  feature selection using generative adversarial networks. In: International
  Conference on Learning Representations (2019)

\bibitem[{Kormaksson et~al.(2021)Kormaksson, Kelly, Zhu, Haemmerle, Pricop, and
  Ohlssen}]{kormaksson2021}
Kormaksson M., Kelly L.~J., Zhu X., Haemmerle S., Pricop L., Ohlssen D.:
  Sequential knockoffs for continuous and categorical predictors: With
  application to a large psoriatic arthritis clinical trial pool. Statistics in
  Medicine 40(14), 3313--3328 (2021) \doi{https://doi.org/10.1002/sim.8955}

\bibitem[{Krishna et~al.(2022)Krishna, Han, Gu, Pombra, Jabbari, Wu, and
  Lakkaraju}]{Satyapriya2022}
Krishna S., Han T., Gu A., Pombra J., Jabbari S., Wu S., Lakkaraju H.: The
  disagreement problem in explainable machine learning: {A} practitioner's
  perspective. ArXiv preprint  (2022)
  \doi{https://doi.org/10.48550/arXiv.2202.01602}

\bibitem[{Kursa and Rudnicki(2010)}]{kursa2010}
Kursa M.~B., Rudnicki W.~R.: Feature selection with the boruta package. Journal
  of Statistical Software 36(11), 1–13 (2010)
  \doi{https://doi.org/10.18637/jss.v036.i11}

\bibitem[{Lei et~al.(2018)Lei, G’Sell, Rinaldo, Tibshirani, and
  Wasserman}]{lei2018}
Lei J., G’Sell M., Rinaldo A., Tibshirani R.~J., Wasserman L.:
  Distribution-free predictive inference for regression. Journal of the
  American Statistical Association 113(523), 1094--1111 (2018)
  \doi{https://doi.org/10.1080/01621459.2017.1307116}

\bibitem[{Linardatos et~al.(2021)Linardatos, Papastefanopoulos, and
  Kotsiantis}]{linardatos2020}
Linardatos P., Papastefanopoulos V., Kotsiantis S.: Explainable {AI}: A review
  of machine learning interpretability methods. Entropy 23(1) (2021)
  \doi{https://doi.org/10.3390/e23010018}

\bibitem[{Liu and Zheng(2018)}]{liu2018}
Liu Y., Zheng C.: Auto-encoding knockoff generator for {FDR} controlled
  variable selection. ArXiv preprint  (2018)
  \doi{https://doi.org/10.48550/ARXIV.1809.10765}

\bibitem[{Lundberg and Lee(2017)}]{lundberg2017}
Lundberg S.~M., Lee S.-I.: A unified approach to interpreting model
  predictions. Advances in neural information processing systems 30 (2017)

\bibitem[{Lundberg et~al.(2020)Lundberg, Erion, Chen, DeGrave, Prutkin, Nair,
  Katz, Himmelfarb, Bansal, and Lee}]{lundberg2020}
Lundberg S.~M., Erion G., Chen H., DeGrave A., Prutkin J.~M., Nair B., Katz R.,
  Himmelfarb J., Bansal N., Lee S.-I.: From local explanations to global
  understanding with explainable {AI} for trees. Nat Mach Intell 2(1), 56--67
  (2020) \doi{https://doi.org/10.1038/s42256-019-0138-9}

\bibitem[{Miller(1988)}]{miller1988}
Miller A.~M.: Jewelry from Antiquity to the Modern Era. In: Gems and Jewelry
  Appraising. Springer, Boston, MA (1988)
  \doi{https://doi.org/10.1007/978-1-4684-1404-2_5}

\bibitem[{Molnar et~al.(2023)Molnar, K{\"o}nig, Bischl, and
  Casalicchio}]{molnar2023}
Molnar C., K{\"o}nig G., Bischl B., Casalicchio G.: Model-agnostic feature
  importance and effects with dependent features: A conditional subgroup
  approach. Data Mining and Knowledge Discovery pp 1--39 (2023)
  \doi{https://doi.org/10.1007/s10618-022-00901-9}

\bibitem[{Pearl(2009)}]{pearl2009}
Pearl J.: Causality. Cambridge University Press, Cambridge, UK (2009)
  \doi{https://doi.org/10.1017/CBO9780511803161}

\bibitem[{Ribeiro et~al.(2016)Ribeiro, Singh, and Guestrin}]{ribeiro2016}
Ribeiro M.~T., Singh S., Guestrin C.: "{W}hy should {I} trust you?"
  {E}xplaining the predictions of any classifier. In: Proceedings of the 22nd
  ACM SIGKDD International Conference on Knowledge Discovery and Data Mining
  (2016) \doi{https://doi.org/10.1145/2939672.2939778}

\bibitem[{Ribeiro et~al.(2018)Ribeiro, Singh, and Guestrin}]{ribeiro2018}
Ribeiro M.~T., Singh S., Guestrin C.: Anchors: High-precision model-agnostic
  explanations. In: Proceedings of the AAAI conference on artificial
  intelligence (2018)

\bibitem[{Rinaldo et~al.(2016)Rinaldo, Wasserman, G'Sell, and
  Lei}]{rinaldo2016}
Rinaldo A., Wasserman L., G'Sell M., Lei J.: Bootstrapping and sample splitting
  for high-dimensional, assumption-free inference (2016)
  \doi{https://doi.org/10.48550/ARXIV.1611.05401}

\bibitem[{Romano et~al.(2020)Romano, Sesia, and Cand{\`e}s}]{romano2020}
Romano Y., Sesia M., Cand{\`e}s E.: Deep knockoffs. Journal of the American
  Statistical Association 115(532), 1861--1872 (2020)
  \doi{https://doi.org/10.1080/01621459.2019.1660174}

\bibitem[{Sesia et~al.(2018)Sesia, Sabatti, and Candès}]{seisa2018}
Sesia M., Sabatti C., Candès E.~J.: {Gene hunting with hidden Markov model
  knockoffs}. Biometrika 106(1), 1--18 (2018)
  \doi{https://doi.org/10.1093/biomet/asy033}

\bibitem[{Shah and Peters(2020)}]{shah2020}
Shah R.~D., Peters J.: The hardness of conditional independence testing and the
  generalised covariance measure. The Annals of Statistics 48(3), 1514--1538
  (2020) \doi{https://doi.org/10.1214/19-AOS1857}

\bibitem[{Shapley(1953)}]{shapley1953}
Shapley L.: A Value for n-Person Games. In: Kuhn, H. and Tucker, A., Eds.,
  Contributions to the Theory of Games II. Princeton University Press,
  Princeton, NJ (1953) \doi{https://doi.org/10.1515/9781400881970-018}

\bibitem[{Shrikumar et~al.(2017)Shrikumar, Greenside, and
  Kundaje}]{shrikumar2017}
Shrikumar A., Greenside P., Kundaje A.: Learning important features through
  propagating activation differences. In: International conference on machine
  learning. PMLR (2017)

\bibitem[{Sudarshan et~al.(2020)Sudarshan, Tansey, and
  Ranganath}]{sudarshan2020}
Sudarshan M., Tansey W., Ranganath R.: Deep direct likelihood knockoffs.
  Advances in neural information processing systems 33 (2020)

\bibitem[{Tsamardinos and Borboudakis(2010)}]{tsamardinos2010}
Tsamardinos I., Borboudakis G.: Permutation testing improves bayesian network
  learning. In: Machine Learning and Knowledge Discovery in Databases: European
  Conference, ECML PKDD (2010)
  \doi{https://doi.org/10.1007/978-3-642-15939-8_21}

\bibitem[{Vanschoren et~al.(2014)Vanschoren, Van~Rijn, Bischl, and
  Torgo}]{openml}
Vanschoren J., Van~Rijn J.~N., Bischl B., Torgo L.: Openml: networked science
  in machine learning. ACM SIGKDD Explorations Newsletter 15(2), 49--60 (2014)

\bibitem[{Watson and Wright(2021)}]{watson2021}
Watson D.~S., Wright M.~N.: Testing conditional independence in supervised
  learning algorithms. Machine Learning 110(8), 2107--2129 (2021)
  \doi{https://doi.org/10.1007/s10994-021-06030-6}

\bibitem[{Watson et~al.(2023)Watson, Blesch, Kapar, and Wright}]{watson2022}
Watson D.~S., Blesch K., Kapar J., Wright M.~N.: Adversarial random forests for
  density estimation and generative modeling. In: Proceedings of the 23rd
  International Conference on Artificial Intelligence and Statistics (2023)

\bibitem[{Williamson et~al.(2021)Williamson, Gilbert, Carone, and
  Simon}]{williamson2021}
Williamson B.~D., Gilbert P.~B., Carone M., Simon N.: Nonparametric variable
  importance assessment using machine learning techniques. Biometrics 77(1),
  9--22 (2021) \doi{https://doi.org/10.1111/biom. 13392}

\bibitem[{Xu et~al.(2019)Xu, Skoularidou, Cuesta-Infante, and
  Veeramachaneni}]{xu2019}
Xu L., Skoularidou M., Cuesta-Infante A., Veeramachaneni K.: Modeling tabular
  data using conditional {GAN}. Advances in Neural Information Processing
  Systems 32 (2019)

\bibitem[{Zou and Hastie(2005)}]{zou2005}
Zou H., Hastie T.: Regularization and variable selection via the elastic net.
  Journal of the royal statistical society: Series B (Statistical Methodology)
  67(2), 301--320 (2005) \doi{https://doi.org/10.1111/j.1467-9868.2005.00503.x}

\end{thebibliography}
\newpage
\appendix
\section{Conditional Independent Pairs Algorithm} \label{app::cond_indep_pairs}
\begin{algorithm}
\caption{Conditional Independent Pairs (Candès et al., 2018)}\label{alg::independent_pairs}
\begin{algorithmic}
\State $j = 1$
\While{ $j \leq p$} 
\State sample $\Tilde{X_j} \sim \mathcal{L}(X_j \mid X_{-j}, \Tilde{X}_{i:j-1}$) 
\State $j = j+1$
\EndWhile
\end{algorithmic}
\end{algorithm}
\section{Simulations in Section 3.1}
\subsection{Experimental Setup}
 For the experiments in Section 3.1,
 we consider a data generating process that allows us to understand the effect of mixed data in conditional feature importance (FI) measurement for a known causal structure. In detail, we generate four variables using the following equations that detail the data generating process:
\begin{align*}
    X_1 &\sim N(0,1) \\
    X_2 &\sim N(0,1) \\
    X_3 &= \beta X_2 + \epsilon_{X_3} \\
    X_4 &=\beta X_1 + \epsilon_{X_4} \\
    & \hspace{3cm}\text{where } \epsilon_{X_3},\epsilon_{X_4} \sim N(0,1)
\end{align*} \label{eq::dgp_sim1}
\noindent The target variable $Y$ is either continuous (regression task) or binary (classification task). We introduce these two scenarios to illustrate that for a given number of observations, the type of target variable has an influence on the FI values, i.e. the power to detect them, through the model fit. However, since this paper analyzes the importance measurement of mixed variables $X_j$, $j \in \{1,2,3,4\}$ and not on $Y$, we evaluate the difference between regression and classification task only for an exemplary case (part of scenario III, with $X_3$ as 10 level categorical variable). The target variable for the respective tasks will be generated according to:
\begin{align*}
        \text{regression task: }Y  &= \beta X_3 +\beta X_4 + \epsilon_{Y} \\
        \text{classification task: }Y  &\sim Bern(\text{logit}^{-1}(\beta X_3 -\beta X_4)) \\
        & \text{where } \epsilon_{Y} \sim N(0,1)
\end{align*}
{  
\noindent This data structure ensures a controlled setting in which all variables are relevant in a marginal sense, but only some are relevant in a conditional sense of FI measurement. For example, $X_1$ affects the target variable $Y$ only through $X_4$. Hence, a conditional FI measure should not attribute any importance to $X_1$ in the prediction of $Y$ given $X_4$.

For variation in effect size $\beta$, we analyze two different settings, namely moderate effect size $\beta =~ 0.5$ and high effect size $\beta =~ 0.9$ to investigate the increase in power when the variables have a strong effect on $Y$.

To illustrate the different ways mixed data can occur in real-world applications, we introduce low-, medium- and high-cardinality categoricals. Scenario II uses binary variables (low-cardinality case) and scenario III investigated 4-level (medium-cardinality case) and 10-level categoricals (high-cardinality case). Categoricals will be generated using the procedure given in \ref{app:generate_cat}.

Another condition we vary is the type of model used for the prediction task. While CPI has been shown to be model-agnostic, i.e. work with any machine learning prediction model, we here exemplify two models to illustrate this point. In detail, we compare the generalized linear model and a random forest model, see further Section \ref{subsec::hyperparameters}.

\noindent In sum, we fit the following configurations:
\begin{enumerate}
    \item Scenario I: All covariates Gaussian, regression outcome, varying moderate and high effect size, fitting both linear and random forest models.  
    \item Scenario II: Either $X_1$ or $X_3$ binary, regression outcome, varying moderate and high effect size, fitting both regression and random forest models.  As a point of reference, we include the case with all covariates are binary, varying moderate and high effect size, fitting both linear and random forest models.
    \item Scenatio III: 
    \begin{itemize}
        \item Medium-cardinality: Either $X_1$ or $X_3$ 4-level categoricals, regression outcome, varying moderate and high effect size, fitting both  and random forest models. 
        As a point of reference, we include the case with all covariates are 4-level categoricals, varying moderate and high effect size, fitting both linear and random forest models
        \item High-cardinality (A): $X_3$ 10-level categorical, regression or classification outcome, high effect size, fitting both linear and random forest models.
        \item High-cardinality (B): Both $X_1$ and $X_3$ 10-level categoricals, regression outcome, moderate effect size, random forest prediction model.
    \end{itemize}
\end{enumerate}

}
\subsection{Generating Categorical Variables}\label{app:generate_cat}
\begin{figure}[h] 
	\centering
	\includegraphics[width=1\textwidth]{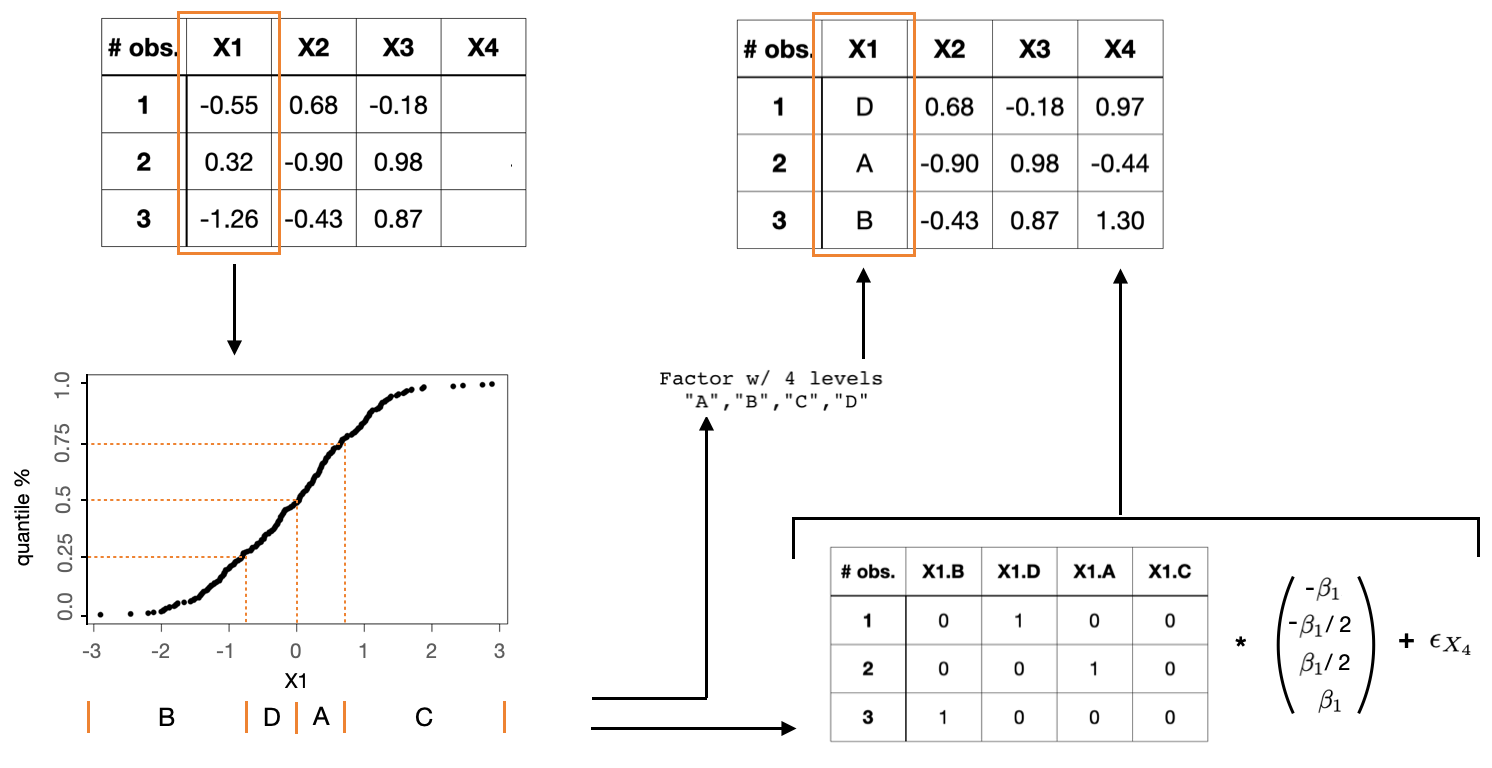}
	\caption{Transforming continuous variables into categorical variables.}
	\label{fig::dgp_cat}
\end{figure}

Synthesizing data sets that contain both continuous and categorical variables with a predetermined correlational structure is a challenging task. In brief, we tackle mixed data generation by defining the categorical variables based on quantiles of continuous variables, see Fig. \ref{fig::dgp_cat} for an intuitive visualization of the procedure.

First, the variable will be generated as a continuous variable according to the formulas given in the data generating process outlined in Section \ref{eq::dgp_sim1}. Then, the respective variable will be cut into quantiles, where the number of quantiles equals the number of desired category levels. Randomly chosen letters will serve as cateogory labels that are assigned to the variable depending on the quantile.  Downstream the causal path (Fig. \ref{fig::sim1}), 
we ensure that the quantile-wise categorized variables are treated as unordered factors. Internally, the categorical variables will be represented as one-hot encoded variables, and each one-hot variable, i.e. category level, will be assigned an effect  $\beta$. These effects expand equidistant from $-\beta$ to $+\beta$ and will be multiplied with the dummy encoded dataset. This results in a numerical representation of the categorical variable that then can be used as an input value to proceed down the causal path. 

Note that the effects of the levels in a categorical variable sum to zero. This means only a procedure that respects the categorical nature of the data, i.e. treats each level individually, will detect the importance of a categorical variable. 

\subsection{Prediction Models} \label{subsec::hyperparameters}
{  The following prediction models are used for the simulation: 
\begin{itemize}
    \item Generalized linear model: Ordinary least squares linear regression model for regression tasks; Logistic regression for classification tasks.
    \item Random Forest: We use the defaults of \texttt{R}-package \texttt{ranger}\footnote{\url{https://cran.r-project.org/web/packages/ranger/ranger.pdf}}, i.e. $500$ trees in the forest.
\end{itemize}
The prediction models are trained on $2/3$ of the data (training data). The remaining $1/3$ of the data (test data) is used to evaluate the models and calculate the respective FI scores. 

Knockoff sampling procedures are the sequential, Gaussian and deep knockoff sampler, as introduced in the main text. The knockoff procedures have access to the test data only and it is worth pointing out that the Gaussian knockoff and Deep knockoff subroutines require dummy encoded data, which significantly expands the dimension of the data matrix if medium- or high-cardinality features are included. In these cases, the prediction models will be defined on the dummy encoded data and CPI evaluated for the variables' respective group of dummy variables.  } 
\subsection{Model Validation}
{  We investigate the model fit of the prediction models used for the experiments of Section 3.1 since a good model fit is substantial for the CPI procedure. We fit the  models on the training data and evaluate the performance on the test data, as it is done within the CPI procedure.\footnote{Note that we also account for the fact that Gaussian and deep knockoffs use dummy encoded data whereas sequential knockoffs can use categorical data directly.} 

For regression tasks, we use $\text{R}^2$ as a measure of model performance and will compare the respective models' $\text{R}^2$ to $\text{R}^{2*}$, which is the optimal $\text{R}^2$ that can be achieved for the dataset. To calculate $\text{R}^{2*}$, we have to determine the signal-to-noise-ratio (SNR) that can be defined as $SNR := \frac{Var(\mathbf{X}\beta)}{Var(\epsilon_Y)}$. From the  data generating process (DGP), we know that $Y = \beta X_3 + \beta X_4 + \epsilon_Y$, where $\epsilon_Y \overset{\mathrm{iid}}{\sim} N(0,1)$. Hence, we have $Var(\epsilon_Y) = 1$ and can calculate $Var(\mathbf{X}\beta) = Var(Y) - Var(\epsilon_Y) =  Var(Y) -1$ from the dataset since we have access to $Y$. This enables us to calculate $SNR = \frac{Var(Y) -1}{1}$ for each dataset we use in Section 3.1 individually. The optimal $\text{R}^{2*}$ is then given by $\text{R}^{*2} = \frac{\textbf{SNR}}{\textbf{SNR} +1}$. The quantity that allows us to study how close to optimality the prediction models are across the dataset configurations we use for the experiments is the discrepancy, i.e. $\text{R}^{2}- \text{R}^{*2}$, which we present in Figure \ref{fig::validata_models_3.1}. 

For classification tasks, we use accuracy as a measure of model performance. Similar to the setup for regression tasks, we again compare the empirically achieved accuracy of the model with the optimal accuracy that can be achieved for each of the datasets used for the experiments in Section 3.1. The DGP for classification tasks generates the target variable as $Y  \sim Bern(p = \text{logit}^{-1}(\beta X_3 -\beta X_4))$. The value of $p$ reveals how accurate an optimal prediction model can perform on the prediction of the binary variable $Y$, since it takes on a value of $1$ with probability $p$ and a value of $0$ with probability $1-p$. That is, the optimal accuracy achievable is given by max($p$, 1-$p$).  We calculate the difference between the empirical accuracy of the model on the test set and the optimal accuracy we can extract when generating each of the datasets during the simulation. The discrepancy across dataset configurations\footnote{Note that we exclude sample size $N=100$ from the Figure because of its high variance (test set consists of only $N = 100*0.33 = 33$ observations).} is given in Figure \ref{fig::validata_models_3.1}.

Overall, the deviations from optimality are small and become more centered around 0 with less data in the tails for larger sample sizes. We therefore conclude that the models have a sufficiently good fit to the data to reliably run the CPI procedure. 

  \begin{figure}[!htbp]
	\centering
	\includegraphics[trim={0 0 0 0cm},clip,width=1\textwidth]{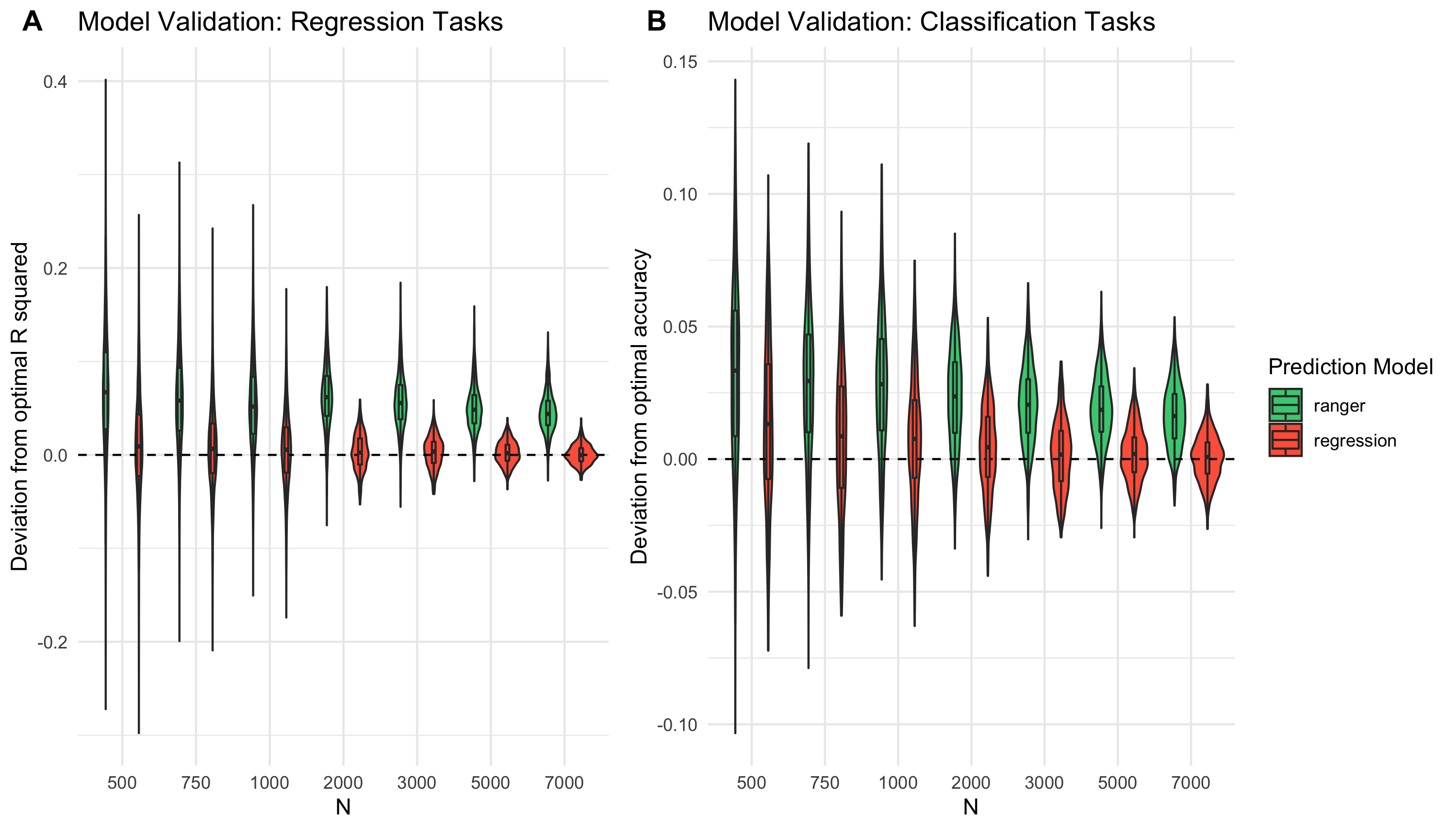}
	\caption{{  Violin- and boxplots of deviation from optimal $\text{R}^{2*}$ (Panel \textbf{A}) and from optimal accuracy (Panel \textbf{B}) for least-squares or logistic regression models and random forest (ranger) models across datasets generated in the simulations of Section 3.1.}}
 \label{fig::validata_models_3.1}
\end{figure}

\clearpage
\subsection{Results}
\subsection*{Scenario (I)}
\begin{figure}[ht!]
\includegraphics[trim={0 0 0 2cm},clip, width=1.1\textwidth]{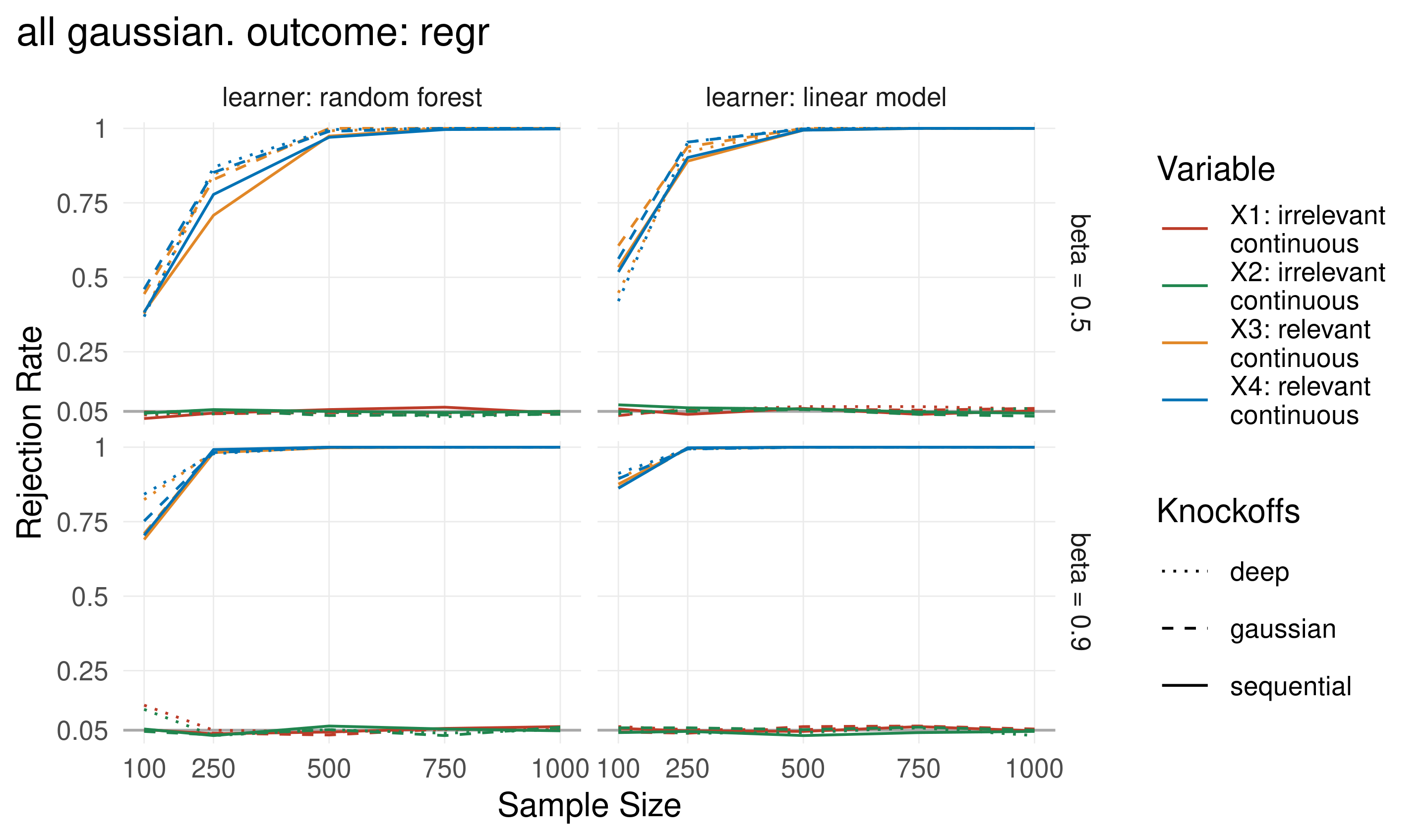}
\centering
\caption{All variables continuous, $Y$ continuous.{  Variation in effect size (beta) and prediction model (random forest, linear model).}}\label{fig::all_gaussian}       
\end{figure} 
\newpage
\FloatBarrier
\subsection*{Scenario (II)}
\begin{figure}[ht!]
\includegraphics[trim={0 0 0 2cm},clip, width=1.1\textwidth]{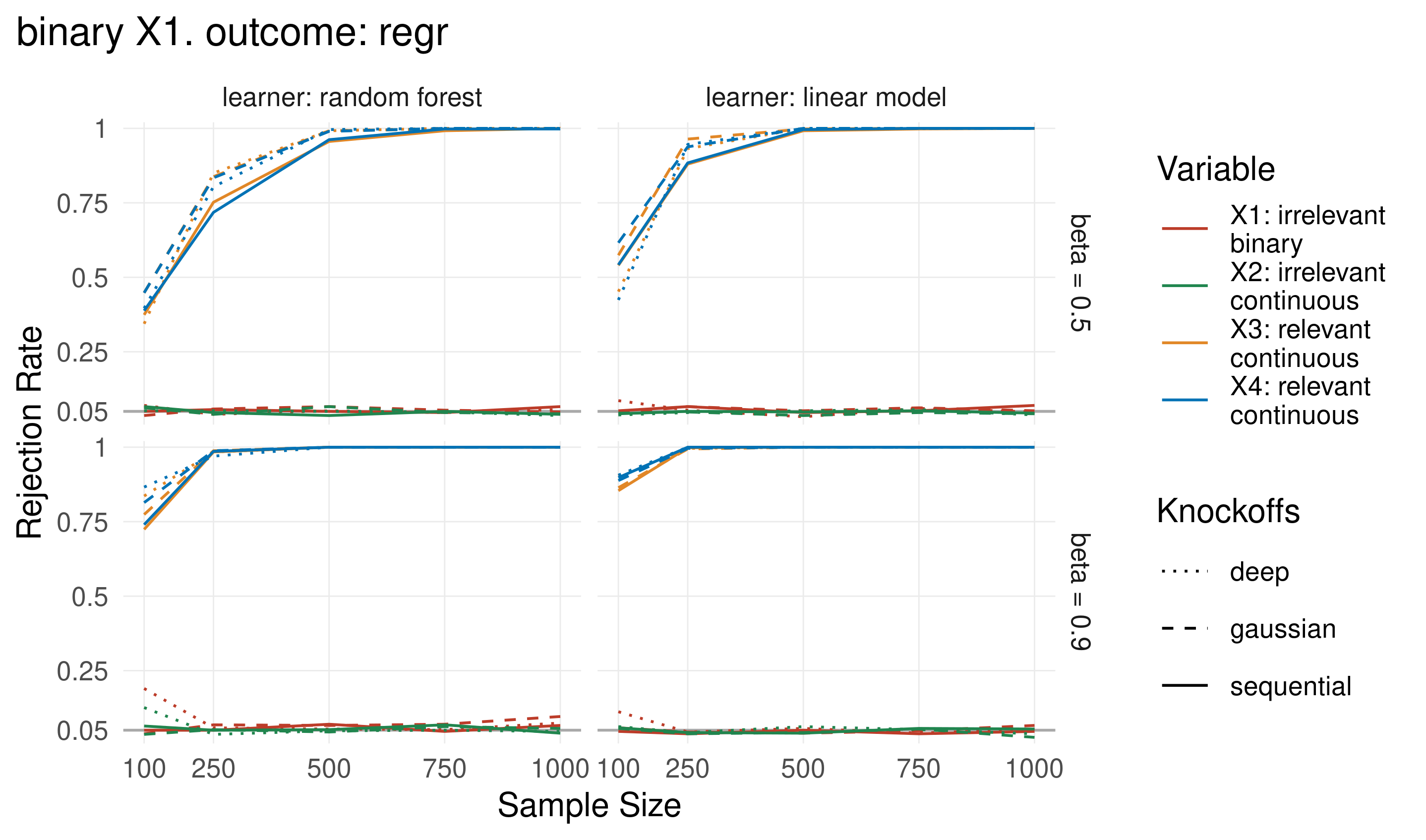}
\centering
\caption{X1 binary, $Y$ continuous.{  Variation in effect size (beta) and prediction model (random forest, linear model).}}\label{fig::X1_binary}       
\end{figure} 

\begin{figure}[ht!]
\includegraphics[trim={0 0 0 2cm},clip, width=1.1\textwidth]{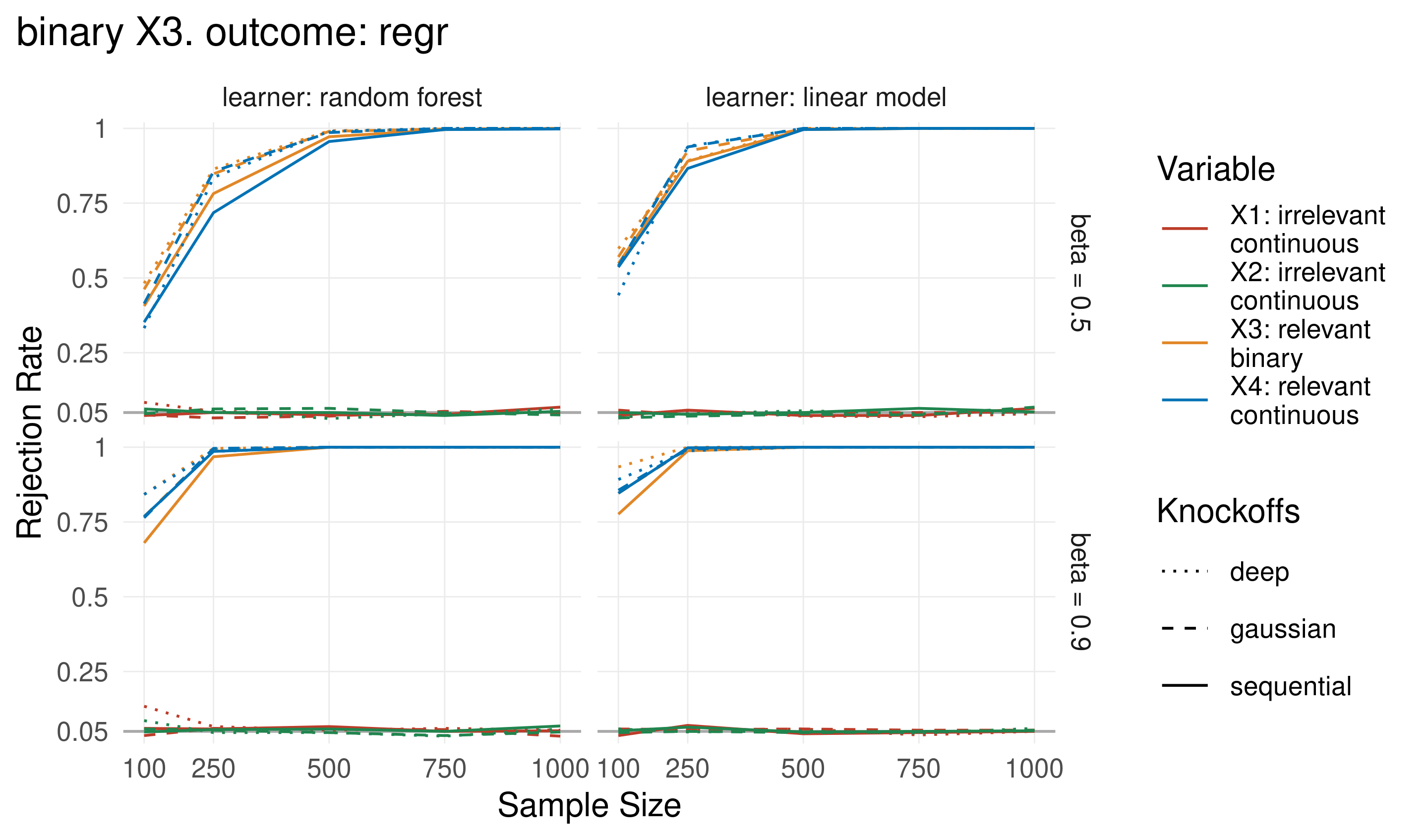}
\centering
\caption{X3 binary, $Y$ continuous.{  Variation in effect size (beta) and prediction model (random forest, linear model).}}\label{fig::X3_binary}       
\end{figure} 

\begin{figure}[ht!]
\includegraphics[trim={0 0 0 2cm},clip, width=1.1\textwidth]{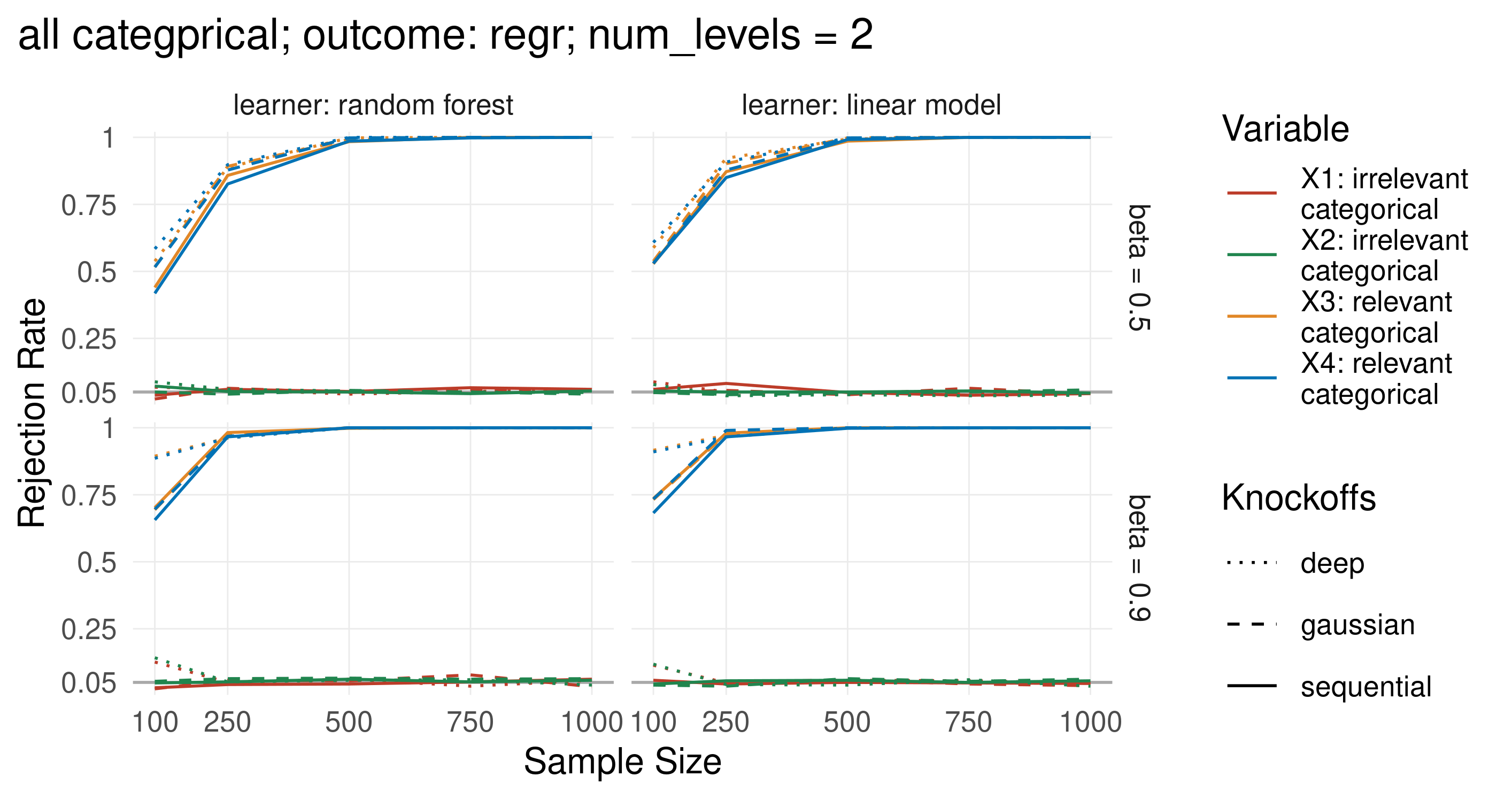}
\centering
\caption{All variables binary, $Y$ continuous. {  Variation in effect size (beta) and prediction model (random forest, linear model).}}\label{fig::binary_all_cat}       
\end{figure} 
\FloatBarrier
\subsection*{Scenario (III)}
\begin{figure}[ht!]
\includegraphics[trim={0 0 0 2cm},clip, width=1.1\textwidth]{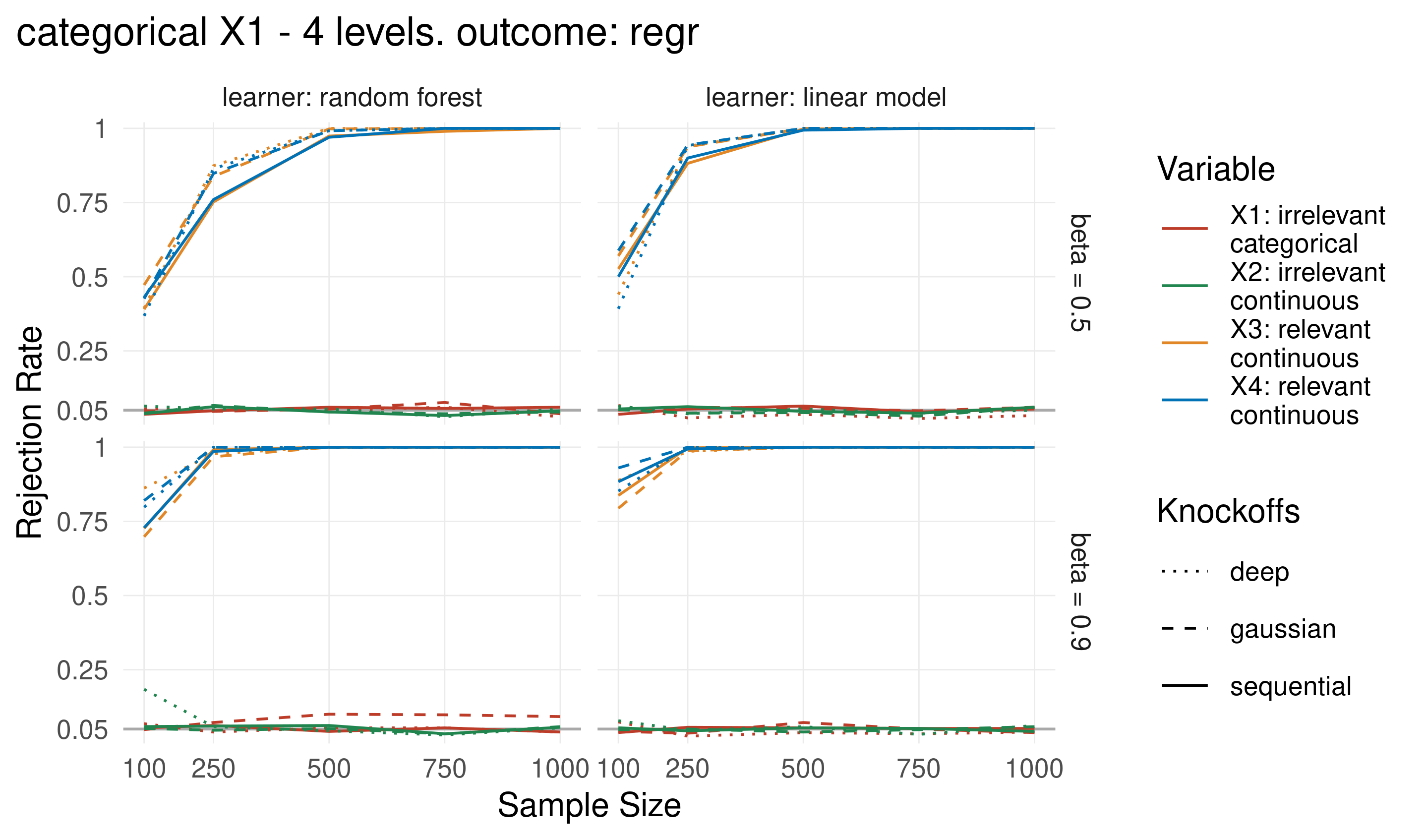}
\centering
\caption{X1 categorical with $4$ levels, $Y$ continuous.{  Variation in effect size (beta) and prediction model (random forest, linear model).}}\label{fig::X1_cat}       
\end{figure} 

\begin{figure}[ht!]
\includegraphics[trim={0 0 0 2cm},clip, width=1.1\textwidth]{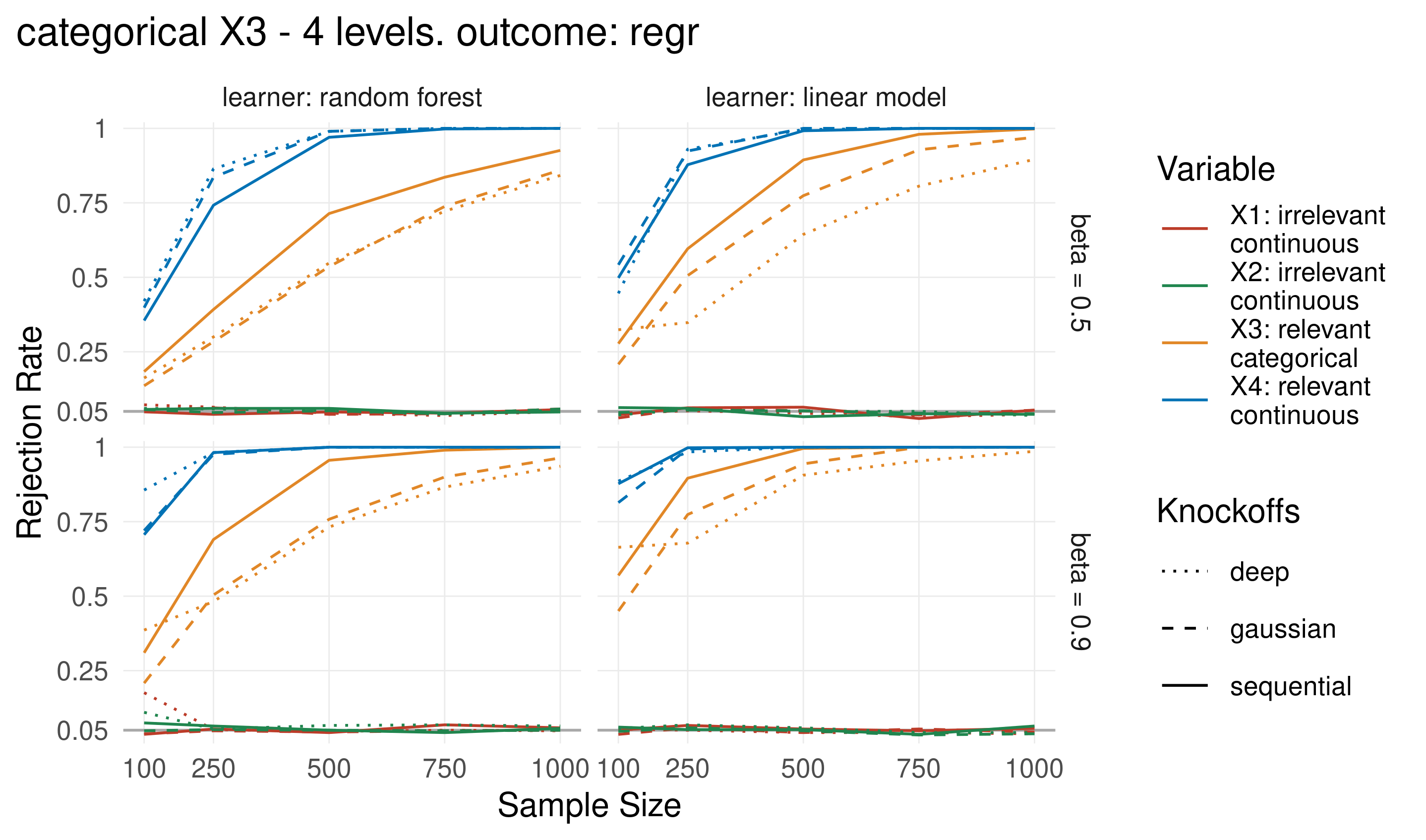}
\centering
\caption{X3 categorical with $4$ levels, $Y$  continuous.{  Variation in effect size (beta) and prediction model (random forest, linear model).}}\label{fig::X3_cat_4}       
\end{figure} 

\begin{figure}[ht!]
\includegraphics[trim={0 0 0 2cm},clip, width=1.1\textwidth]{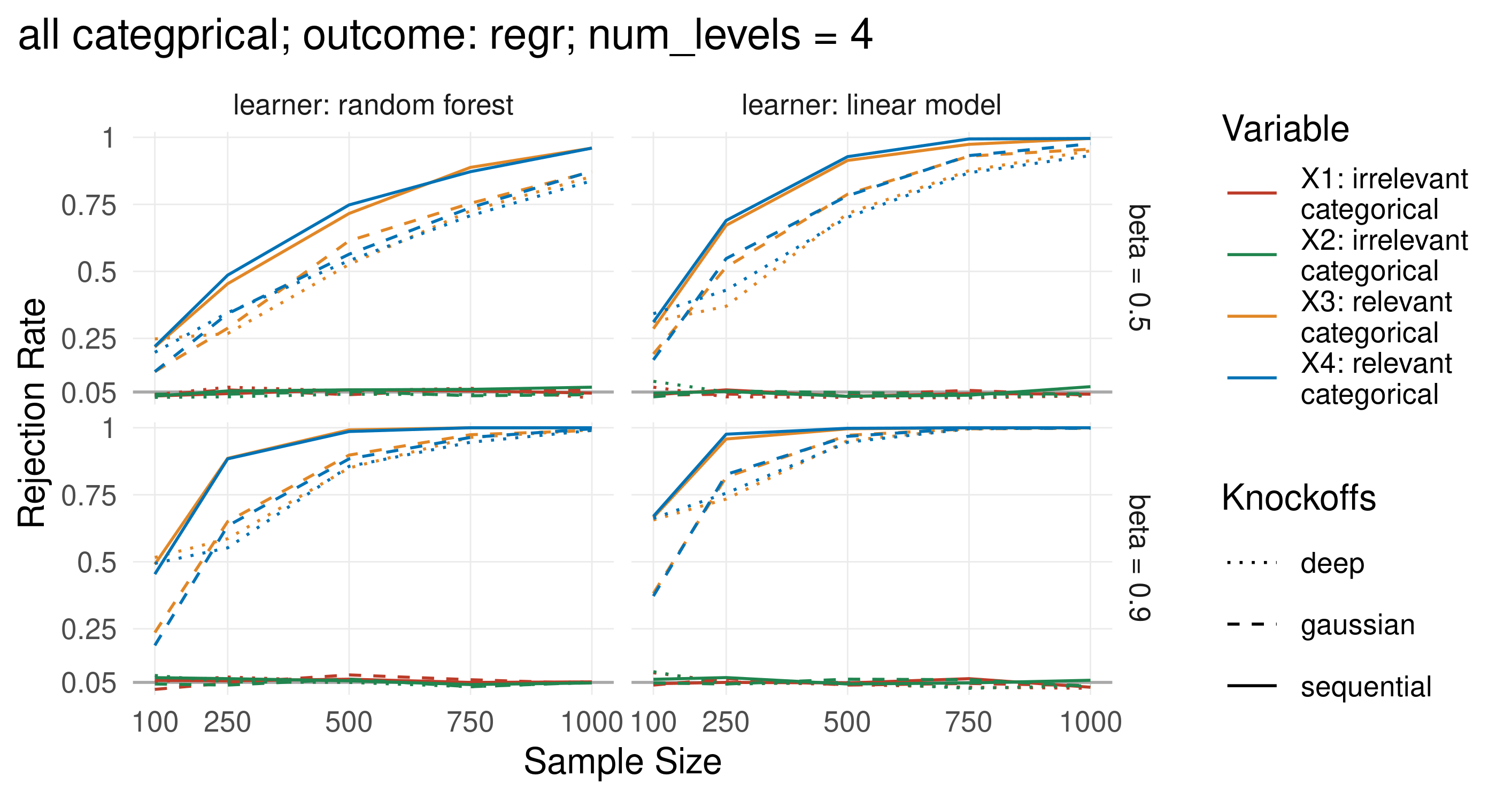}
\centering
\caption{All variables categorical with $4$ levels, $Y$ continuous. {  Variation in effect size (beta) and prediction model (random forest, linear model).}}\label{fig::cat_all_cat}       
\end{figure} 
\clearpage
\begin{figure}[t!]
\includegraphics[trim={0 0 0 1.19cm},clip, width=1.1\textwidth]{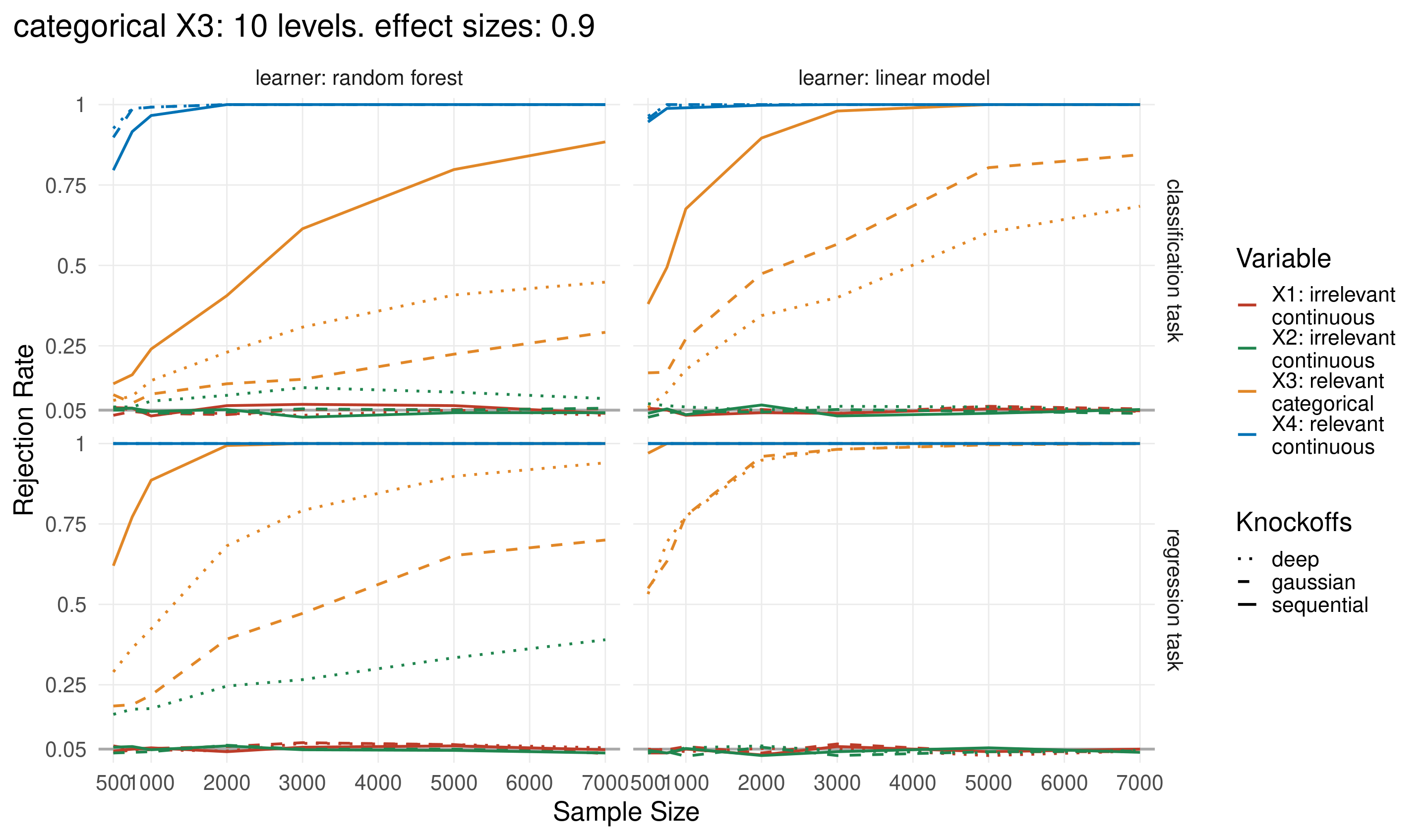}
\centering
\caption{X3 categorical with $10$ levels. Effect size beta fixed at 0.9.{   Variation type of target variable (classification or regression task) and prediction model (random forest, linear model).}}\label{fig::X3_cat_10}       
\end{figure} 
}
\section{Simulations in Section 3.2}
\subsection{Experimental Setup}\label{app::experiments_2}
For the experiments in Section 3.2 
, the data generating process is based on sampling variables from multivariate normal distribution with a predefined correlation structure. We generate data $\mathbf{X}^{n \times 12} \sim N(\mu, \mathbf{\Sigma})$, where the covariance matrix is such that exactly one relevant and irrelevant variable of the same type (Table \ref{tab::dgp_sim2}), are correlated by $\rho$, i.e.
\begin{align*}
    \mu &= \begin{bmatrix}
           \mu_{1} \\
           \mu_{2} \\
           \vdots \\
           \mu_{12}
         \end{bmatrix} = \begin{bmatrix} 
           0 \\
           0 \\
           \vdots \\
           0
         \end{bmatrix} \text{ \hspace{1.5cm} }
    \Sigma = \begin{bmatrix}
    1 & \rho & 0 & 0 & \hdots & 0 & 0 & 0 & 0\\
    \rho & 1 & 0 & 0 & \hdots & 0 & 0 & 0 & 0\\
    0 & 0 & 1 & \rho & \hdots & 0 & 0 & 0 & 0\\
    0 & 0 & \rho & 1 & \hdots & 0 & 0 & 0 & 0\\
    \vdots & \vdots &  \vdots & \vdots  &  \ddots & \vdots & \vdots  & \vdots & \vdots \\
    0 & 0 & 0 & 0 & \hdots & 1 & \rho  & 0 & 0\\
    0 & 0 & 0 & 0 & \hdots & \rho & 1  & 0 & 0\\
    0 & 0 & 0 & 0 & \hdots & 0 & 0 & 1 & \rho \\
    0 & 0 & 0 & 0 & \hdots & 0 & 0 & \rho & 1 \\
    \end{bmatrix}
\end{align*}

\begin{table}[h]
\centering
        \begin{tabular}{lcclll}
            \toprule
            &  Type &Effect on $Y$: &Effect on $Y$:&\\
            &  & Transformation & Strength  & &   \\ \midrule
        $X_1$&     &  & no effect & \multicolumn{1}{l}{$\beta =$ 0} \\\cline{4-5}
        $X_2$&    continuous                      & no transformation& weak effect & \multicolumn{1}{l}{$\beta =$ 1} \\\cmidrule{4-5}
        $X_3$&       linear                  && no effect & \multicolumn{1}{l}{$\beta =$ 0} \\\cline{4-5}
        $X_4$&                         & & strong effect & \multicolumn{1}{l}{$\beta =$ 3} \\\cmidrule{2-5}
        $X_5$&                       &   & no effect & \multicolumn{1}{l}{$\beta =$ 0} \\\cmidrule{4-5}
        $X_6$&       continuous                  &\multirow{3}{*}{ $x_{ij}= $
       $\begin{cases}
    +1, & \text{if } \Phi^{-1}(0.25) \leq x_{ij} \leq \Phi^{-1}(0.75)\\
    -1,              & \text{otherwise} \end{cases}$
}& weak effect & \multicolumn{1}{l}{$\beta =$ 1} \\\cmidrule{4-5}
        $X_7$&     nonlinear                    && no effect & \multicolumn{1}{l}{$\beta =$ 0} \\\cmidrule{4-5}
        $X_8$&                         && strong effect & \multicolumn{1}{l}{$\beta =$ 3} \\\cmidrule{2-5}
        $X_9$&   & & no effect & \multicolumn{1}{l}{$\beta =$ 0} \\\cmidrule{4-5}
        $X_{10}$&           categorical              &see Appendix \ref{app:generate_cat} & weak effect & \multicolumn{1}{l}{$\beta =$ 1} \\\cmidrule{4-5}
        $X_{11}$&                         && no effect & \multicolumn{1}{l}{$\beta =$ 0} \\\cmidrule{4-5}
        $X_{12}$&                         & & strong effect & \multicolumn{1}{l}{$\beta =$ 3} \\\bottomrule
        \end{tabular}
    \caption{Variable Types}
    \label{tab::dgp_sim2}
\end{table}

The target variable $Y$ again can be either continuous (regression task) or binary (classification task).
\begin{align*}
        \text{regression task: }Y  &= \sum_{j=1}^{p}\beta_j X_j + \epsilon_{R} \\
        \text{classification task: }Y  &\sim Bern(\text{logit}^{-1}(\beta_{BER} \cdot\sum_{j=1}^{p}\beta_j X_j) ) \\
        & \text{where } \epsilon_{R} \text{ is such that the signal-to-noise ratio } \text{SNR} = 2\\
        & \text{and } \beta_{BER} \text{ is such that the Bayes error rate } \text{BER} = 0.2
\end{align*}
The SNR is achieved by sampling $\epsilon_C$ from a normal distribution $N(0, \sigma^2)$, where $\sigma$ is chosen such that the desired level of $SNR = var(\mathbf{X \beta})/\sigma^2 = 2$ is achieved. For classifcation tasks, the BER is achieved through multiplication of $\sum_{j=1}^{p}\beta_j X_j) )$ with $\beta_{BER}$, where $\beta_{BER}$ is determined through minimizing the difference between achieved and empirical Bayesian error rate. 
{  
Finally, let us summarize the parameter configurations considered for the experiments in Section 3.2. We calculate the results for all possible combinations (24 in total) of the following parameters:
\begin{itemize}
    \item Correlation strength between covariates: $\rho = 0.5$ or $\rho = 0.8$
    \item Cardinality of categorical variables: 2 or 5
    \item Target variable $Y$: binary (classification task) or continuous (regression task)
    \item Prediction model (learner): regression, random forest or neural network
    
\end{itemize} }
\subsection{Prediction Models}
{  The following prediction models are used for the simulation: }
\begin{itemize}
    \item Generalized linear model: Ordinary least squares linear regression model for regression tasks; Logistic regression for classification tasks.
    \item Random Forest: We use the defaults of \texttt{R}-package \texttt{ranger}\footnote{\url{https://cran.r-project.org/web/packages/ranger/ranger.pdf}}, i.e. $500$ trees in the forest.
    \item Neural Network: We use a single hidden layer network as given by the the defaults of \texttt{R}-package \texttt{nnet}\footnote{\url{https://cran.r-project.org/web/packages/nnet/nnet.pdf}} with the modifications to the size of the hidden layers (number of units in the hidden layer is set to 20) and a weight decay of $0.1$.
\end{itemize}
{  The prediction models are trained on $2/3$ of the data (training data). The remaining $1/3$ of the data (test data) is used to evaluate the models and calculate the respective FI scores. 

Knockoff sampling procedures are the sequential, Gaussian and deep knockoff sampler, as introduced in the main text. The knockoff procedures have access to the test data only and it is worth pointing out that the Gaussian knockoff and Deep knockoff subroutines require dummy encoded data, which significantly expands the dimension of the data matrix if medium- or high-cardinality features are included. In these cases, the prediction models will be defined on the dummy encoded data and CPI evaluated for the variables' respective group of dummys.

Other FI measure procuedures will also calculate their importance scores on the test data set.} 
\subsection{Evaluation Strategy}
The major difficulty in finding an appropriate evaluation strategy is to unify the various methods on a single metric. While the CPI based methods come with inference procedures that allow for testing non-zero importance values, most of the other methods only provide the user with importance scores that indicate a certain ranking of variables according to their assigned importance score. Of course, permutation tests could be performed with these methods, however, because of computational intensity, this is infeasible in practice. We therefore compare methods by contrasting the order in which they rank the variables. 

 In this setting, the goal of the variable importance methods is to recapture the true ranking of variables as precisely as possible. Aggregating such ranking results is, however, a challenging task. While some approaches from voting theory, e.g. the Borda count, can be applied to cases where there is an exact descending order of variables, we here face the difficulty of even determining a true ranking that involves both continuous and categorical variables. In other words, generating a controlled setting in which a categorical variable is more (or less) important to the prediction than a or continuous variable is challenging, since the different levels of the categorical variable each contribute in a different way. We therefore simplify the true ranking to distinguishing between relevant and irrelevant variables. 
 
 According to the DGP, there are 12 variables in total of which 6 are relevant and 6 are not. If a relevant variable is within the top 6 of ranks, we classify it as ``detected" whereas if it was ranked amongst ranks 7-12 it was ``not detected" as relevant by the respective method. This simplification allows us to report the sensitivity and 1-specificity for each of the methods.
 
\begin{figure}[!htbp]
	\centering
	\includegraphics[trim={0 0 0 1cm},clip,width=1\textwidth]{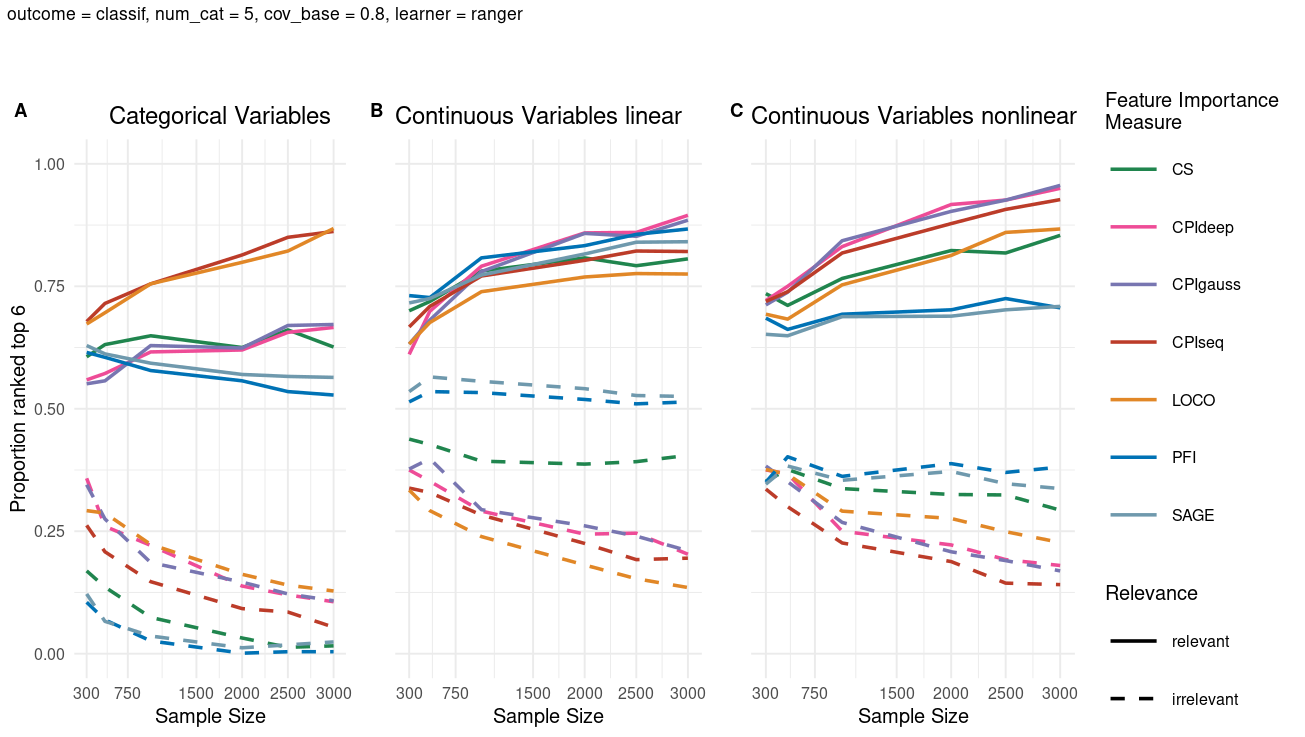}
	\caption{{ Proportion of variables being ranked amongst the top $6$ out of all $12$ variables by type of variable. The solid lines (relevant variables) correspond to sensitivity, whereas the dashed lines (irrelevant variables) correspond to 1-specificity. Categorical variables exhibit $c=5$ levels,  pairwise correlation is $\rho = 0.8$  supervised learner is a random forest, $Y$ binary. }}
	\label{fig::trio_2}
\end{figure}

\begin{figure}[!htbp]
	\centering
	\includegraphics[trim={0 0 0 1cm},clip,width=1\textwidth]{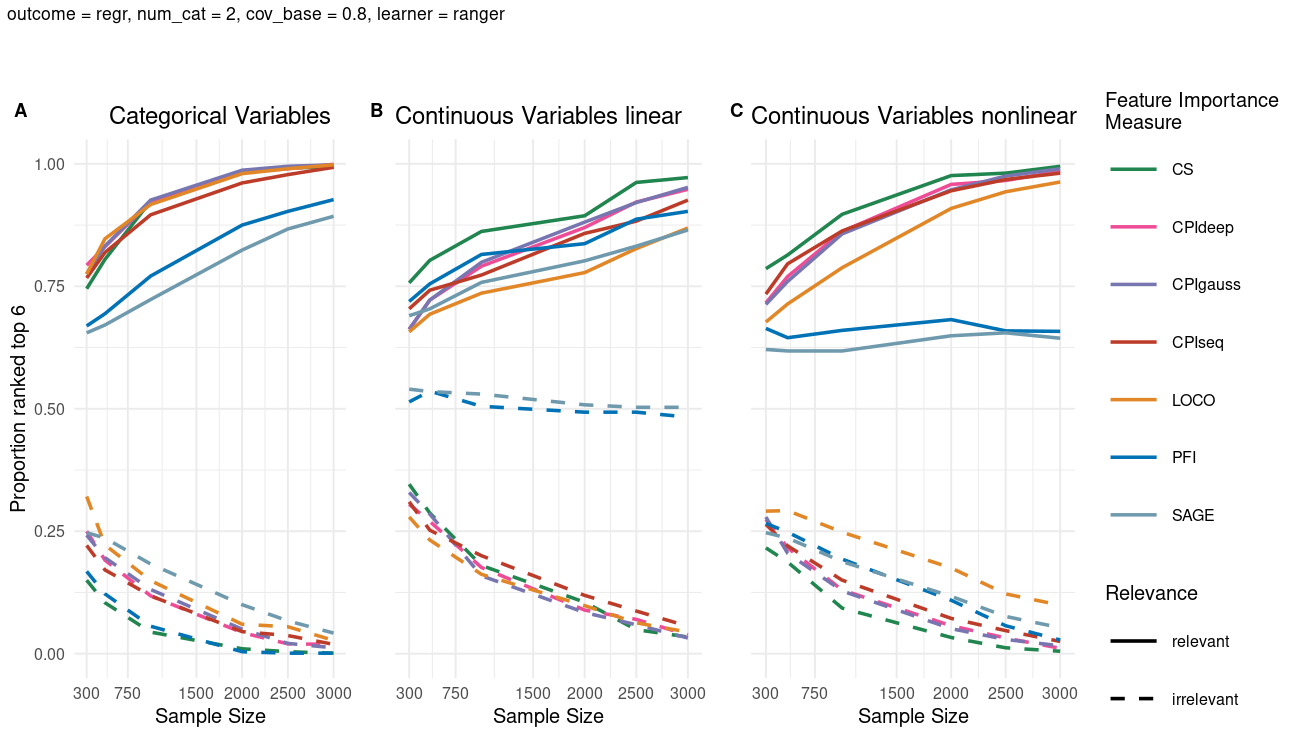}
	\caption{{ Proportion of variables being ranked amongst the top $6$ out of all $12$ variables by type of variable. The solid lines (relevant variables) correspond to sensitivity, whereas the dashed lines (irrelevant variables) correspond to 1-specificity. Categorical variables exhibit $c=2$ levels, pairwise correlation is $\rho = 0.8$, supervised learner is a random forest, $Y$ continuous. }}
	\label{fig::trio_3}
\end{figure}

\begin{figure}[!htbp]
	\centering
	\includegraphics[trim={0 0 0 1cm},clip,width=1\textwidth]{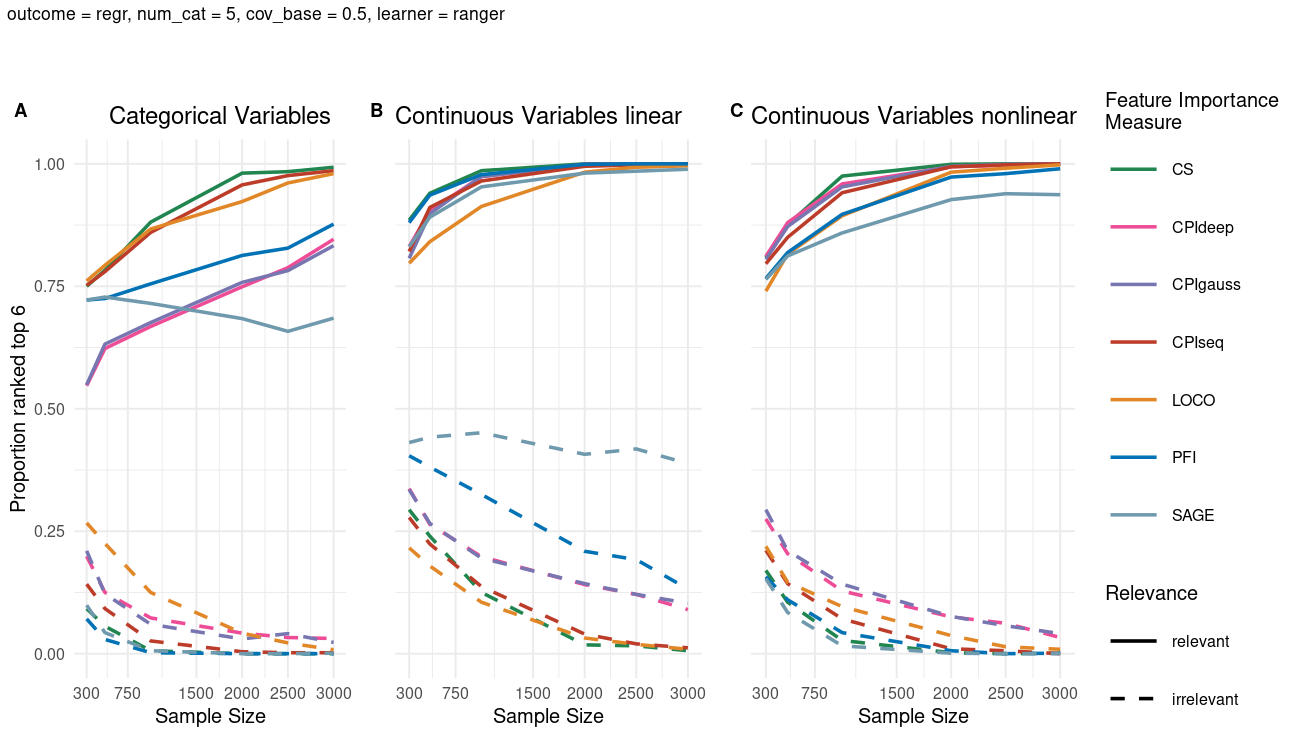}
	\caption{{ Proportion of variables being ranked amongst the top $6$ out of all $12$ variables by type of variable. The solid lines (relevant variables) correspond to sensitivity, whereas the dashed lines (irrelevant variables) correspond to 1-specificity. Categorical variables exhibit $c=5$ levels, pairwise correlation is $\rho = 0.5$, supervised learner is a random forest, $Y$ continuous. }}
	\label{fig::trio_4}
\end{figure}

\begin{figure}[!htbp]
	\centering
	\includegraphics[trim={0 0 0 1cm},clip,width=1\textwidth]{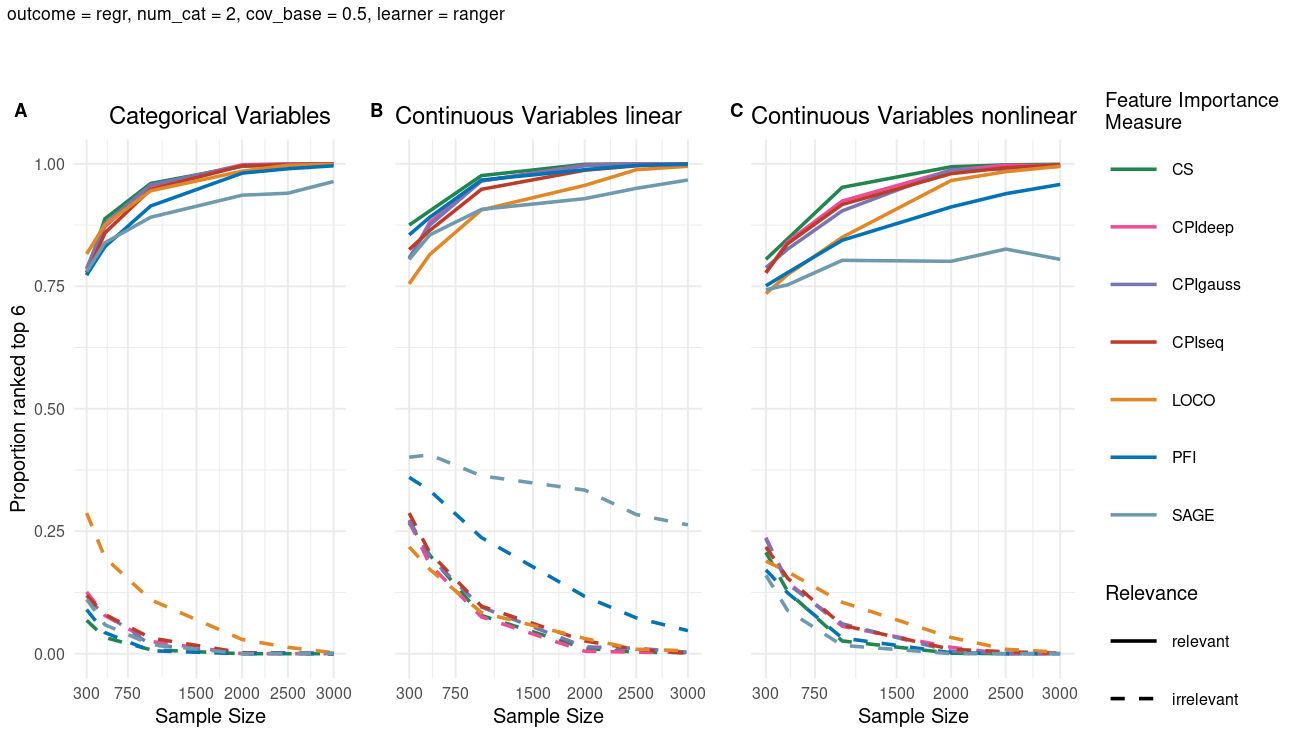}
	\caption{{ Proportion of variables being ranked amongst the top $6$ out of all $12$ variables by type of variable. The solid lines (relevant variables) correspond to sensitivity, whereas the dashed lines (irrelevant variables) correspond to 1-specificity. Categorical variables exhibit $c=2$ levels, pairwise correlation is $\rho = 0.5$, supervised learner is a random forest, $Y$ continuous. }}
	\label{fig::trio_5}
\end{figure}

\begin{figure}[!htbp]
	\centering
	\includegraphics[trim={0 0 0 1cm},clip,width=1\textwidth]{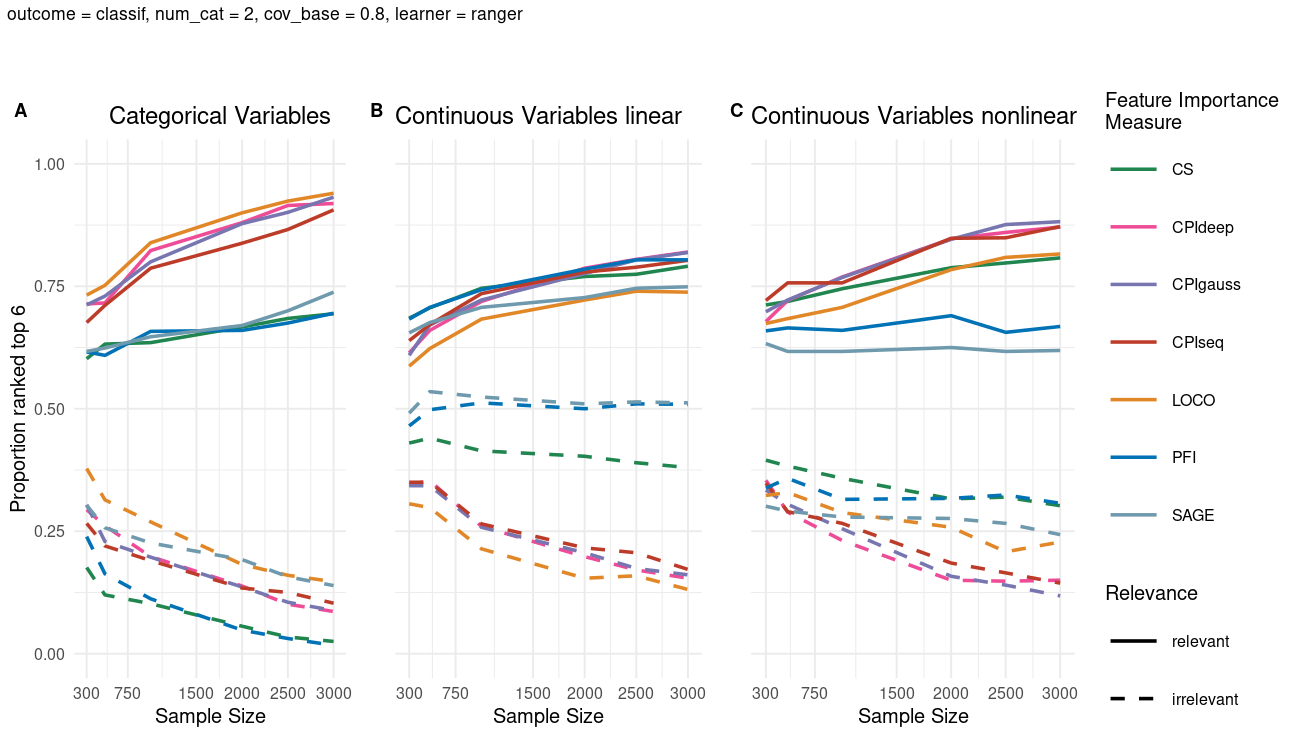}
	\caption{{ Proportion of variables being ranked amongst the top $6$ out of all $12$ variables by type of variable. The solid lines (relevant variables) correspond to sensitivity, whereas the dashed lines (irrelevant variables) correspond to 1-specificity. Categorical variables exhibit $c=2$ levels, pairwise correlation is $\rho = 0.8$, supervised learner is a random forest, $Y$ binary. }}
	\label{fig::trio_6}
\end{figure}

\begin{figure}[!htbp]
	\centering
	\includegraphics[trim={0 0 0 1cm},clip,width=1\textwidth]{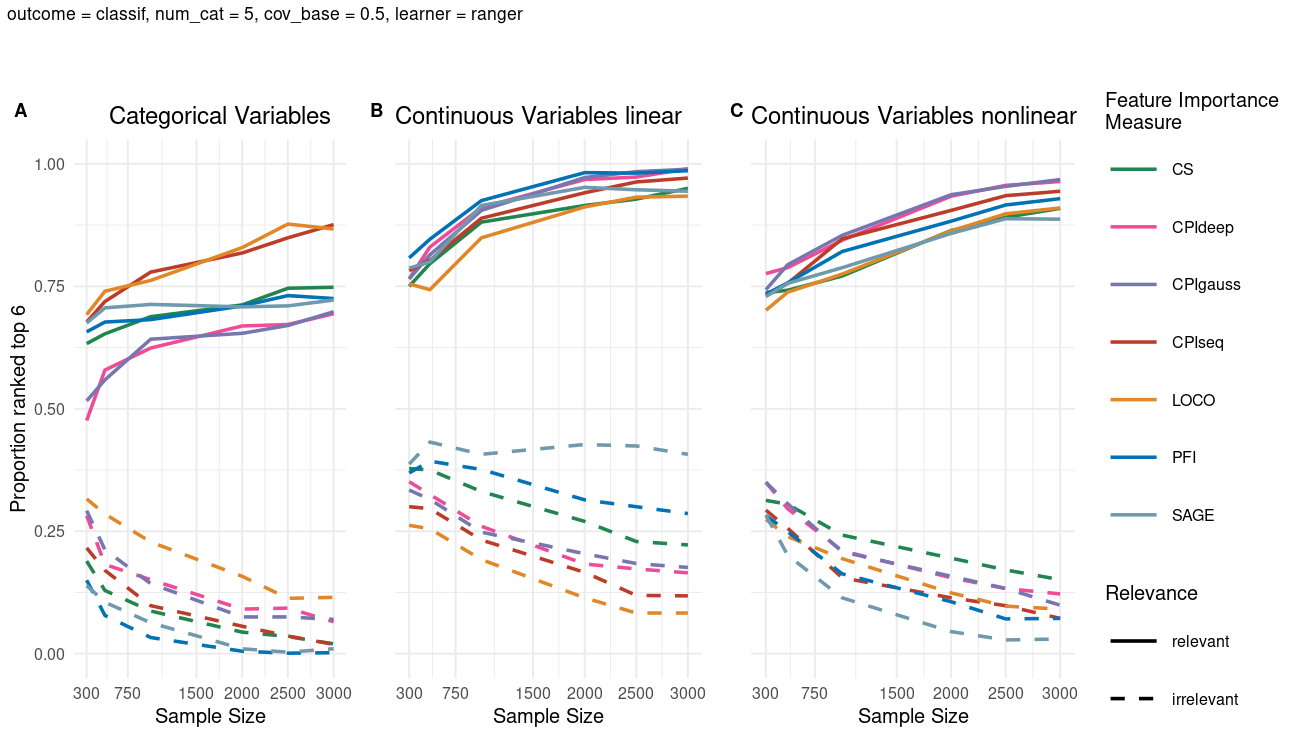}
	\caption{{ Proportion of variables being ranked amongst the top $6$ out of all $12$ variables by type of variable. The solid lines (relevant variables) correspond to sensitivity, whereas the dashed lines (irrelevant variables) correspond to 1-specificity. Categorical variables exhibit $c=5$ levels, pairwise correlation is $\rho = 0.5$, supervised learner is a random forest, $Y$ binary. }}
	\label{fig::trio_7}
\end{figure}

\begin{figure}[!htbp]
	\centering
	\includegraphics[trim={0 0 0 1cm},clip,width=1\textwidth]{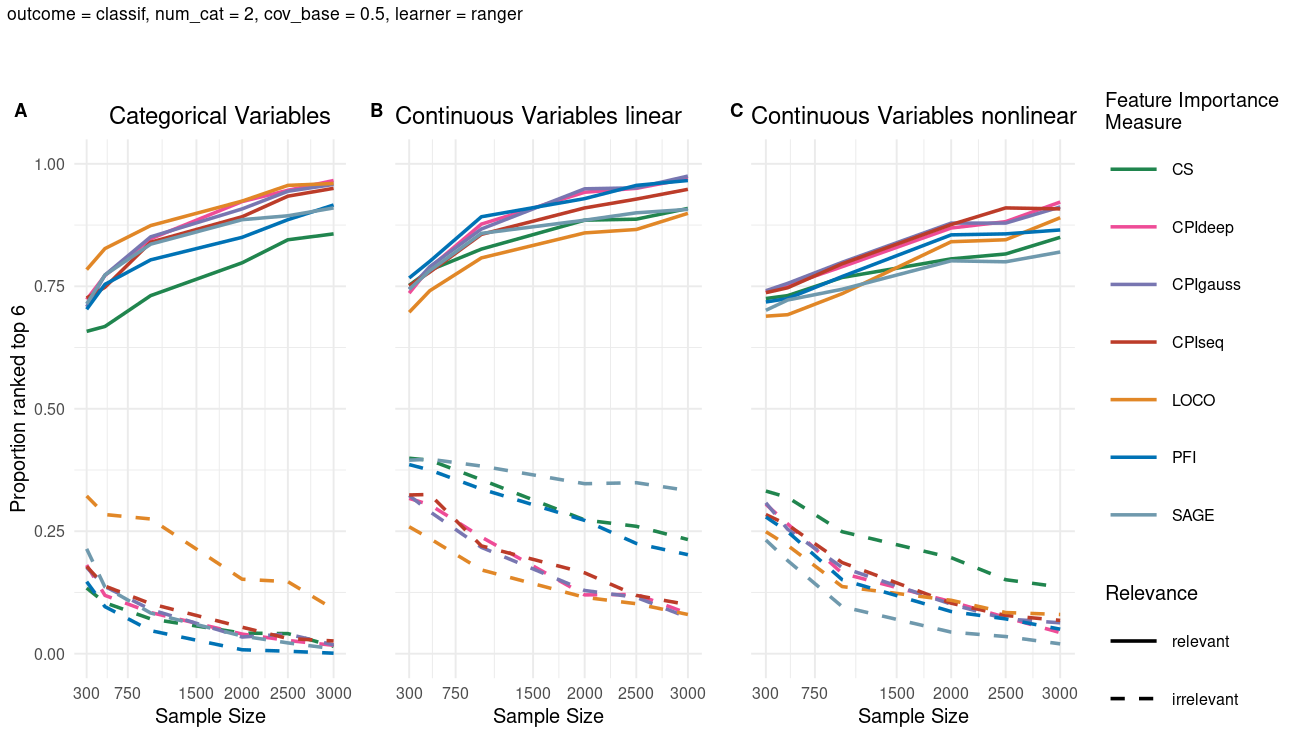}
	\caption{{ Proportion of variables being ranked amongst the top $6$ out of all $12$ variables by type of variable. The solid lines (relevant variables) correspond to sensitivity, whereas the dashed lines (irrelevant variables) correspond to 1-specificity. Categorical variables exhibit $c=2$ levels, pairwise correlation is $\rho = 0.5$, supervised learner is a random forest, $Y$ binary. }}
	\label{fig::trio_8}
\end{figure}

\begin{figure}[!htbp]
	\centering
	\includegraphics[trim={0 0 0 1cm},clip,width=1\textwidth]{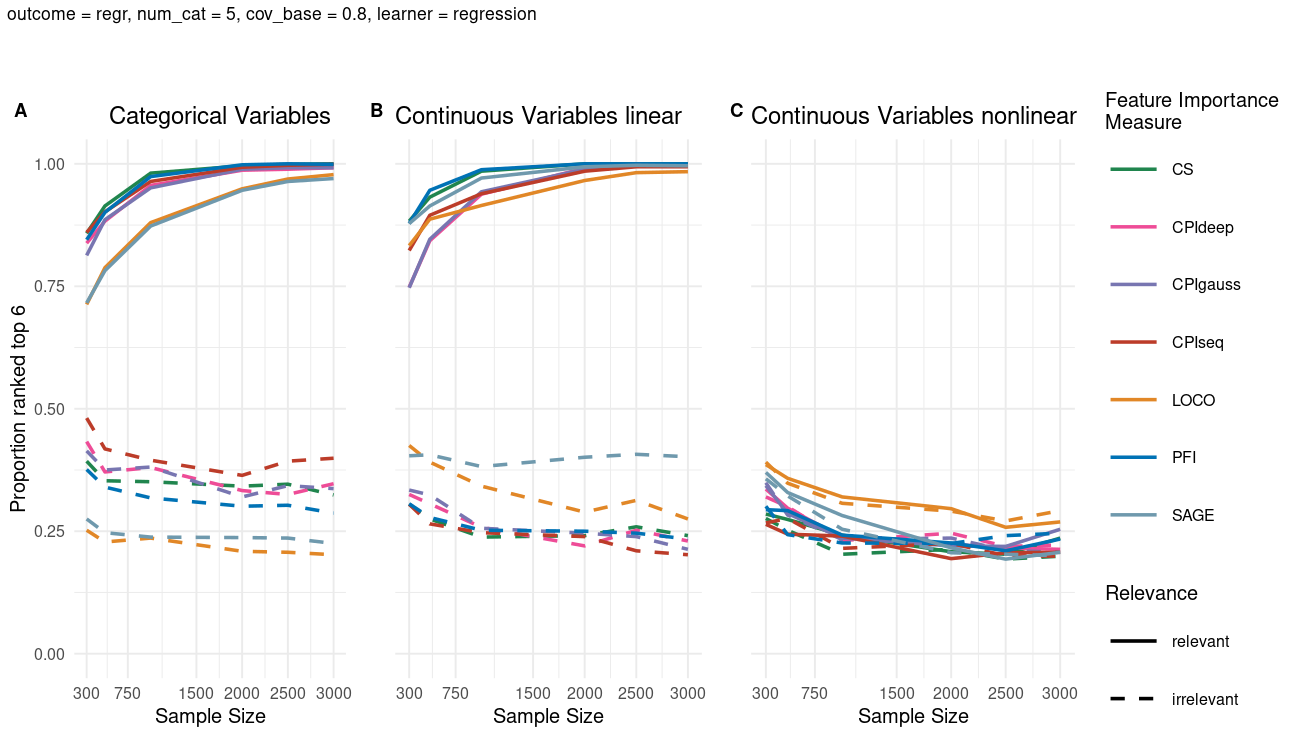}
	\caption{{ Proportion of variables being ranked amongst the top $6$ out of all $12$ variables by type of variable. The solid lines (relevant variables) correspond to sensitivity, whereas the dashed lines (irrelevant variables) correspond to 1-specificity. Categorical variables exhibit $c=5$ levels, signal to noise ratio is set to $2$, pairwise correlation is $\rho = 0.8$ supervised learner is a linear regression model, $Y$ continuous. }}
	\label{fig::trio_9}
\end{figure}

\begin{figure}[!htbp]
	\centering
	\includegraphics[trim={0 0 0 1cm},clip,width=1\textwidth]{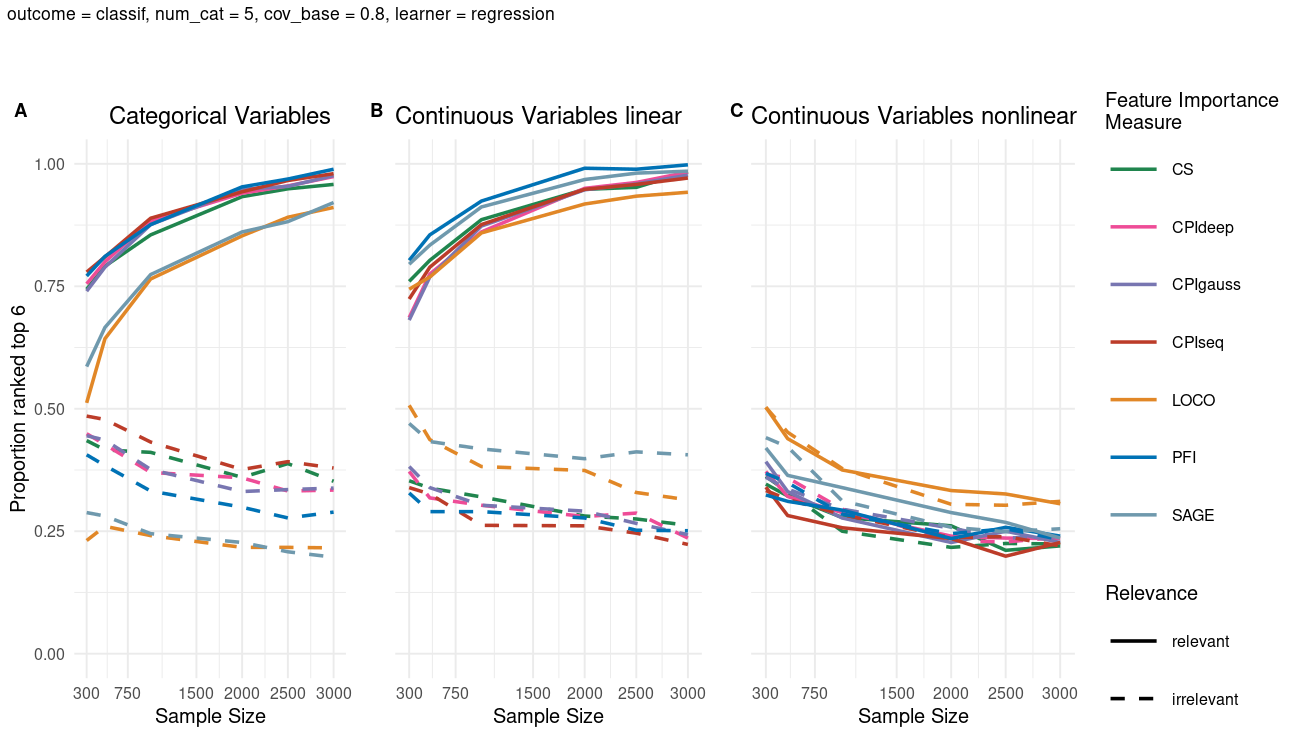}
	\caption{{ Proportion of variables being ranked amongst the top $6$ out of all $12$ variables by type of variable. The solid lines (relevant variables) correspond to sensitivity, whereas the dashed lines (irrelevant variables) correspond to 1-specificity. Categorical variables exhibit $c=5$ levels, signal to noise ratio is set to $2$, pairwise correlation is $\rho = 0.8$ supervised learner is a logistic regression model, $Y$ binary. }}
	\label{fig::trio_10}
\end{figure}

\begin{figure}[!htbp]
	\centering
	\includegraphics[trim={0 0 0 1cm},clip,width=1\textwidth]{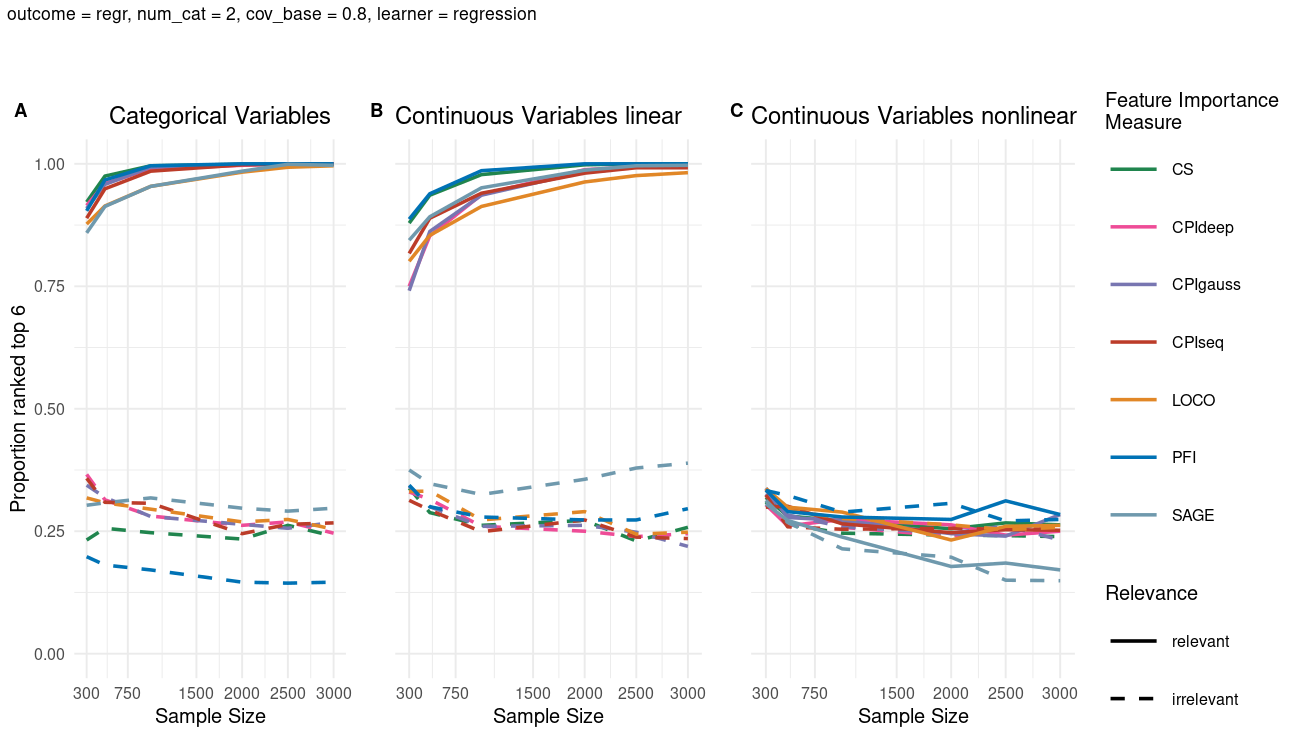}
	\caption{{ Proportion of variables being ranked amongst the top $6$ out of all $12$ variables by type of variable. The solid lines (relevant variables) correspond to sensitivity, whereas the dashed lines (irrelevant variables) correspond to 1-specificity. Categorical variables exhibit $c=2$ levels, signal to noise ratio is set to $2$, pairwise correlation is $\rho = 0.8$ supervised learner is a linear regression model, $Y$ continuous. }}
	\label{fig::trio_11}
\end{figure}

\begin{figure}[!htbp]
	\centering
	\includegraphics[trim={0 0 0 1cm},clip,width=1\textwidth]{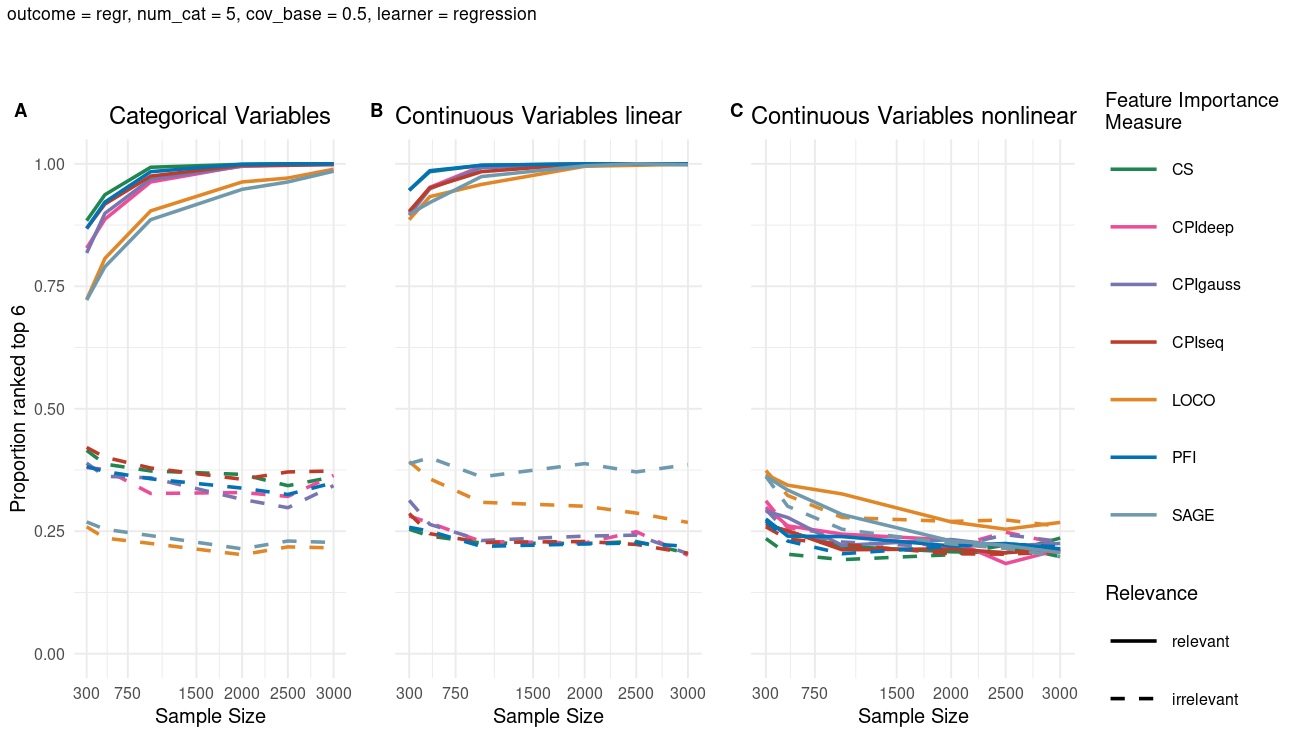}
	\caption{{ Proportion of variables being ranked amongst the top $6$ out of all $12$ variables by type of variable. The solid lines (relevant variables) correspond to sensitivity, whereas the dashed lines (irrelevant variables) correspond to 1-specificity. Categorical variables exhibit $c=5$ levels, signal to noise ratio is set to $2$, pairwise correlation is $\rho = 0.5$ supervised learner is a linear regression model, $Y$ continuous. }}
	\label{fig::trio_12}
\end{figure}

\begin{figure}[!htbp]
	\centering
	\includegraphics[trim={0 0 0 1cm},clip,width=1\textwidth]{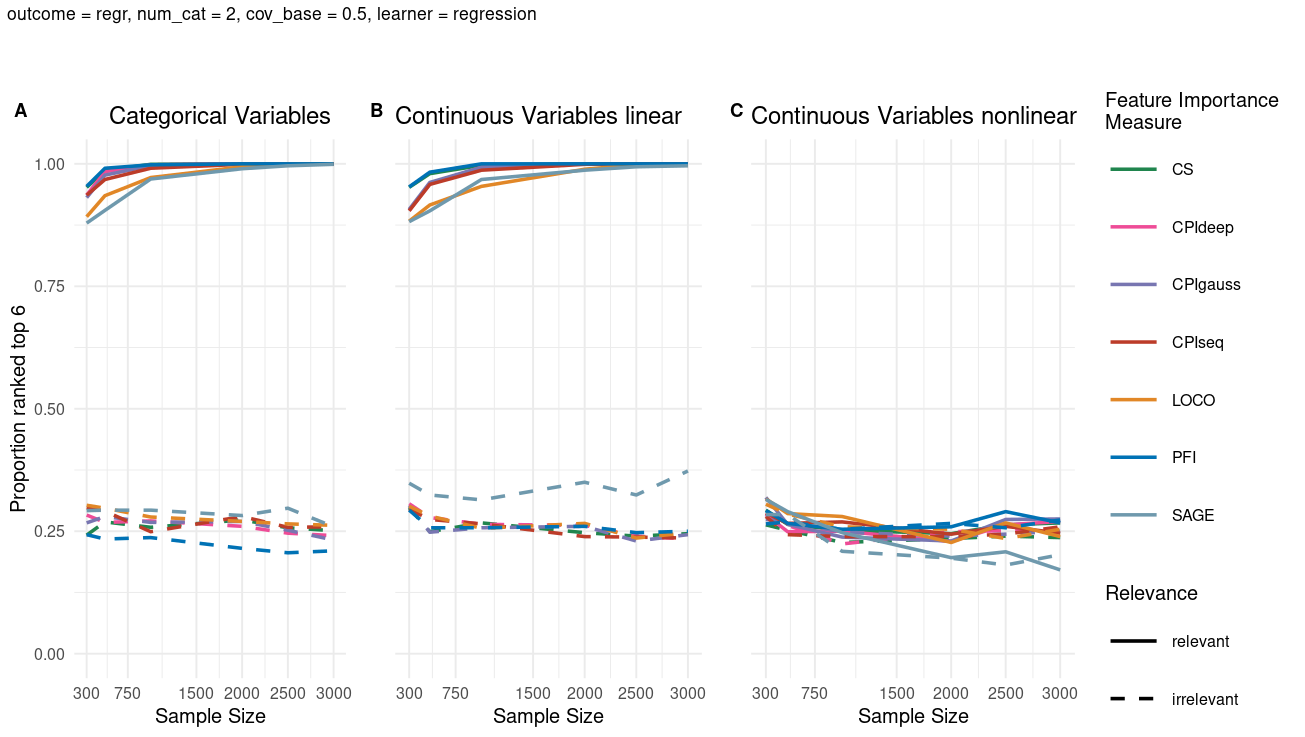}
	\caption{{ Proportion of variables being ranked amongst the top $6$ out of all $12$ variables by type of variable. The solid lines (relevant variables) correspond to sensitivity, whereas the dashed lines (irrelevant variables) correspond to 1-specificity. Categorical variables exhibit $c=2$ levels, signal to noise ratio is set to $2$, pairwise correlation is $\rho = 0.5$ supervised learner is a linear regression model, $Y$ continuous. }}
	\label{fig::trio_13}
\end{figure}

\begin{figure}[!htbp]
	\centering
	\includegraphics[trim={0 0 0 1cm},clip,width=1\textwidth]{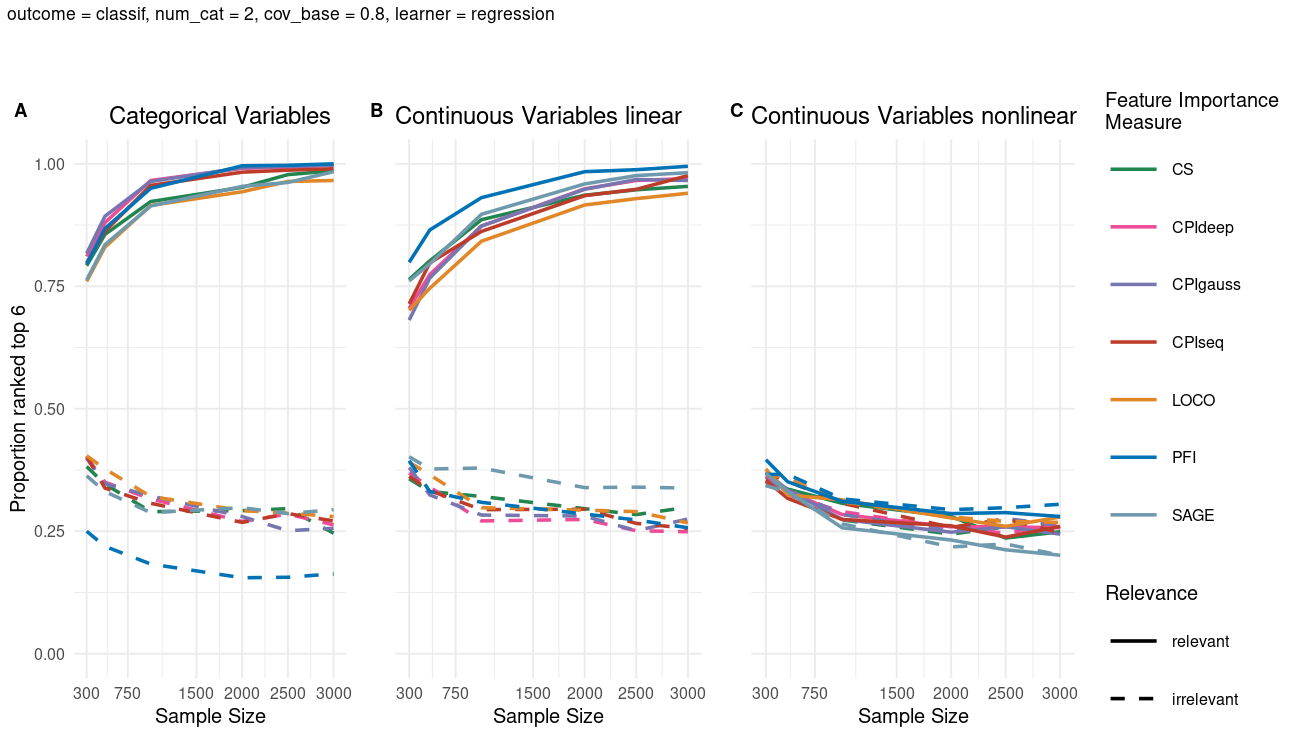}
	\caption{{ Proportion of variables being ranked amongst the top $6$ out of all $12$ variables by type of variable. The solid lines (relevant variables) correspond to sensitivity, whereas the dashed lines (irrelevant variables) correspond to 1-specificity. Categorical variables exhibit $c=2$ levels, signal to noise ratio is set to $2$, pairwise correlation is $\rho = 0.8$ supervised learner is a logistic regression model, $Y$ binary. }}
	\label{fig::trio_14}
\end{figure}

\begin{figure}[!htbp]
	\centering
	\includegraphics[trim={0 0 0 1cm},clip,width=1\textwidth]{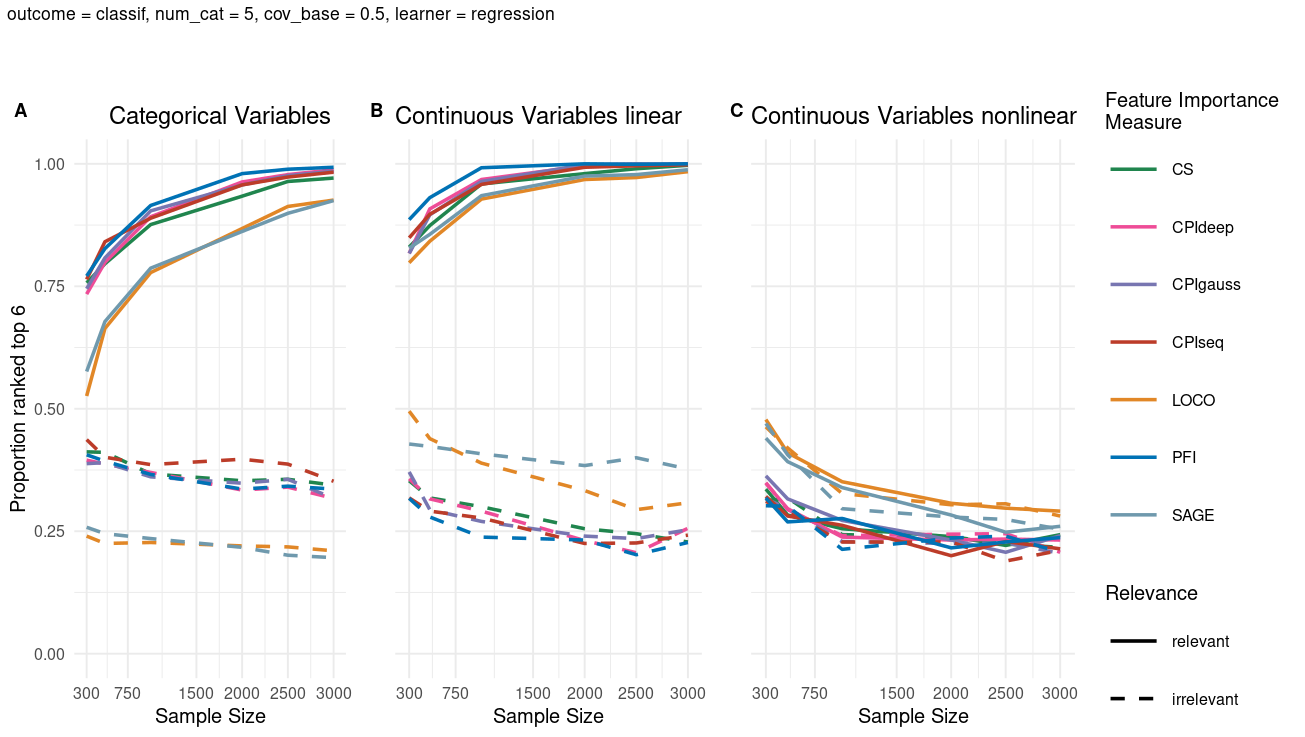}
	\caption{{ Proportion of variables being ranked amongst the top $6$ out of all $12$ variables by type of variable. The solid lines (relevant variables) correspond to sensitivity, whereas the dashed lines (irrelevant variables) correspond to 1-specificity. Categorical variables exhibit $c=5$ levels, signal to noise ratio is set to $2$, pairwise correlation is $\rho = 0.5$ supervised learner is a logistic regression model, $Y$ binary. }}
	\label{fig::trio_15}
\end{figure}

\begin{figure}[!htbp]
	\centering
	\includegraphics[trim={0 0 0 1cm},clip,width=1\textwidth]{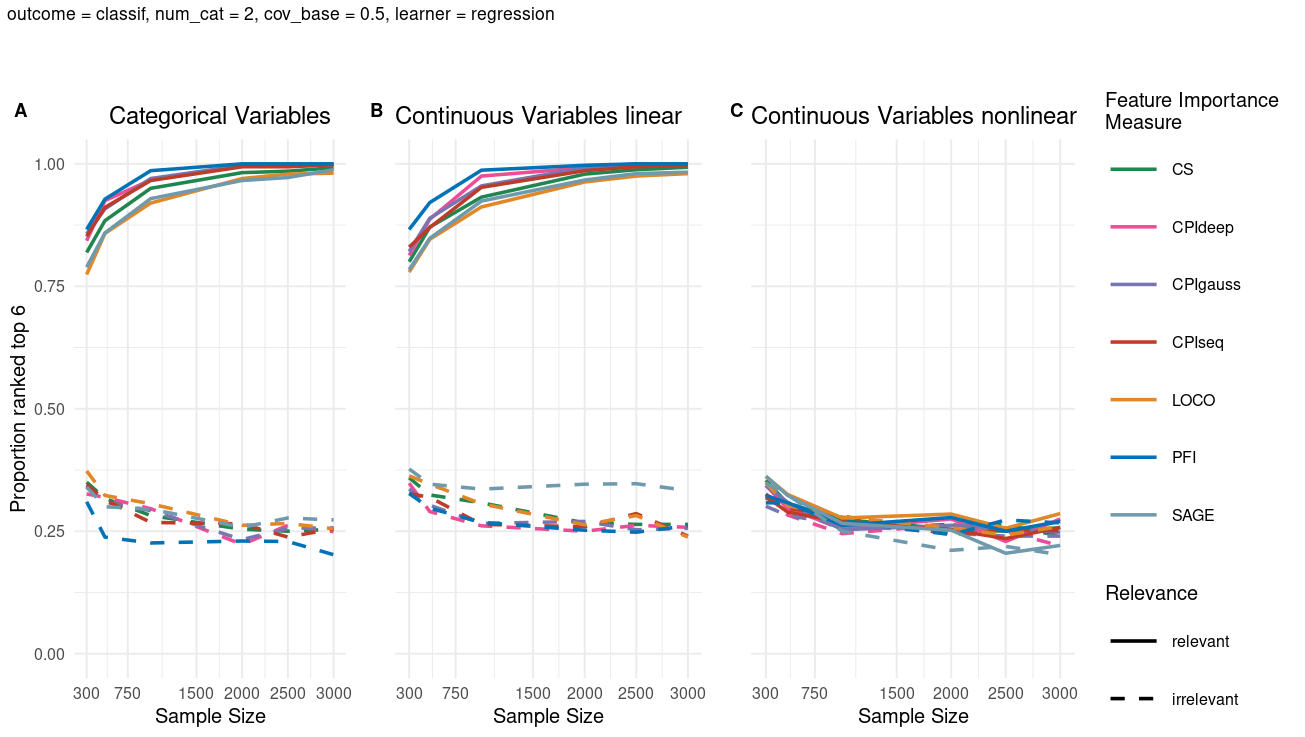}
	\caption{{ Proportion of variables being ranked amongst the top $6$ out of all $12$ variables by type of variable. The solid lines (relevant variables) correspond to sensitivity, whereas the dashed lines (irrelevant variables) correspond to 1-specificity. Categorical variables exhibit $c=5$ levels, signal to noise ratio is set to $2$, pairwise correlation is $\rho = 0.5$ supervised learner is a logistic regression model, $Y$ binary. }}
	\label{fig::trio_16}
\end{figure}

\begin{figure}[!htbp]
	\centering
	\includegraphics[trim={0 0 0 1cm},clip,width=1\textwidth]{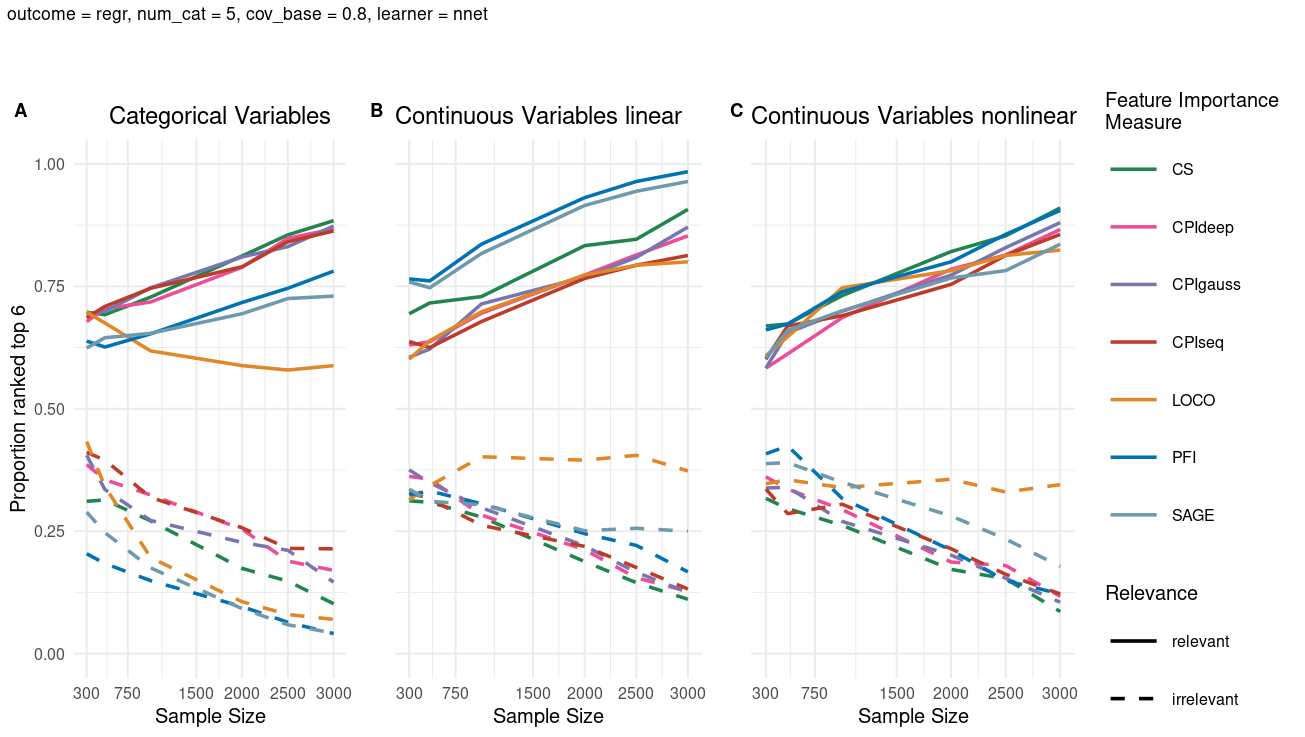}
	\caption{{ Proportion of variables being ranked amongst the top $6$ out of all $12$ variables by type of variable. The solid lines (relevant variables) correspond to sensitivity, whereas the dashed lines (irrelevant variables) correspond to 1-specificity. Categorical variables exhibit $c=5$ levels, signal to noise ratio is set to $2$, pairwise correlation is $\rho = 0.8$ and supervised learner is a neural network, $Y$ continuous. }}
	\label{fig::trio_17}
\end{figure}

\begin{figure}[!htbp]
	\centering
	\includegraphics[trim={0 0 0 1cm},clip,width=1\textwidth]{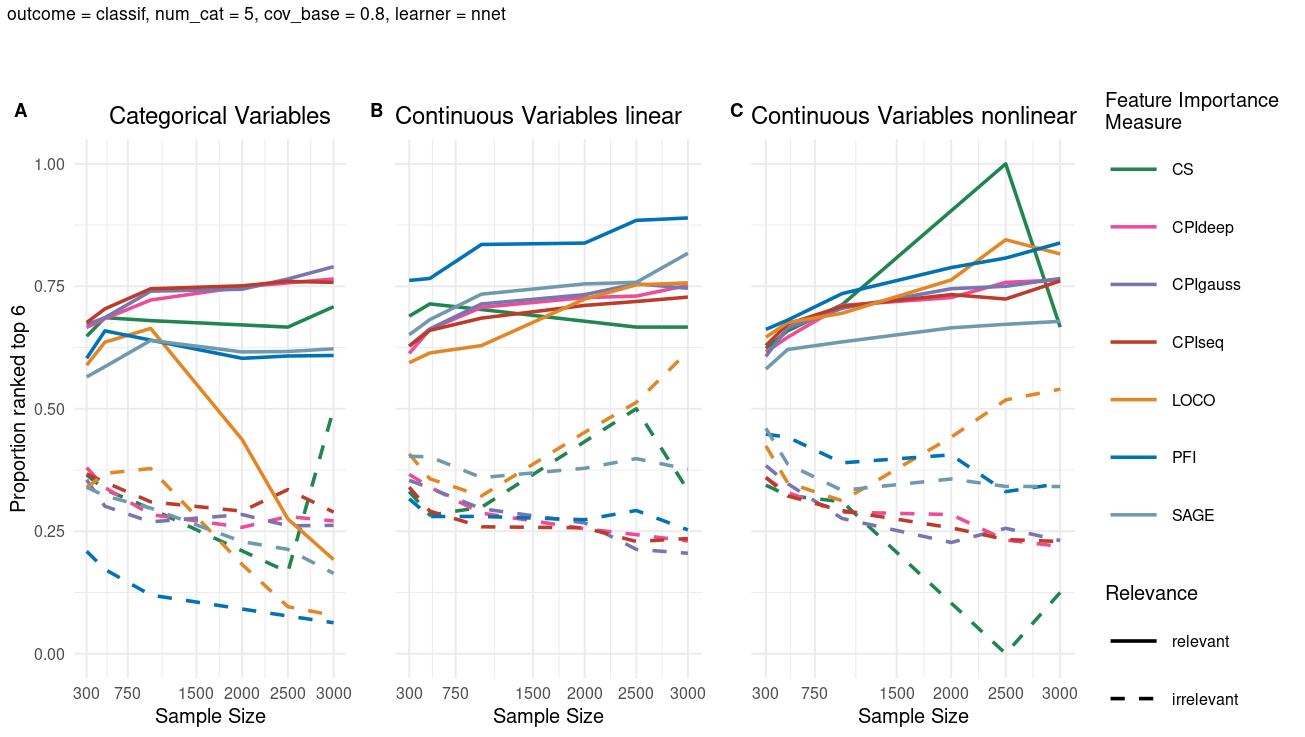}
	\caption{{ Proportion of variables being ranked amongst the top $6$ out of all $12$ variables by type of variable. The solid lines (relevant variables) correspond to sensitivity, whereas the dashed lines (irrelevant variables) correspond to 1-specificity. Categorical variables exhibit $c=5$ levels, signal to noise ratio is set to $2$, pairwise correlation is $\rho = 0.8$ and supervised learner is a neural network, $Y$ binary. }}
	\label{fig::trio_18}
\end{figure}

\begin{figure}[!htbp]
	\centering
	\includegraphics[trim={0 0 0 1cm},clip,width=1\textwidth]{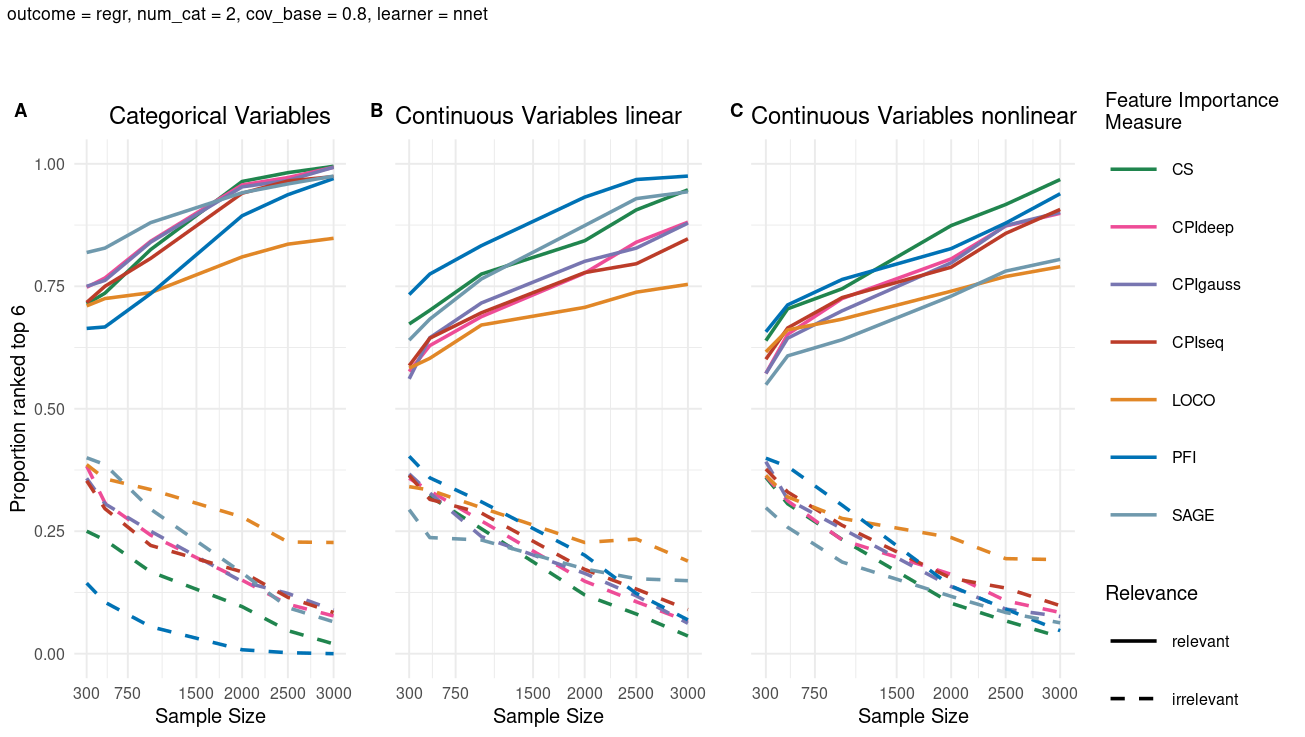}
	\caption{{ Proportion of variables being ranked amongst the top $6$ out of all $12$ variables by type of variable. The solid lines (relevant variables) correspond to sensitivity, whereas the dashed lines (irrelevant variables) correspond to 1-specificity. Categorical variables exhibit $c=2$ levels, signal to noise ratio is set to $2$, pairwise correlation is $\rho = 0.8$ and supervised learner is a neural network, $Y$ continuous. }}
	\label{fig::trio_19}
\end{figure}

\begin{figure}[!htbp]
	\centering
	\includegraphics[trim={0 0 0 1cm},clip,width=1\textwidth]{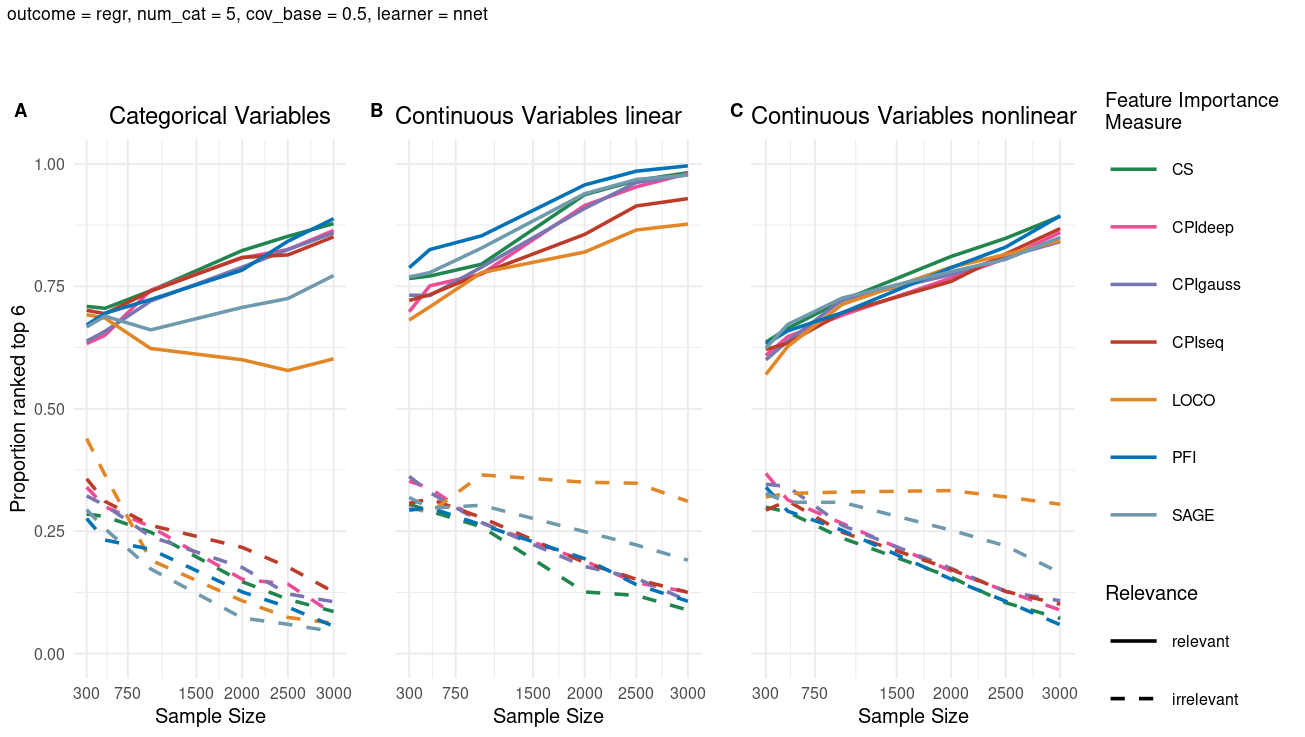}
	\caption{{ Proportion of variables being ranked amongst the top $6$ out of all $12$ variables by type of variable. The solid lines (relevant variables) correspond to sensitivity, whereas the dashed lines (irrelevant variables) correspond to 1-specificity. Categorical variables exhibit $c=5$ levels, signal to noise ratio is set to $2$, pairwise correlation is $\rho = 0.5$ and supervised learner is a neural network, $Y$ continuous. }}
	\label{fig::trio_20}
\end{figure}

\begin{figure}[!htbp]
	\centering
	\includegraphics[trim={0 0 0 1cm},clip,width=1\textwidth]{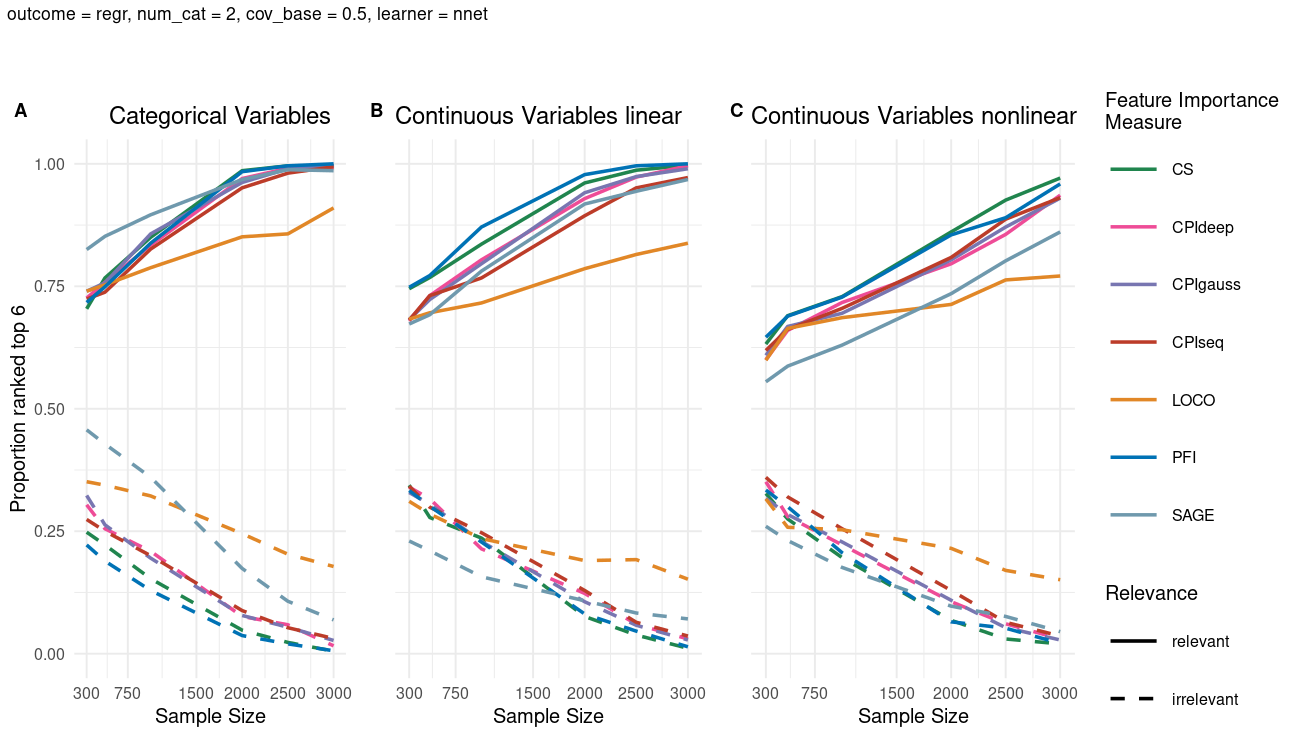}
	\caption{{ Proportion of variables being ranked amongst the top $6$ out of all $12$ variables by type of variable. The solid lines (relevant variables) correspond to sensitivity, whereas the dashed lines (irrelevant variables) correspond to 1-specificity. Categorical variables exhibit $c=2$ levels, signal to noise ratio is set to $2$, pairwise correlation is $\rho = 0.8$ and supervised learner is a neural network, $Y$ continuous. }}
	\label{fig::trio_21}
\end{figure}

\begin{figure}[!htbp]
	\centering
	\includegraphics[trim={0 0 0 1cm},clip,width=1\textwidth]{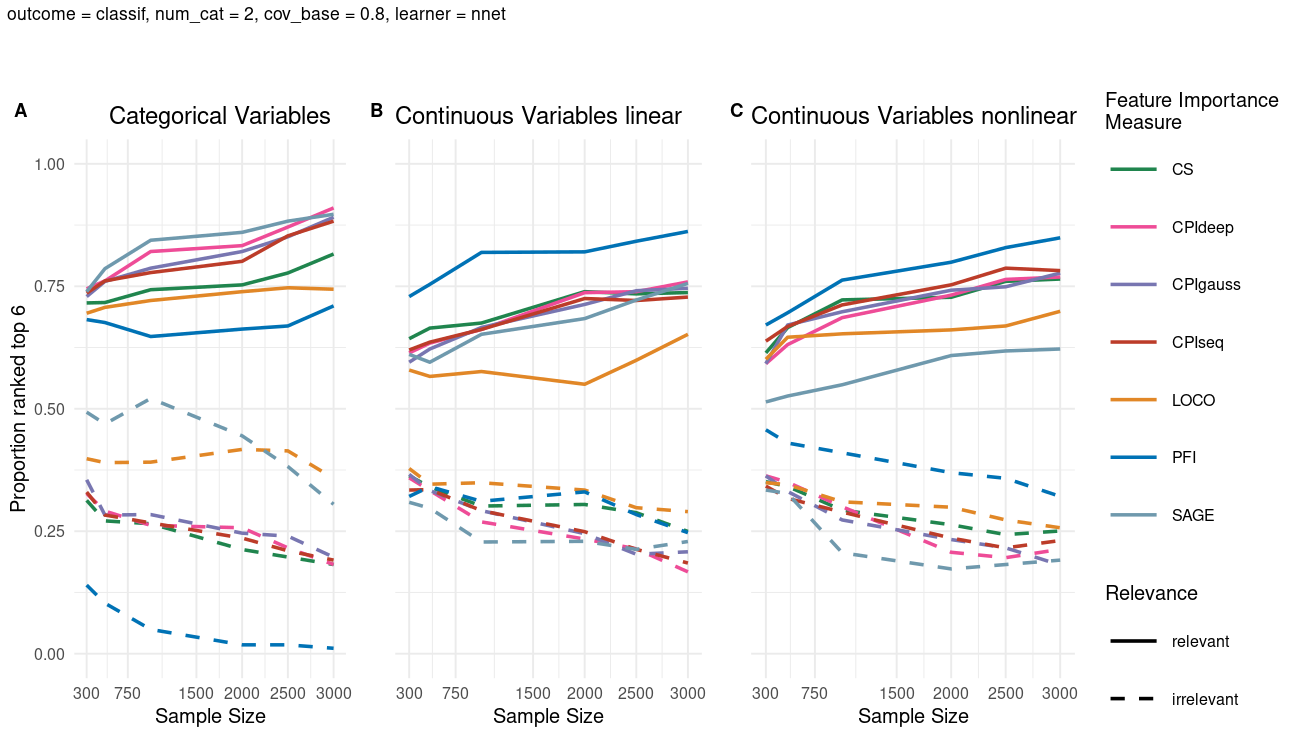}
	\caption{{ Proportion of variables being ranked amongst the top $6$ out of all $12$ variables by type of variable. The solid lines (relevant variables) correspond to sensitivity, whereas the dashed lines (irrelevant variables) correspond to 1-specificity. Categorical variables exhibit $c=2$ levels, signal to noise ratio is set to $2$, pairwise correlation is $\rho = 0.8$ and supervised learner is a neural network, $Y$ binary. }}
	\label{fig::trio_22}
\end{figure}

\begin{figure}[!htbp]
	\centering
	\includegraphics[trim={0 0 0 1cm},clip,width=1\textwidth]{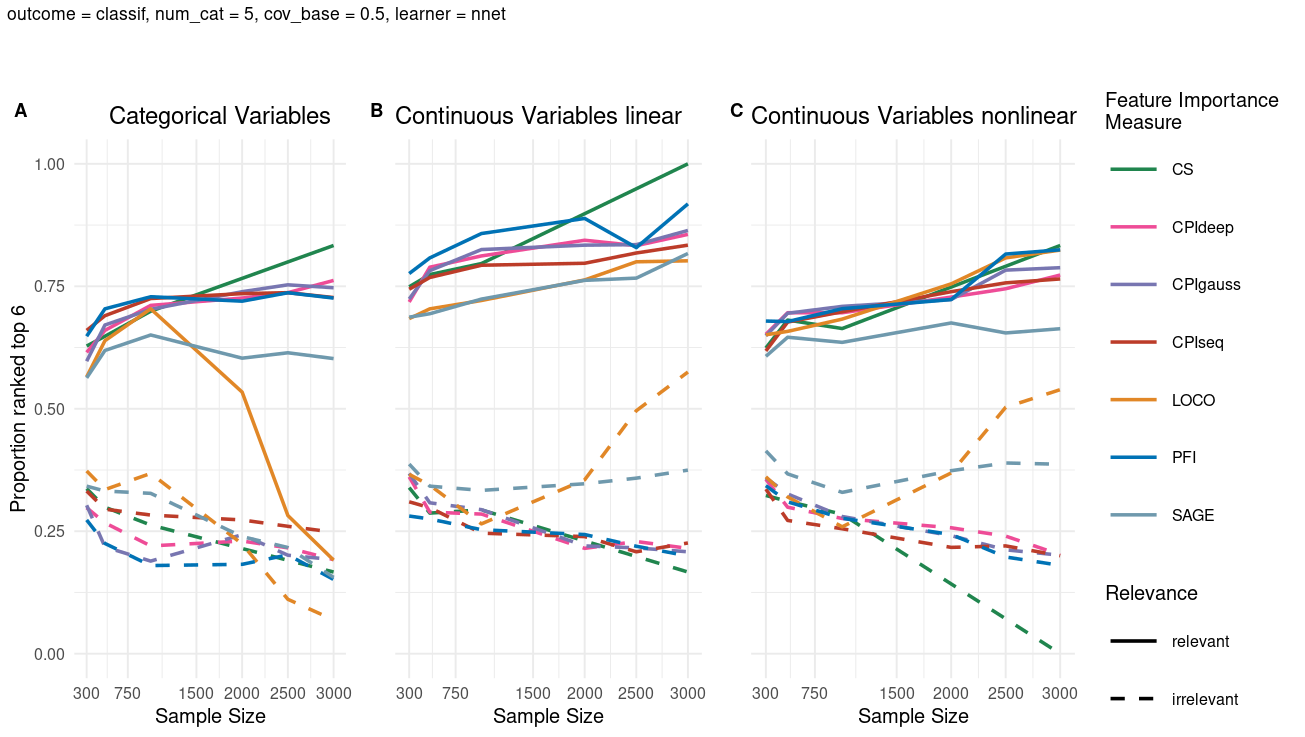}
	\caption{{ Proportion of variables being ranked amongst the top $6$ out of all $12$ variables by type of variable. The solid lines (relevant variables) correspond to sensitivity, whereas the dashed lines (irrelevant variables) correspond to 1-specificity. Categorical variables exhibit $c=5$ levels, signal to noise ratio is set to $2$, pairwise correlation is $\rho = 0.5$ and supervised learner is a neural network, $Y$ binary. }}
	\label{fig::trio_23}
\end{figure}

\begin{figure}[!htbp]
	\centering
	\includegraphics[trim={0 0 0 1cm},clip,width=1\textwidth]{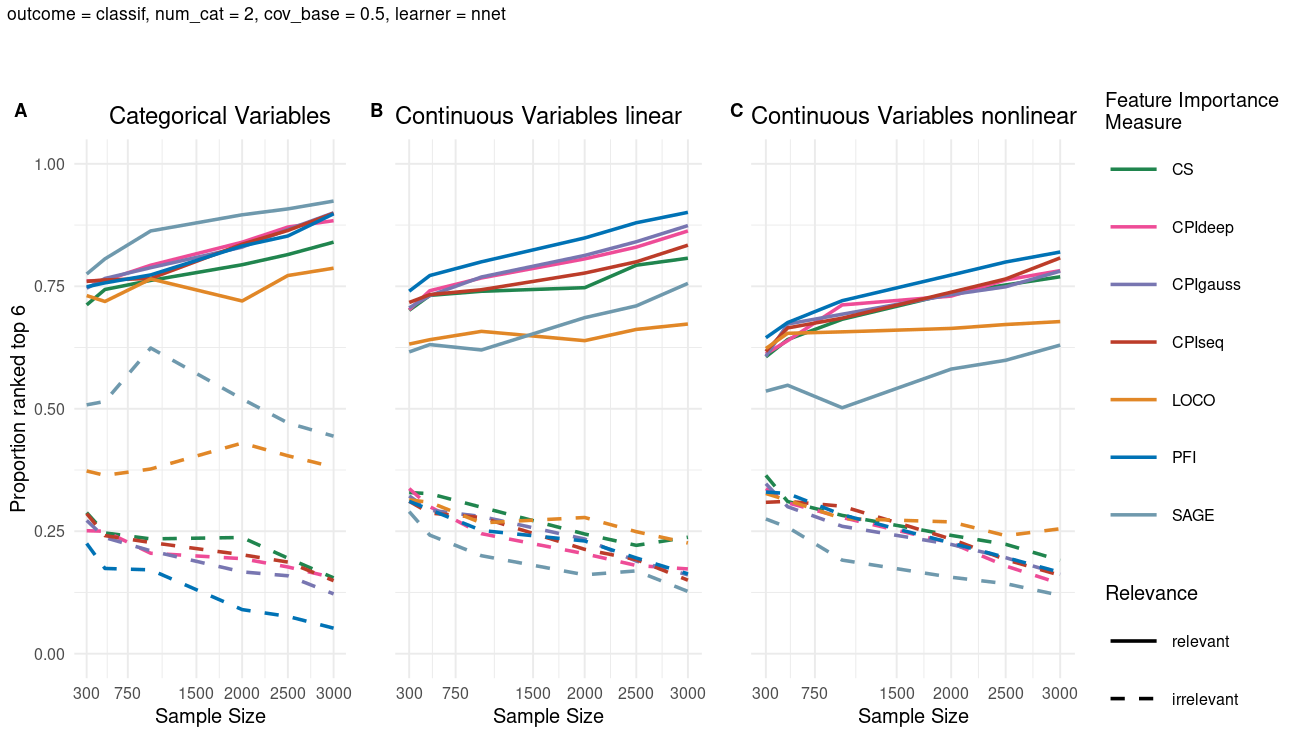}
	\caption{{ Proportion of variables being ranked amongst the top $6$ out of all $12$ variables by type of variable. The solid lines (relevant variables) correspond to sensitivity, whereas the dashed lines (irrelevant variables) correspond to 1-specificity. Categorical variables exhibit $c=2$ levels, signal to noise ratio is set to $2$, pairwise correlation is $\rho = 0.5$ and supervised learner is a neural network, $Y$ binary. }}
	\label{fig::trio_24}
\end{figure}

 Here, sensitivity, in the sense of a true positive rate, is the frequency with which important variables have been detected as important across replicates. Contrary, 1-specificity, i.e. the rate of false positives, is given by the frequency with which irrelevant variables were mistakenly detected as relevant across replicates.{  
 Further, we report area under the receiver operating curve (AUC) as a metric for model evaluation. Given our simulation setup, 6 out of the 12 variables are important to the outcome and the goal for the FI procedures is to detect them by ranking them above the unimportant features. This is akin to a standard classification problem (important vs. unimportant), with FI serving as our score instead of predicted class probabilities. Note that in the present case, ranking more than 6 variables in the top 6 ranks, or more than 6 variables in the bottom ranks, respectively, corresponds to ties in the ranking. We can interpret the resulting AUC as the probability that the variable importance procedure will correctly rank a randomly selected important variable above a randomly selected unimportant variable.

 \subsection{Model Validation}
A good model fit is an important prerequisite for reliable model-agnostic FI measurement. To investigate the model fit for the simulations in Section 3.2, we validate the models by comparing their performance to the best possible fit (oracle prediction model) on the generated data. 

For regression tasks, the data generating process of the experiments in Section 3.2 is designed such that $SNR = 2$. We use $\text{R}^2$ as a measure of model performance and the SNR directly implies the optimum $\text{R}^{*2}$ that can be achieved. That is, $\text{R}^{*2} = \frac{\text{SNR}}{\text{SNR} + 1} = 2/3$.

For classification tasks, the data generating process of the experiments in Section 3.2 is designed such that $\text{BER} = 0.2$. Here, we use accuracy as a measure of model performance and hence, the best possible accuracy an oracle prediction model can achieve is $= 1 -\text{BER} =0.8$. 

To validate the models used in our experiments, we report the achieved $\text{R}^{2}$ and accuracy of the respective models when fitted on the training data -- as it is done by the FI measurement procedures -- and evaluated on the test data. In Figure \ref{fig::validata_models_3.2}, we report the average score and standard deviation across all datasets that were used during the Section 3.2 simulations, i.e. 500 replicates for each of the 24 DGP parameter combinations, see further Section \ref{app::experiments_2}.

We find the models to achieve a good fit overall for the respective datasets, especially for large sample sizes. However, the nnet algorithm appears to require a large training sample sizes to achieve a good fit. Further, we note that the regression model reaches a performance plateau, which is reasonable since this model cannot capture the effects of the nonlinear variables contained in the data sets. As a consequence, the FI measurement for nonlinear variables is hindered when using a linear or logistic regression model (see further Section \ref{sec::results_regression}). 

  \begin{figure}[!htbp]
	\centering
	\includegraphics[trim={0 0 0 0cm},clip,width=1\textwidth]{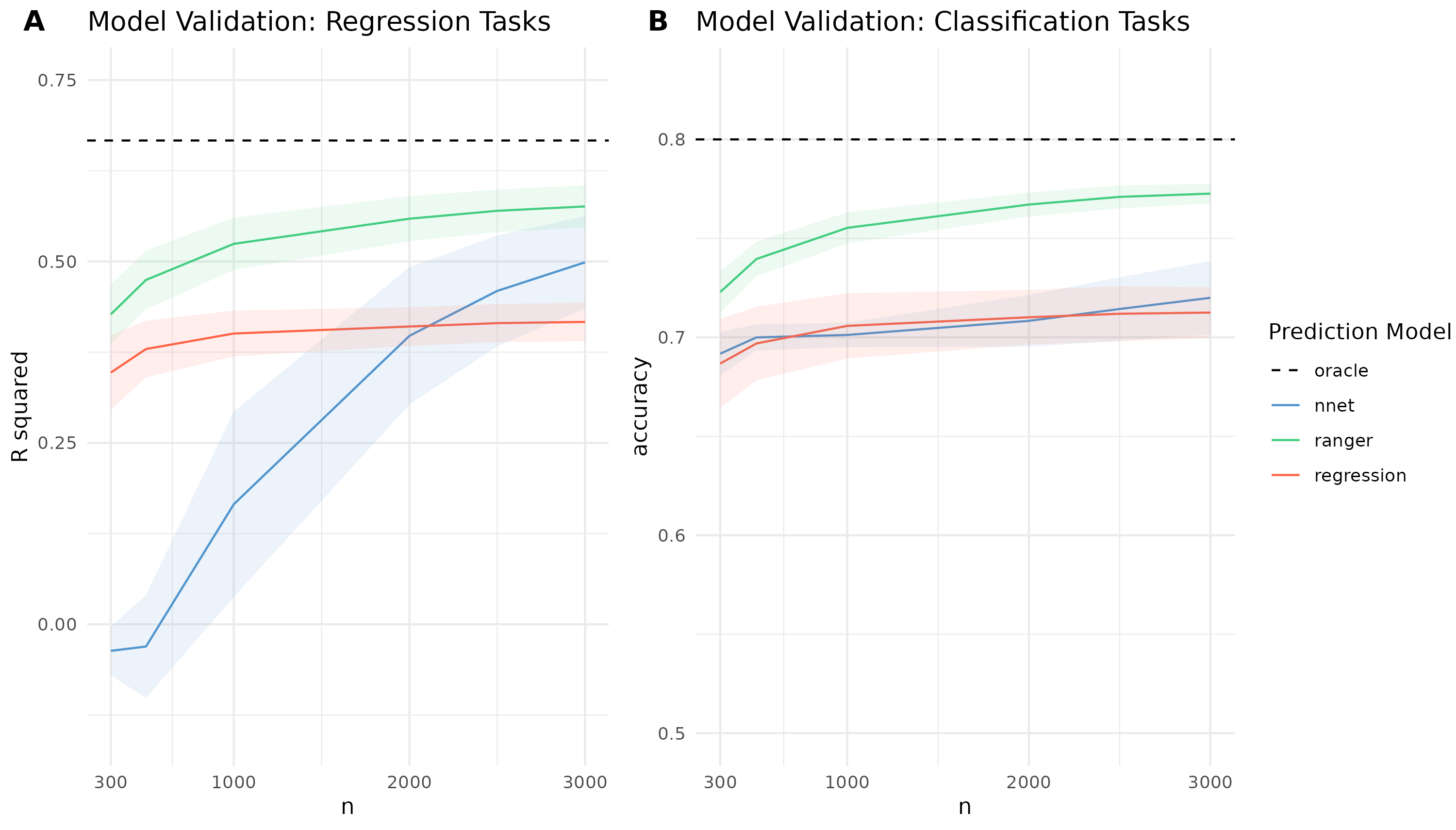}
	\caption{ {   Average score and standard deviation of $\text{R}^{2}$ (Panel \textbf{A}) and accuracy (Panel \textbf{B}) for least-squares regression or logistic regression models, random forest (ranger) models and a neural network (nnet) models across datasets generated in the simulations of Section 3.2.}}
 \label{fig::validata_models_3.2}
\end{figure}
\newpage
 \subsection{Results}
 In Figs. \ref{fig::trio_2} -  \ref{fig::trio_24}, we can see sensitivity (relevant variables, solid lines) and 1-specificity (irrelevant variables, dashed lines) according to the types of variables for various parameter settings.

We group the results according to the prediction model used. In section \ref{sec::results_ranger}, results for the random forest prediction model are given. For these results, we can see that the CS approach loses power to detect relevant categorical variables with a binary target variable more pronounced than competing models. 

Section \ref{sec::results_regression} gives the results for regression models, i.e. the least-squares regression model for continuous $Y$ and the logistic regression model for binary $Y$. Clearly, the prediction model is not able to capture the importance of nonlinear variables appropriately, which is expected due to the linear nature of the prediction model. Resulting from this, the detection frequency of relevant nonlinear continuous variables is similar to the detection rate of irrelevant nonlinear variables. This effect is apparent in across all parameter configurations and variable importance measures. 

From the results for the neural network prediction model (Section \ref{sec::results_nnet}), we can again see the consequences of the chosen prediction model for the variable importance measures. The neural network is quite unstable in the sense that it yields diverse fits for various sample sizes and feature spaces. This results in volatile results for all variable importance measures considered and in particular, the LOCO method suffers from this issue due to the high number of model refits required during variable importance evaluation.
}
\newpage
 
\subsubsection{Prediction Model: Random Forest} \label{sec::results_ranger}

\FloatBarrier
\subsubsection{Prediction Model: Linear Model}\label{sec::results_regression}

\FloatBarrier
\subsubsection{Prediction Model: Neural Network}\label{sec::results_nnet}

\FloatBarrier
{  
\subsection{Comparison with a model-specific procedure: Boruta}
While model-specific procedures are not part of the model-agnostic study presented here, the random forest model-specific Boruta procedure \citep{kursa2010} might stand out to some readers as a similar approach to the CPI procedure. This circumstance merits further discussion and hence we will shed light on the relationship of this procedure to CPIseq in this SI Section.

Summarizing the Boruta procedure briefly, we can note that the basic idea is to add so-called shadow features to a random forest model and compare the random forest variable importance metrics calculated for both original and shadow features. Depending on how often an original features variable importance score exceeds the maximum variable importance score of all shadow features, the original features relevance will be classified as 'confirmed'. The goal of Boruta is to detect all relevant variables by repeatedly adding shadow features and compare variable importance scores. 

The Boruta algorithm requires having a variable importance measure at hand, in this case the random forest variable importance (a marginal measure of FI), and then wraps around the Boruta algorithm. In other words, Boruta is not a variable importance measure itself, but rather a procedure that utilizes an existing variable importance measure to detect all relevant variables. Therefore, this procedure is not a direct competitor of the methods we present in the paper. Further, Boruta was developed as a model-specific procedure for random forests, hence it is not suitable for our model-agnostic study. 

Nonetheless, we acknowledge that there are some similarities in the procedures from a first glance. The motivation to determine all relevant variables through statistical testing is similar in both Boruta and CPI. Moreover, adding knockoffs to the prediction model in CPI might seem similar to the Boruta algorithm which adds shadow features. However, Boruta compares the importance scores of the variables with the maximum importance score of any shadow variable, whereas CPI compares the difference in loss when the prediction model uses the original versus its corresponding knockoff variable. Another key difference is that Boruta does add shadow features by shuffling the variables values, which, similar to permutation FI, breaks correlational structures with both the target and other variables in the model. On the contrary, the knockoffs added by CPI maintain the correlational dependencies with other covariates. 

In sum, the Boruta approach turns out as quite different from the other approaches considered in this paper. Nonetheless, a brief comparison in the performance might be of interest to some readers. 

We ran Boruta on the experimental setup of Section 3.2 and find that Boruta  tends to label every variable as ‘important’ with increasing sample size, see Figure \ref{fig::boruta}.  The reason for this behavior is that Boruta relies on a marginal feature importance measure and thus has no type I error control for the conditional hypothesis. A similar picture arises when applying Boruta to the real world data example (Section 3.3), where the relevance of all variables is ‘confirmed’ by Boruta. 

 \begin{figure}[!htbp]
	\centering
	\includegraphics[trim={0 0 0 1cm},clip,width=1\textwidth]{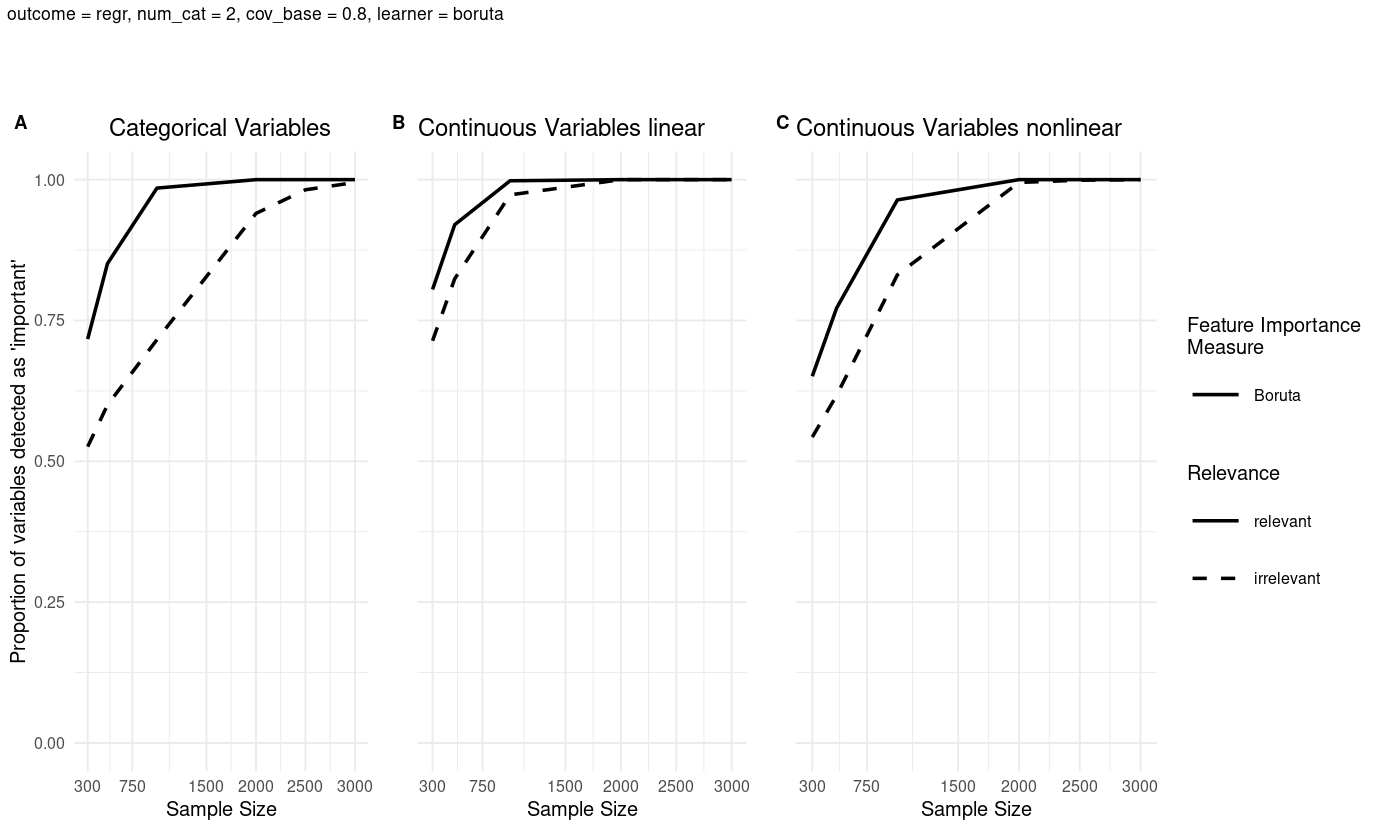}
	\caption{{ Proportion of variables for which relevance was classified as 'confirmed' by the Boruta procedure which relies on a random forest prediction model. The solid lines indicate relevant variables, whereas the dashed lines indicate irrelevant variables. Categorical variables exhibit $c=5$ levels, signal to noise ratio is set to $2$, pairwise correlation is $\rho = 0.8$ and $Y$ continuous.}}
	\label{fig::boruta}
\end{figure}
}
\section{Real-World Data}
\label{app::real_data}
We use the on OpenML publicly available \textit{diamonds} dataset (task ID $42225$).\footnote{\url{https://www.openml.org/search?type=data&sort=runs&id=42225&status=active}} A brief description of the covariates included is given in Table \ref{tab::diamonds}.{  
 From the publicly available \texttt{diamonds} data set, we use only round diamonds, i.e. diamonds that only deviate from a perfectly round shape by \texttt{x\_length} - \texttt{y\_width} $< 0.02$mm, yielding a subset of $N ~= 4\,463$ observations.
\begin{table}[h]
\centering
        \begin{tabular}{lll}
            \toprule
            Covariate & Data type & Description\\ \midrule
            \texttt{price} & continuous & price in USD \\
            \texttt{carat} & continuous & weight of the diamond in carat \\
            \texttt{cut} & categorical, $5$ levels & quality of the cut (Fair, Good, Very Good, Premium, Ideal)\\
            \texttt{color} &  categorical, $7$ levels & diamond colour, from J (worst) to D (best) \\
            \texttt{clarity} & categorical, $8$ levels &  measurement of how clear the diamond is \\
            &&(I1 (worst), SI2, SI1, VS2, VS1, VVS2, VVS1, IF (best))\\
            \texttt{x\_length} & continuous & length in mm\\
            \texttt{y\_width} & continuous & width in mm\\
            \texttt{z\_depth} & continuous & depth in mm\\
            \texttt{depth\_perc} & continuous & depth percentage = \texttt{z\_depth} / mean(\texttt{x\_length}, \texttt{y\_width}) \\
            && \hspace{2,35cm}= 2 $\cdot$ \texttt{z\_depth} / (\texttt{x\_length} + \texttt{y\_width}) \\
            \texttt{table} & continuous & width of diamond top relative to widest point\\\bottomrule
        \end{tabular}
    \caption{Codebook for the \textit{diamonds} data set.}
    \label{tab::diamonds}
\end{table}

\subsection*{Prediction Models}
We use a random forest with defaults of \texttt{R}-package \texttt{ranger}\footnote{\url{https://cran.r-project.org/web/packages/ranger/ranger.pdf}}, i.e. $500$ trees in the forest as prediction model. The model trained on $2/3$ of the data set and evaluated on the remaining $1/3$ of the data yields a good fit with $R^2 = 0.94$. The results are presented in Fig. \ref{fig::diamonds} in the main text.

For the sake of comparison, we recalculate the analysis for a neural network as prediction model. We use a single hidden layer network as given by the the defaults of \texttt{R}-package \texttt{nnet}\footnote{\url{https://cran.r-project.org/web/packages/nnet/nnet.pdf}} with modifications to the size of the hidden layers (number of units in the hidden layer is set to 100) and a weight decay of $0.15$. The model is trained on the same 2:1 train/test split as with the random forest prediction model and we find a similar good fit. Namely, the neural network achieves $R^2= 0.93$ on the test dataset. However, we can see LOCO losing power in statistical testing. We attribute this to instabilities in neural network model retraining, which is an inevitable part of the LOCO subroutine. Further, the neural network seems to be affected by the high colinearity of \texttt{x\_length} and \texttt{y\_width}, which may explain the different importance score outcomes for the conditional subgroup approach and SAGE in comparison to the random forest model. Besides this, the results appear to be robust against the change in prediction model, which underpins the model-agnosticity of the FI measurement procedures.  We present results from the neural network learner in Figure \ref{fig::diamonds_nnet}. 

\begin{figure}[ht!]
\includegraphics[trim={0 0 0 0cm},clip, width=1\textwidth]{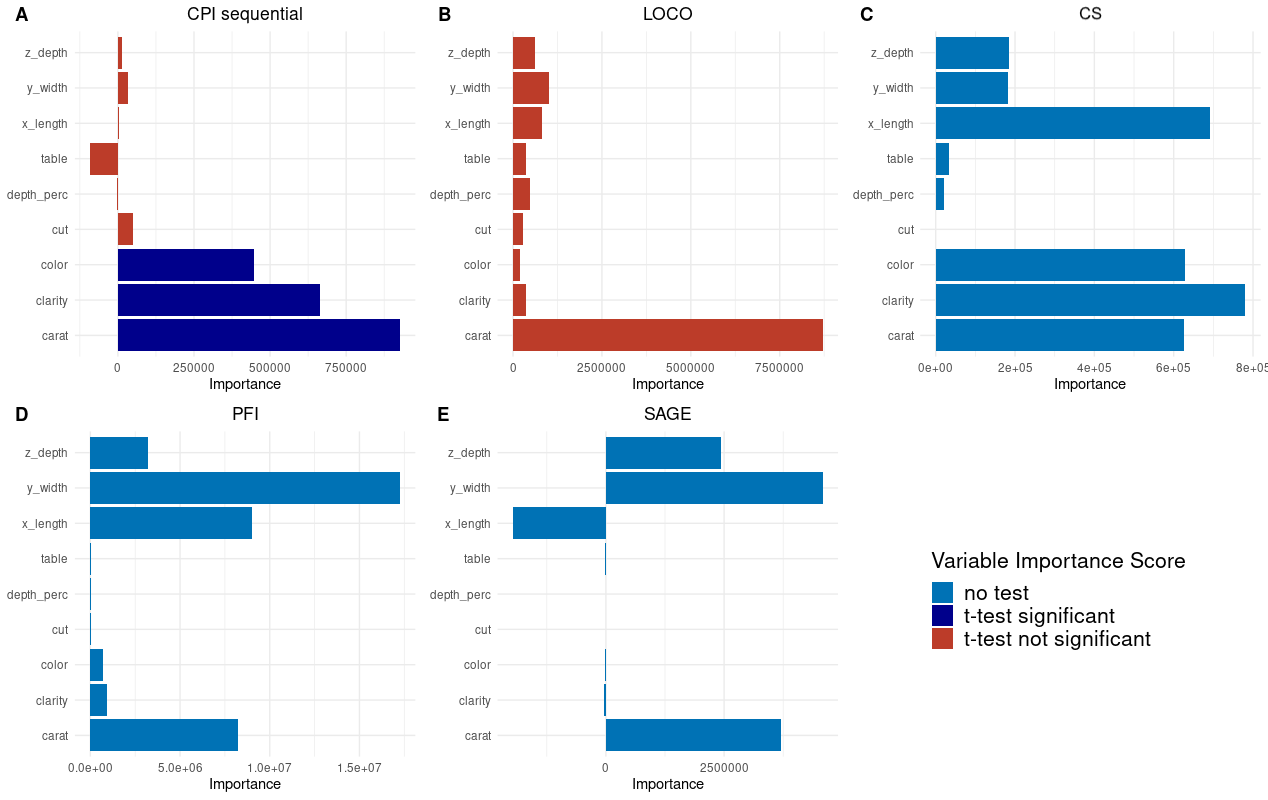}
\centering
\caption{{ Feature importance scores for predicting the selling price of diamonds using a neural network model. For the CPIseq and LOCO, t-tests are at $\alpha ~= 5\%$, using the Holm procedure to adjust for multiple testing.}}\label{fig::diamonds_nnet}       
\end{figure} 
}

\end{document}